\documentclass[a4paper,fleqn]{cas-sc}

\usepackage{longtable,subfigure,lineno}
\usepackage[numbers]{natbib}
\usepackage{placeins}

\usepackage{etoolbox}
\AtBeginDocument{%
  \robustify\cite
}

\newcommand{\sitecount}{205 }
\newcommand{\siteyears}{275 }
\newcommand{\scencecount}{38,187 }
\newcommand{\scencecountround}{38,000 }
\newcommand{\coo}{$CO_2$ }

\newcommand{\fluxunit}{$\mu\mathrm{mol}\,\mathrm{CO_2}\,\mathrm{m}^{-2}\,\mathrm{s}^{-1}$}
\newcommand{\yearlyfluxunit}{$\mathrm{g}\,\mathrm{C}\,\mathrm{m}^{-2}\,\mathrm{yr}^{-1}$}

\def\tsc#1{\csdef{#1}{\textsc{\lowercase{#1}}\xspace}}
\tsc{WGM}
\tsc{QE}
\tsc{EP}
\tsc{PMS}
\tsc{BEC}
\tsc{DE}

\begin{document}
\let\WriteBookmarks\relax
\def\floatpagepagefraction{1}
\def\textpagefraction{.001}
\shorttitle{FAR}

\shortauthors{J. Searcy et~al.}

\title [mode = title]{A Footprint-Aware, High-Resolution Approach for Carbon Flux
Prediction Across Diverse Ecosystems}                      



\author[1]{Jacob Searcy}[orcid=0000-0003-3544-3066]
\cormark[1]
\ead{jsearcy@uoregon.edu}
\credit{Conceptualization, Data curation, Software, Methodology, Visualization, Funding acquisition, Supervision, Writing -- original draft}

\author[2]{Anish Dulal}
\credit{Data curation, Software, Methodology, Visualization, Writing -- review \& editing}

\author[3]{Courtney Mathers}
\credit{Writing -- review \& editing}

\author[3,4,5]{Scott D. Bridgham}
\credit{Funding acquisition, Writing -- review \& editing}

\author[1,4]{Ashley Cordes}
\credit{Funding acquisition, Writing -- original draft, Writing -- review \& editing}

\author[3,4]{Lillian Aoki}
\credit{Funding acquisition, Writing -- review \& editing}

\author[3,5]{Brendan Bohannan}
\credit{Funding acquisition, Writing -- review \& editing}

\author[6]{Qing Zhu}
\credit{Data curation, Writing -- review \& editing}

\author[3,4,5]{Lucas C. R. Silva}
\credit{Conceptualization, Funding acquisition, Project administration, Writing -- review \& editing}

\affiliation[1]{organization={Department of Data Science, University of Oregon}}

\affiliation[2]{organization={Department of Computer Science, University of Oregon}}

\affiliation[3]{organization={Institute of Ecology and Evolution, University of Oregon}}

\affiliation[4]{organization={Environmental Studies Program, University of Oregon}}

\affiliation[5]{organization={Department of Biology, University of Oregon}}

\affiliation[6]{organization={Climate and Ecosystem Sciences Division, Lawrence Berkeley National Laboratory}}

\cortext[cor1]{Corresponding author}

\begin{abstract}
Eddy-covariance (EC) flux towers provide in situ
measurements of \coo flux and serve as the ground-truth data for
predictive `upscaling' models derived from satellite products. However,
many satellites now resolve spatial scales smaller than
an EC tower's footprint. We show theoretically that upscaling models trained on high-resolution data in heterogeneous landscapes must account
for an EC tower's footprint to avoid bias in pixel-level predictors. To address this problem, we introduce Footprint-Aware Regression (FAR),
a deep-learning framework that simultaneously predicts spatial footprints and
pixel-level estimates of \coo flux, and show it yields unbiased pixel-level predictions given sufficient training data. We demonstrate FAR on our AMERI-FAR25 dataset, which combines \siteyears site-years of tower data with corresponding Landsat scenes, and show that FAR outperforms non-footprint-aware models. FAR increased half-hourly $R^2$ from approximately 0.575 to 0.634 and reduced RMSE by about 7\% relative to the best fixed-footprint baseline on a dataset of withheld sites. Gains are larger for monthly and yearly averages relative to a coarser 990~m baseline. Site-level analyses show that these gains extend across multiple ecosystem types. These performance gains are observed whether footprints are learned jointly or estimated independently using an established footprint model, despite substantial variation in footprint size between methods.
In a regional comparison over the Western Cascades, FAR produces flux estimates with a distribution comparable to that of existing high-resolution process-based models.
\end{abstract}

\begin{highlights}
\item FAR jointly learns dynamic tower footprints and 30~m resolution \coo flux predictions.
\item AMERI-FAR25 pairs 275 site-years of EC data with \scencecount Landsat scenes.
\item Footprint-aware models outperform non-footprint-aware baselines.
\end{highlights}


\boolfalse{usecasgrabsbox}
\boolfalse{usecashlsbox}
\maketitle

\section{Introduction}

Eddy covariance (EC) towers provide in situ land-based measurements of
\coo flux by tracking rapid fluctuations in vertical wind
speed and gas concentrations as atmospheric eddies transport air parcels
between the surface and the atmosphere. EC tower data are increasingly used
to train `upscaling' models used to predict \coo fluxes
from satellite data. However, an obstacle in utilizing EC tower data for
predictive models is their measurement footprint, which describes the land area that contributes to the observed flux. This upwind
footprint varies with tower height, ecosystem properties, and
atmospheric conditions, with 80\% footprint climatologies (areas containing, on average, 80\% of a tower's footprint) spanning 
$10^3$--$10^7~\mathrm{m}^2$~\cite{Chu2021RepresentativenessSites}. This setting creates several scale mismatches that need to be addressed by any upscaling model. Spatially, an EC tower’s footprint changes with wind conditions, resulting in a variable measurement area, whereas satellite products are measured on fixed grids. Grid-cell size also varies substantially among satellite products; for example, MODIS products are commonly available at 250--1{,}000~m resolution~\cite{justice1998modis_land_remote_sensing}, whereas Landsat products provide 30~m resolution~\cite{eros2020landsat}. Depending on the satellite product, tower, and atmospheric conditions, a tower's footprint may overlap only part of a satellite pixel or span multiple pixels. Temporally, EC towers often provide data at half-hourly intervals, whereas satellite sampling frequency depends on factors such as orbit, swath width, and constellation size. Coarser-resolution products such as MODIS can provide approximately daily observations, while Landsat provides 30~m measurements at an approximately eight-day revisit interval or longer. In practice, the availability of usable satellite observations may be further constrained by cloud cover.

A typical approach to this problem is to temporally process and spatially average data to coarser, matching resolutions. Examples of this approach include FLUXCOM~\cite{Jung2020ScalingApproach}, EC-GPR-TCFs~\cite{Reyes-Munoz2025Tower-to-globalData}, LPJ-GUESS~\cite{MartinBelda2022LPJ-GUESS/LSMv1.0:Realism}, and SCOPE-GPR-TCF~\cite{Reyes-Munoz2024InferringModels}, with SCOPE-GPR-TCF providing the highest-resolution product, with predictions at 300~m resolution every 8 days. However, due to its footprint, EC tower data cannot be exactly mapped to a fixed grid without assumptions that the average area measured by the tower's footprint is representative of the spatial area covered by the target grid. In this work, we explore the use of satellite predictors using pixels smaller than that of a footprint, and leverage the fact that while satellite data are static between measurements, each tower's footprint is not, which can potentially allow us to learn more about how each pixel contributes to the EC tower's measurement.

The use of higher resolution, both temporally and spatially, substantially increases computational burden and complexity, but has the potential to improve flux predictions, particularly when those models are trained on EC towers situated in spatially heterogeneous regions. There is growing evidence that accounting for this spatial heterogeneity is crucial for accurate gas flux predictions.
For example, spatial representativeness within a heterogeneous landscape was the dominant uncertainty in upscaling net ecosystem production (NEP) in a case study of 17 EC towers~\cite{Ran2016SpatialScale}. Furthermore, tower footprints may cover more than one land cover type~\cite{Chu2021RepresentativenessSites,Fang2024AssessingFLUXNET2015}, which
has the potential to bias predictors. Difficulties
with spatial heterogeneity have also been found in EC tower measurements
of methane~\cite{Peltola2015StudyingHeight}. There is evidence that such spatial
heterogeneity could be accounted for by using high-resolution satellite
data. For example, high-resolution satellite measurements within an EC
tower's footprint have improved correlations with EC tower flux measurements~\cite{Kong2022MatchingEstimates}. However, there is currently no analysis method
that can be used to account for spatial heterogeneity across large multi-site EC tower datasets. Additionally, development toward this goal is limited by a lack of existing datasets and benchmarks with
spatially resolved satellite imagery~\cite{fortier2025carbonsense}.
Previous work toward spatially explicit predictions at single towers includes environmental response functions, which relate flux measurements to variables that are dynamically weighted by footprints~\cite{Xu2017UpscalingFunctions}. We expand on this work by introducing footprint-aware regression (FAR), a methodology for training unbiased pixel-level estimators, and AMERI-FAR25, a dataset of
\siteyears site-years of EC tower data and over \scencecountround associated 30~m $\times$ 30~m resolution
Landsat scenes (satellite image acquisitions). FAR utilizes data streams at different spatial and temporal resolutions, with the goal of extracting ecosystem properties from satellite data and modeling the \coo flux of that ecosystem as a function of environmental drivers such as incoming solar radiation. It utilizes a novel attention mechanism~\cite{Ilse2018Attention-basedLearning}, a commonly used modeling technique that enables a model to identify and give greater weight to the most informative regions of the input data. We modify the standard approach by predicting the ``attention weights'' from atmospheric variables and treating them as a learned footprint. We demonstrate theoretically that this approach can produce unbiased predictions of \coo flux at high spatial resolution, and show empirically that this approach generally improves agreement with EC tower measurements when compared to spatially aggregated models.

\section{Theoretical framing}

In the following sections, we motivate the need for improved machine learning models to predict \coo flux at higher spatial resolutions by
describing two related issues: (i) bias induced by \emph{scale mismatch}
when applying predictors trained on aggregated inputs to fine-scale
targets, and (ii) \emph{footprint bias} arising from heterogeneous land
cover coupled with changing footprint weights. We then describe FAR,
which jointly learns a pixel-scale flux model and footprint weights from
tower-scale supervision, and derive conditions in which it is identifiable. Throughout this section, we use ``bias'' to refer to systematic error in the
conditional mean. A predictor is biased if, among observations with the same
predictor values, its errors do not average to zero, and is unbiased otherwise.

We model the flux measured by an eddy-covariance (EC) tower as a
footprint-weighted average of fine-scale surface exchange discretized
into $N$ pixels indexed by \(i \in \{ 1,\ldots,N\}\) with each measurement indexed by \(t\). Note that FAR does not use historical data to make its prediction. Let \(y_{t,i}\) denote the fine-scale flux at pixel \(i\)
and time \(t\). Let \(\alpha_{t,i}\) denote the true tower footprint weight
for pixel \(i\) at time \(t\), satisfying
\begin{equation}
\alpha_{t,i} \geq 0,\quad\quad\sum_{i = 1}^{N}\alpha_{t,i} = 1.
\label{eq:obs}
\end{equation}

The tower observation is

\begin{equation}Y_{t}\mspace{6mu} = \mspace{6mu}\sum_{i = 1}^{N}\alpha_{t,i}\, y_{t,i}\mspace{6mu} + \mspace{6mu}\varepsilon_{t},
\label{eq:aggregate}
\end{equation}

where \(\varepsilon_{t}\) represents tower measurement error.

\subsection{High-resolution scale mismatch}
\label{sec:resolution_mismatch}
Our objective is to estimate the pixel-level flux \(y_{t,i}\) from the
available tower-scale aggregate \(Y_{t}\) (see Equation~\ref{eq:aggregate}). A common strategy used for upscaling is to aggregate pixel predictors to match a tower-level model.
Let \(X_{t,i}\) denote pixel-level predictors for pixel \(i\) at time
\(t\). Define a weighted average predictor:
\(w_{t,i}\):    
\begin{equation}{\bar{X}}_{t}\mspace{6mu} = \mspace{6mu}\sum_{i = 1}^{N}{w}_{t,i}\, X_{t,i},\quad\quad{w}_{t,i} \geq 0,\quad\sum_{i = 1}^{N}{w}_{t,i} = 1.\end{equation}

A tower-scale predictor is then fit using this average predictor as

\begin{equation}{\widehat{Y}}_{t}\mspace{6mu} = \mspace{6mu} f_{\theta}\left( {\bar{X}}_{t} \right),\end{equation}

with parameters \(\theta\) learned from tower observations with a machine learning model. We say the
tower-scale predictor is \emph{unbiased} with respect to the available
aggregated predictors if it has zero conditional mean error,

\begin{equation}\mathbb{E}\left\lbrack Y_{t} - f_{\theta}\left( {\bar{X}}_{t} \right)\, \middle| \,{\bar{X}}_{t} \right\rbrack = 0,\end{equation}
or equivalently
\(\mathbb{E}\left\lbrack Y_{t} \mid {\bar{X}}_{t} \right\rbrack = f_{\theta}\left( {\bar{X}}_{t} \right)\).

The key question is whether a predictor that is an unbiased estimator at the tower
scale can also be an unbiased estimator of fine-scale \(y_{t,i}\). If we
assume this is true; then:

\begin{equation}\mathbb{E}\left\lbrack \sum_{i = 1}^{N}\alpha_{t,i}f_{\theta}\left( X_{t,i} \right)\, \middle| X_{t,i} \right\rbrack\mspace{6mu} = \mspace{6mu}\mathbb{E}\left\lbrack f_{\theta}\left( \sum_{i = 1}^{N}{w}_{t,i}X_{t,i} \right)\, \middle| \, X_{t,i} \right\rbrack,\end{equation}

using our assumption of unbiased estimators implies

\begin{equation}\sum_{i = 1}^{N}\alpha_{t,i}f_{\theta}\left( X_{t,i} \right)\mspace{6mu} = f_{\theta}\left( \sum_{i = 1}^{N}{w}_{t,i}X_{t,i} \right)
\label{eq:jsen-condition}
\end{equation}

However, based on Jensen's equality condition~\cite{Durrett2019ProbabilityExamples}, the equality above only
holds generally if \(f_{\theta}\) is affine (i.e., a linear model) and
\({w}_{t,i} = \alpha_{t,i}\) or if \(X_{t,i}\) is
constant across pixels (\(X_{t,i} = c_{t}\)). For general nonlinear
\(f_{\theta}\), Equation~\ref{eq:jsen-condition} does not hold even if \(w_{t,i} = \alpha_{t,i}\).
Consequently, even if \(f_{\theta}\left( {\bar{X}}_{t} \right)\) is an
unbiased estimator of the tower aggregate \(Y_{t}\), the same mapping
\(f_{\theta}\left( X_{t,i} \right)\) cannot, in general, be an unbiased
estimator of the fine-scale fluxes \(y_{t,i}\).

\subsection{Footprint bias}
\label{sec:footprintbias}
A related issue arises when model training and application occur under
different footprint weights, such as might occur in a heterogeneous landscape with changing footprints. Consider two cases that share the same
pixel-level fluxes \(y_{i}\) and site-level predictors ($\bar{X}^{(1)}=\bar{X}^{(2)}$) but have different footprint weights
\(\alpha^{(1)}\) and \(\alpha^{(2)}\):

\begin{equation}Y^{(1)} = \sum_{i = 1}^{N}\alpha_{i}^{(1)}\, y_{i},\quad\quad Y^{(2)} = \sum_{i = 1}^{N}\alpha_{i}^{(2)}\, y_{i}.\end{equation}

Let \(f_{\theta}\left( \bar{X} \right)\) be an unbiased estimator of
\(Y^{(1)}\), i.e., an estimator trained on aggregated data from case (1). Now consider applying the same predictor \(f_{\theta}\) under
case (2).

In this case, the predictor output is the same in both cases:
\(f_{\theta}\left( {\bar{X}}^{(2)} \right) = f_{\theta}\left( {\bar{X}}^{(1)} \right)\).
Therefore, the conditional bias of this predictor for case (2) is

\begin{equation}\begin{matrix}
{Bias}^{(2)} & {: =}\mathbb{E}\left\lbrack f_{\theta}\left( {\bar{X}}^{(2)} \right) - Y^{(2)}\, \middle| \,{\bar{X}}^{(2)} \right\rbrack \\
 & = \mathbb{E}\left\lbrack f_{\theta}\left( {\bar{X}}^{(1)} \right) - Y^{(2)}\, \middle| \,{\bar{X}}^{(1)} \right\rbrack \\
 & = \mathbb{E}\left\lbrack Y^{(1)} - Y^{(2)}\, \middle| \,{\bar{X}}^{(1)} \right\rbrack \\
 & = \sum_{i = 1}^{N}\left( \alpha_{i}^{(1)} - \alpha_{i}^{(2)} \right)\, y_{i}.
 \label{eq:footprint_bias}
\end{matrix}\end{equation}

Equation~\ref{eq:footprint_bias} is zero when
\(y_{i}\) is constant over space, when the footprints match
(\(\alpha^{(1)} = \alpha^{(2)}\)), or under incidental cancellations
between fluxes. Otherwise, an unbiased predictor trained
under case (1) will generally exhibit nonzero bias when applied under
case (2), even when the predictor inputs are unchanged, because the
target aggregate differs solely due to a change in footprint weights.
In other words, the exact same land can produce different flux predictions, and
from the previous argument, neither of these measurements when used to train a predictor with aggregated input features necessarily produces an unbiased estimator of the
true flux.

It is important to note that many flux prediction algorithms do approximate the requirements
for unbiased estimators. EC towers are often placed to maximize the homogeneous land around them by design, and measured predictors such as the normalized
difference vegetation index (NDVI)~\cite{Tucker1979RedVegetation,Rouse1974MonitoringERTS} can exhibit linear relationships with monthly measured gross primary productivity. However, our aim is to enable unbiased predictors in heterogeneous landscapes with any nonlinear predictor.

\subsection{Footprint-Aware Regression (FAR)}
\label{sec:far_theory}

In its most general form, FAR jointly estimates (i) a pixel-scale flux predictor based on predictors $x^{flux}_t$ (see Section~\ref{sec:far_dnn}) and parameterized by $\theta$, and (ii) footprint weights predicted from observable atmospheric conditions $x^{footprint}_t$ (see Section~\ref{sec:far_dnn}) and parameterized by $\phi$.

\begin{equation}\hat{Y}^{obs}_{t,\theta,\phi} = \sum_i \hat{\alpha}_{t,i,\phi} \hat{y}_{t,i,\theta} = \sum_i f^\alpha_{\phi,i}\!\left(x^{footprint}_t\right) f^y_\theta\!\left(x^{flux}_{t,i}\right)\end{equation}

\begin{equation}
\hat{\alpha}_{t,i,\phi}\geq 0,
\qquad
\sum_i \hat{\alpha}_{t,i,\phi}=1
\end{equation}

In this section, we address the properties that FAR and its dataset must have to learn an unbiased pixel-level predictor, with FAR's specific deep learning implementation described in Section~\ref{sec:far_dnn}. We consider the dataset optimization problem

\begin{equation}
(\theta^\star,\phi^\star)
=
\arg\min_{\theta,\phi}
\mathcal L(\theta,\phi),
\qquad
\mathcal L(\theta,\phi)
=
\mathbb E\left[
\left(
Y_t^{obs}
-
\sum_i
f^\alpha_{\phi,i}(x_t^{footprint})
f^y_\theta(x_{t,i}^{flux})
\right)^2
\right].
\end{equation}

The central question is when the minimizer
$f^y_{\theta^\star}$ is an unbiased pixel-level predictor. To start, define (i) $m^\alpha_{t,i}$ the
conditional-mean of the true contribution of pixel $i$ given the observed atmospheric conditions, and (ii) $m^y_{t,i}$ the conditional-mean of the true pixel-level flux of pixel $i$ given the measured satellite and environmental variables. 
\begin{equation}
m^\alpha_{t,i}
:=
\mathbb{E}\left[
\alpha_{t,i}
\mid
x_t^{footprint}
\right],
\qquad
m^y_{t,i}
:=
\mathbb{E}\left[
y_{t,i}
\mid
x_{t,i}^{flux}
\right].
\label{eq:conditional_components}
\end{equation}

We assume that, for each pixel $i$, 
\begin{equation}
\begin{aligned}
&\mathbb{E}\left[
\alpha_{t,i}y_{t,i}
\;\middle|\;
x_t^{footprint},
\left\{x_{t,j}^{flux}\right\}_{j=1}^{N}
\right] \\
&\qquad =
\mathbb{E}\left[
\alpha_{t,i}
\mid
x_t^{footprint}
\right]
\mathbb{E}\left[
y_{t,i}
\mid
x_{t,i}^{flux}
\right],
\end{aligned}
\label{eq:conditional_factorization}
\end{equation}
This assumption implies that after accounting for the measured environmental and atmospheric variables, any unexplained variation does not systematically alter how strongly that location contributes to the tower measurement. If we also assume that the tower measurement error has conditional mean zero, then the conditional mean of the observed tower flux ($\mu_t$) is the expected footprint contribution times the expected pixel flux:
\begin{equation}
\begin{aligned}
\mu_t
&:=
\mathbb{E}\left[
Y_t^{\mathrm{obs}}
\;\middle|\;
x_t^{footprint},
\left\{x_{t,i}^{flux}\right\}_{i=1}^{N}
\right] \\
&=
\mathbb{E}\left[
\sum_{i=1}^{N}\alpha_{t,i}y_{t,i}+\varepsilon_t
\;\middle|\;
x_t^{footprint},
\left\{x_{t,i}^{flux}\right\}_{i=1}^{N}
\right] \\
&=
\sum_{i=1}^{N}
\mathbb{E}\left[
\alpha_{t,i}
\mid
x_t^{footprint}
\right]
\mathbb{E}\left[
y_{t,i}
\mid
x_{t,i}^{flux}
\right] \\
&=
\sum_{i=1}^{N}m^\alpha_{t,i}m^y_{t,i}.
\end{aligned}
\label{eq:conditional_mean}
\end{equation}

Therefore, if the FAR predictor is sufficiently flexible to represent
$\mu_t$ and the dataset is sufficiently large, then the minimizer will satisfy.
\begin{equation}
\sum_{i=1}^{N}
f^\alpha_{\phi^\star,i}\left(x_t^{footprint}\right)
f^y_{\theta^\star}\left(x_{t,i}^{flux}\right)
=
\mu_t
\end{equation}

In other words, the prediction of the aggregated flux is expected to be unbiased by construction. However, we need to establish whether the factorization of $\mu_t$ into footprint weights and flux components can recover an unbiased pixel-level flux prediction. To examine this, we write the flux prediction as the predictable conditional mean of the true fine-scale flux and a residual.
\begin{equation}
f^y_\theta(x_{t,i}^{flux})=m^y_{t,i}+b_{t,i}(\theta),
\end{equation}

and similarly represent the learned footprint weights with their residual.

\begin{equation}
f^\alpha_{\phi,i}(x_t^{footprint})=m^\alpha_{t,i}+\delta_{t,i}(\phi) \;\mathrm{where}\;\sum_i\delta_{i,t}=0.
\end{equation}

Substituting into the aggregated predictor yields

\begin{equation}
\sum_i
f^\alpha_{\phi^\star,i}(x_{t}^{footprint})
f^y_{\theta^\star}(x_{t,i}^{flux})
=
\sum_i m^\alpha_{t,i} m^y_{t,i}
+
\sum_i m^\alpha_{t,i} b_{t,i}(\theta^\star)
+
\sum_i \delta_{t,i}(\phi^\star) m^y_{t,i}
+
\sum_i \delta_{t,i}(\phi^\star) b_{t,i}(\theta^\star)=\mu_t.
\end{equation}

From Equation~\ref{eq:conditional_mean}, the first term $\sum_i m^\alpha_{t,i} m^y_{t,i}=\mu_t$, i.e., the expected tower flux, implying the error terms 
\begin{equation}
\sum_i m^\alpha_{t,i} b_{t,i}(\theta^\star)
+
\sum_i \delta_{t,i}(\phi^\star) m^y_{t,i}
+
\sum_i \delta_{t,i}(\phi^\star) b_{t,i}(\theta^\star)=0.
\label{eq:bias}
\end{equation}
The above equation holds trivially when the bias of the predicted pixel fluxes $b_{t,i}(\theta^\star)=0$ and the error in the predicted footprint weights $\delta_{t,i}(\phi^\star)=0$. However, non-zero bias may remain as long as Equation~\ref{eq:bias} holds.
This implies that pixel-level bias can persist if it is offset by an error in the learned footprint weights, or if there are systematic cancellations in the bias such that $\sum_i m^\alpha_{t,i} b_{t,i}(\theta^\star)=0$. However, this can be further constrained by data.
\\\\
\textbf{Homogeneous contributing regions:} Assume there exist periods in which the fine-scale flux and input variables are (approximately) constant over the contributing footprint region, where the true and expected footprint weights $\alpha_{t,i}\ne0$ or $ \hat{\alpha}_{t,i}\ne0$. In this case, the bias and conditional mean flux are constant for a given $t$.
\begin{equation}b_{t,i}(\theta)=\bar{b}_t \;\; m^y_{t,i}=\bar{m}_t\end{equation}
Equation~\ref{eq:bias} becomes:
\begin{equation}
 \bar{b}_{t}\sum_i m^\alpha_{t,i}
+
\bar{m}^y_{t}\sum_i \delta_{t,i}(\phi^\star) 
+
\bar{b}_{t}\sum_i \delta_{t,i}(\phi^\star) =\bar{b}_{t}=0.
\end{equation}

This implies that FAR becomes an unbiased ($b_t$=0) flux predictor if homogeneous regions exist for all relevant values of $x^{flux}$. However, this is not true for $\delta(\phi^\star)$ which may continue to have non-zero bias, since with $b_{t,i}=0$ Equation~\ref{eq:bias} becomes:

\begin{equation}
\sum_i \delta_{t,i}(\phi) m^y_{t,i}=0
\end{equation},

This is true for any $m^y_{t,i}$ constant over non-zero $\delta_{t,i}(\phi^\star)$, and could potentially hold in the presence of incidental flux cancellations. Similarly, substantial heterogeneity is required to constrain the footprint weights $f^\alpha_{\phi,i}(x_t^{footprint})$  because they only come from differences in the aggregated fluxes such as those described in Section~\ref{sec:footprintbias}. However, even if the footprint is not perfectly constrained, this does not impact the unbiased prediction of the pixel-level flux. Note that this is significantly less stringent than assuming homogeneity across the whole input for all times, which is what must be assumed with a fixed footprint.

The above proof relies on an assumption that sufficient data exist to anchor FAR across the relevant input space. This may not hold in practice, so additional model constraints are imposed to limit the potential for bias on the pixel-level flux prediction. Equation~\ref{eq:bias} shows that coordinated errors in the inferred footprint weights could offset corresponding changes in the predicted pixel-level flux. We therefore limit the capacity for such coordination in two ways. First, augmentations are applied by rotating the image and wind vector to produce a new artificial scene that is identical to the original, but rotated by an arbitrary angle, with the wind direction rotated by the same angle. This removes one possible source of coordination, where the footprint model could potentially access flux through correlations with the prevailing wind direction. Second, the flux predictors ($x_{t,i}^{flux}$) contain no spatial information about their location in the image, so the flux predictor cannot assign systematically different flux values solely because a pixel lies, for example, on the left rather than the right side of the image.

In this section, we have shown that scale mismatches and footprint bias both have the potential to create biased pixel-level predictors. FAR addresses both issues by learning
pixel-scale predictions and footprint weights jointly, enforcing consistency with the tower aggregation across time while exploiting both homogeneous and heterogeneous regimes to constrain the fit. We also show that if there is sufficient data where footprints cover homogeneous areas, FAR becomes an unbiased pixel-level predictor. Below in Section~\ref{sec:far_dnn}, we address how we implemented the pixel-level flux and footprint predictions with a two-armed deep learning model, and the dataset we used to train FAR.

\section{Methods}

In this section, we describe the data, training, and evaluation of FAR. FAR requires data from several sources, EC towers for fluxes and locally measured environmental drivers and satellite data for remotely sensed environmental properties. For upscaling when tower data is not available FAR requires estimates of these environmental drivers. These data are described in Section~\ref{sec:data}. These data are then used to train a deep learning model described in Section~\ref{sec:far_dnn}, which is evaluated based on criteria in Section~\ref{sec:eval}.

\subsection{AMERI-FAR25 Dataset}
\label{sec:data}
Although satellite data have long been combined with EC measurements, relatively few reference datasets are distributed for direct algorithm comparison~\cite{fortier2025carbonsense}, and those datasets lack spatial imagery at the scales considered here. To train FAR, we created the AMERI-FAR25 Dataset, which is a composite of \siteyears site-years of EC tower data combined with \scencecount satellite scenes and represents the first publicly available dataset that provides a large-scale pairing of satellite imagery and EC tower measurements. In this section, we describe the processing and variables available in our dataset.

\subsubsection{EC tower data}

We obtained EC tower data from the AmeriFlux BASE product
~\cite{Chu2023AmeriFluxSharing} for all sites releasing data under the CC-BY-4.0
license with downloadable half-hour flux data, and we follow its variable naming convention below. We required sites to have
measurements of wind direction (WD [$^\circ$]), wind speed (WS [m~s$^{-1}$]), carbon flux (FC [$\mu$mol~CO$_2$~m$^{-2}$~s$^{-1}$]),
friction velocity (USTAR [m~s$^{-1}$]), air temperature (TA [$^\circ$C]), sensible heat turbulent
flux (H [W~m$^{-2}$]), and relative humidity (RH [\%]); any sites without these variables
were excluded. The remaining sites were required to have a measure of
incoming radiation, either shortwave (SW\_IN [W~m$^{-2}$]) or photosynthetic flux
density (PPFD\_IN [$\mu$mol~photons~m$^{-2}$~s$^{-1}$]). SW\_IN was used for sites where SW\_IN was
available. For sites with only PPFD\_IN available, SW\_IN was predicted
with a linear scaling of PPFD\_IN. Sites without existing tower height
metadata were then dropped, resulting in \sitecount sites shown in Table~\ref{site-table}. For stations with
multiple sensors measuring the same quantity, the highest sensor(s) were
utilized, and replicates at this height were averaged. For sites reporting a storage correction term (SC [\fluxunit]), this term was added to the measured flux (FC [\fluxunit]). Outlier FC
measurements below the 0.5\% or above the 99.5\% percentiles were dropped to
prevent divergences during training, as were implausible negative SW\_IN values and measurements where carbon drawdown (FC \textless{} 0) was measured with
no incoming radiation (e.g. nighttime where SW\_IN=0). Each half-hour measurement
was marked as passing or failing a $u_*$ threshold derived using the moving point method described in~\cite{Papale2006TowardsEstimation}. Failing measurements were removed from the training data. 

\subsubsection{Meteorological data for upscaling}

In addition to local tower measurements, we utilize 30-year normals obtained from PRISM~\cite{PRISMUniversity}. Each normal represents a historical average value of each variable at a given place and time during the year. We utilize 800~m daily normals for TA and RH (treated as constant
throughout the day) at 800~m resolution and monthly normals for effective cloud transmittance, which is an estimate of the ratio of ground-level solar radiation to that expected under clear-sky conditions. Monthly effective cloud transmittance was used in combination with the software package \emph{pysolar}~\cite{stafford_pysolar} to estimate half-hour solar radiation at each site (SW\_IN). A current limitation of this approach is the lack of half-hour resolution for TA, RH, andeffective cloud transmittance, which we compare to full-resolution tower data.

\subsubsection{Landsat Data}

Satellite imagery was taken from Landsat 8 and 9 Level 1 product. This product contains 11 bands: eight multispectral bands at 30~m
resolution, one panchromatic band at 15~m resolution, and two thermal
infrared bands acquired at 100~m resolution and resampled to
30~m. These bands
are Coastal aerosol (0.43--0.45 $\mu m$), Blue (0.45--0.51 $\mu m$), Green
(0.53--0.59 $\mu m$), Red (0.64--0.67 $\mu m$), Near Infrared (NIR) (0.85--0.88 $\mu m$),
Shortwave Infrared (SWIR) 1 (1.57--1.65 $\mu m$), SWIR 2 (2.11--2.29 $\mu m$),
Panchromatic (0.50--0.68 $\mu m$), Cirrus (1.28--1.36 $\mu m$), Thermal Infrared
(TIRS) 1 (10.6--11.19 $\mu m$) and TIRS 2 (11.50--12.51 $\mu m$). Together
these bands provide significant information and are currently used to
derive a variety of environmental proxies such as NDVI~\cite{Tucker1979RedVegetation,Rouse1974MonitoringERTS}, the enhanced
vegetation index (EVI)~\cite{Huete2002OverviewIndices}, the normalized
difference water index (NDWI), and normalized burn ratio (NBR)~\cite{Key1999TheSeverity.}. Landsat scenes (satellite image acquisitions) that contained a flux tower were
downloaded with the \textit{landsatxplore} tool~\cite{forget_landsatxplore}. Patches of
128 $\times$ 128 pixels (3.84~km~$\times$~3.84~km) centered on each tower's location were extracted from all
Landsat scenes resulting in a total of \scencecount Landsat patches in the dataset. The original Landsat Level-1 digital values are used in the dataset.
Scenes were processed so pixels that were missing, cloudy, or had missing data were replaced with the most recently
available value for that pixel. Specifically, pixels were retained using the Landsat 8/9 Collection 2 QA\_PIXEL, selecting land and water pixels with low probability of cloud, cloud-shadow, snow/ice, and cirrus contamination (QA\_PIXEL$\in \{21824, 21888, 21952\}$). This reflects a workflow in which
the most recently available information is used for prediction. A one-month buffer before the start date of each site was used, and any scenes with pixels having no current or previous value were excluded. For each Landsat patch, all bands except for the panchromatic one (which provides
broad-spectrum measurements at a higher spatial resolution) are
concatenated with four orientation angles (Solar/Sensor, Azimuth/Zenith), and a binary mask denoting whether a pixel was replaced to form our input \(x_T^{landsat} \in \mathbb{R}^{128\times128\times15}\).

\subsubsection{Training and Evaluation Splits}

AMERI-FAR25 was split into different training, validation, and test sets. Training
sets were used during training to fit model weights. The validation sets
were used during training to determine if a model was overfit, and the
test sets were only used for evaluation. For testing FAR's upscaling,
the \sitecount sites were grouped by International Geosphere--Biosphere
Programme (IGBP) definition codes~\cite{Loveland01012000} (denoting ecosystem types, e.g.
deciduous forest, wetland, etc.), and for all IGBP groups with 10 or
more sites, 40\% of those sites were withheld and evenly split between
the validation (\textit{val\_site}) and test set (\textit{test\_site}). This limits our evaluation to wetlands (WET), open shrublands (OSH), grasslands (GRA), evergreen needleleaf forests (ENF), deciduous broadleaf forests (DBF), and croplands (CRO).
AMERI-FAR25 also provides data splits to test forecasting or temporal drift:
\textit{test\_future} and \textit{val\_future}, which consist of the last
year of data from every site with multiple years of data. In addition,
20\% of points were withheld at random from the remaining data to make a
final validation set (\textit{val}).

\subsection{FAR's Implementation}
\label{sec:far_dnn}
\begin{figure}
    \centering
    \includegraphics[width=0.95\linewidth]{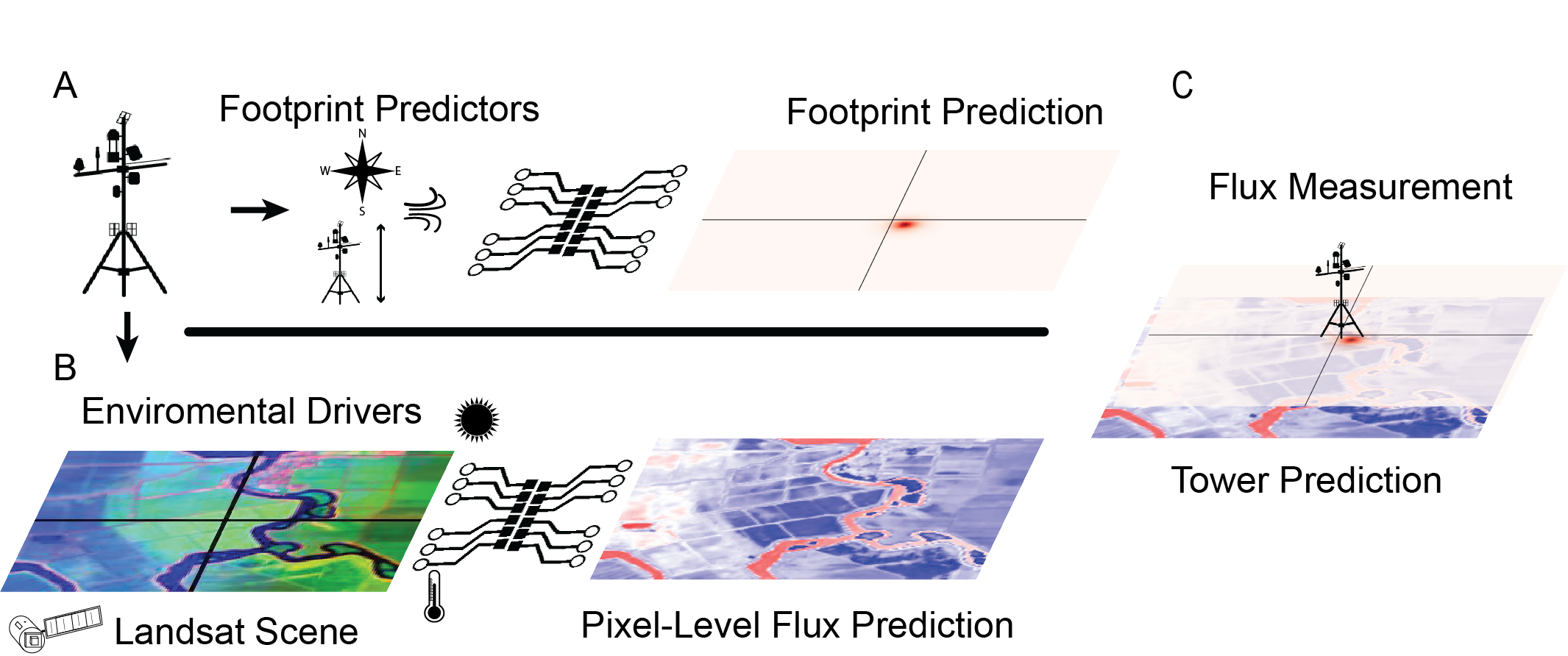}
    \caption{FAR Overview schematic. FAR is implemented as a two-arm deep neural network with a footprint prediction arm (A) and a flux prediction arm (B). These arms are then combined into a single aggregated prediction of the measured tower flux (C).
    }
    \label{fig:far_overview}
\end{figure}

FAR's implementation is depicted in Figure~\ref{fig:far_overview} and is divided into two parts that are
aggregated to predict EC tower measurements. First, the pixel-level flux
prediction model $( f^y_\theta(x^{flux}_{t,i})$) utilizes a grid of pixels from multi-band satellites
centered at each flux tower with length \(\mathbf{L}\), width
\(\mathbf{W}\)\textbf{,} and \textbf{D} bands. In this study, we utilize Landsat data denoted \(x_T^{landsat}\mathbf{\in \ }\mathbb{R}^{\mathbf{LxWxD}}\).
For brevity, we index all spatial coordinates by $i$. These satellite data can be thought of as a proxy for the environment, but must be augmented with data at higher temporal resolution to predict half-hour fluxes. To address this, we also add 'environmental' drivers, which are \textbf{N} local meteorological data points measured by
the tower (\(x_t^{drivers}\mathbf{\in \ }\mathbb{R}^{\mathbf{N}}\)), in particular SW\_IN, TA, and RH. Above, \(t\) denotes
the sampling of tower-measured variables (nominally every half-hour) and
\(T\) denotes the sampling of the Landsat data (nominally every 16 days per satellite). To align the different spatial and temporal resolutions of the inputs, each Landsat measurement is reused for every half-hour time step until the next Landsat measurement becomes available. Conversely, each half-hour tower measurement is converted to an image-sized array by assigning that same scalar value to every spatial location in the grid.
\begin{equation}
 x_{t,i}^{flux} \,\mathrel{:=}\, \Big(\,x_{\tau(t),i}^{landsat},\; x_t^{drivers}\,\Big), \qquad \tau(t) \,\mathrel{:=}\, \max\{\,T_k : T_k \le t\,\}.
\end{equation}
This model predicts \(\hat{y}\ \mathbf{\in \ }\mathbb{R}^{\mathbf{LxW}}\), a gridded pixel-level prediction of \coo flux. The flux-prediction model was implemented with a simple multi-layer
perceptron (MLP)~\cite{Rumelhart1986LearningErrors}. This MLP is applied independently to each input pixel.

Second, the footprint prediction model ($f^\alpha_{\phi,i}(x_t^{footprint})$) is inspired by attention mechanisms that are frequently used in deep learning methods from computer vision~\cite{dosovitskiy2021image}, natural language processing~\cite{vaswani2017attention}, and multi-instance learning~\cite{Ilse2018Attention-basedLearning}. These attention mechanisms allow for identifying regions of a model's input by providing a weight map (one weight per input instance), which is generally required to sum to one and is used to create a weighted mean of the input features. This is analogous to the problem presented above, where we wish to identify which pixels are contributing to the tower-aggregated measurement ($\hat{\alpha}_{t,i,\phi}$). Unlike the attention mechanisms used in computer vision, which use the input features themselves to learn weights, FAR's footprint prediction model utilizes \textbf{M}
measurements related to atmospheric conditions at the tower
(\(x_t^{wind}\mathbf{\in \ }\mathbb{R}^{\mathbf{M}}\)). These measurements are also held constant and spatially expanded to match the pixel resolution; additionally, the normalized pixel coordinates ($r_i=(w_i,l_i)\in(0,1)$) are concatenated to form the input.

\begin{equation}
 x_{t,i}^{footprint} \,\mathrel{:=}\, \Big(\,x_t^{wind},\; r_i\Big)
\end{equation}

with the predicted measurement footprint \(\hat{\alpha}\ \mathbf{\in \ }\mathbb{R}^{\mathbf{LxW}}\). The
footprint prediction model is also implemented with a simple MLP applied pixel-wise. Similarly to other attention-based models, the final output is constrained to sum to 1 by applying a softmax to the output of the MLP over the two spatial dimensions. $x_t^{wind}$ are the concatenation of WD, WS, USTAR, TA, and H. Both MLPs are implemented with 3 hidden layers each with 512 neurons. This architecture was chosen over other possibilities such as U-Net~\cite{Ronneberger2015U-Net:Segmentation} to preserve the independence of pixel predictions and maintain a simple realization of the algorithm. 

Both arms were trained simultaneously over each half-hour measurement in the training dataset to minimize the mean squared error loss described in Section~\ref{sec:far_theory}. In Section~\ref{sec:far_theory}, we showed that several footprints may yield the same aggregated flux, so to investigate the plausible range of the learned footprints, we included an entropy regularization term.
\begin{equation}
\mathcal L
= \sum_{t}
\left(
\left(Y_t^{obs} - \hat{Y}_t^{obs}\right)^2
- \lambda \sum_i \hat{\alpha}_{t,i,\phi}\,\ln\!\left(\hat{\alpha}_{t,i,\phi}\right)
\right).
\end{equation}

Where $-\lambda \sum_i \hat{\alpha}_{t,i,\phi}\,\ln\!\left(\hat{\alpha}_{t,i,\phi}\right)$ is
an entropy regularization term, with $\lambda > 0$ minimizing entropy
to prefer small footprints, \(\lambda < 0\) maximizing entropy to prefer
large footprints, and \(\lambda = 0\) applying no preference. FAR used a default value of $\lambda=10^{-3}$. After training, the flux prediction stage can then be
used independently to obtain pixel-level flux predictions.

The model was implemented in PyTorch~\cite{Paszke2019PyTorch:Library} and trained with the Adam optimizer
~\cite{Kingma2014Adam:Optimization} using a learning rate of $10^{-4}$.

Before training, input features and target fluxes were scaled by subtracting their mean value and dividing by their standard deviation as calculated from the training data to facilitate neural network convergence. As discussed in Section~\ref{sec:far_theory}, to limit any correlations between the footprint model and flux predictions due to prevailing wind directions, data augmentation was
applied to each Landsat scene by rotating it and the wind direction by
an arbitrary angle. FAR was trained until no further improvement was
observed in the loss function in any of the validation sets. The
\textit{val} dataset achieved the lowest loss value, followed by
\textit{val\_future} and \textit{val\_site}, which both achieved their
minimum loss earlier in the training process. The weights with the lowest
loss on the \textit{val\_site} dataset were then used for subsequent
evaluation, since it represents the best test of the upscaling task.
Training was performed on the University of Oregon's high-performance
computer, Talapas, and took about 7 days on an A100 in MIG mode
configured with 40 GB of GPU memory.

\begin{figure}
    \centering
    \includegraphics[width=0.75\linewidth]{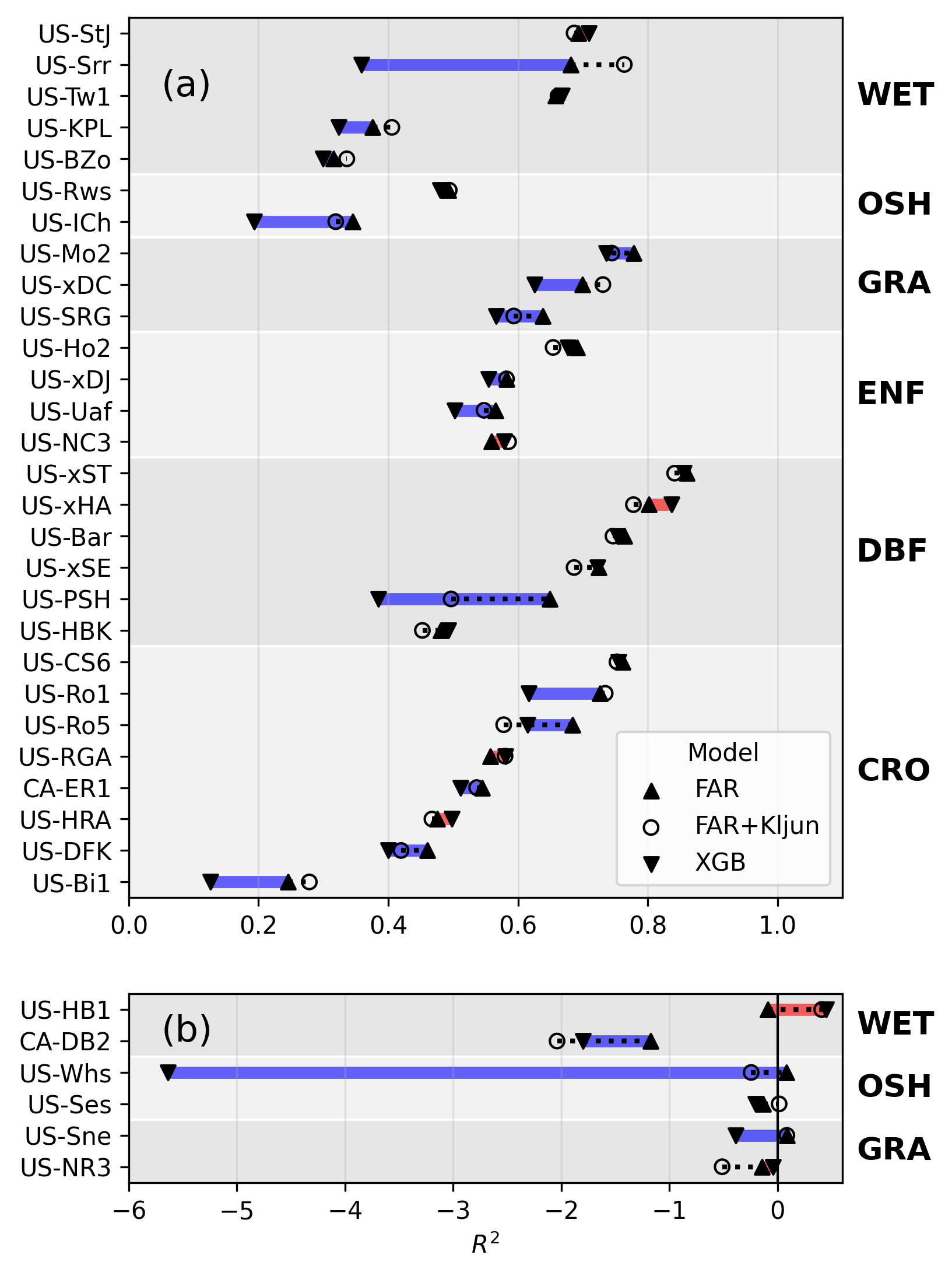}
    \caption{Comparison of $R^2$ scores by site for half-hour predictions. FAR, FAR$+$Kljun, and XGBoost (XGB) are shown for all sites in the \textit{test\_site} data split. Panel (a) shows all sites with $R^2 >0$ for all predictions. Panel (b) shows all sites for which at least one model has \(R^2 < 0\). Blue/red bars denote a positive/negative change in $R^2$ between XGBoost and FAR. The dashed line represents changes between FAR and FAR+Kljun. Bands denote the IGBP category of each site among wetlands (WET), Open Shrublands (OSH), Grasslands (GRA), Evergreen Needleleaf Forest (ENF), Deciduous Broadleaf Forest (DBF), and Croplands (CRO).}
    \label{fig:site_r2}
\end{figure}

\subsection{Evaluation}
\label{sec:eval}

To evaluate FAR, we assessed its performance in our \textit{test\_site}
split across a variety of sites and IGBP codes using root mean squared
error (RMSE) and coefficients of determination ($R^2$) calculated using the Nash-Sutcliffe efficiency construction~\cite{nash1970river}, which may drop below 0 if the regression prediction is a worse predictor than the mean of the distribution. FAR produces half-hour predictions, but to demonstrate how errors aggregate over time, we also combined measurements into monthly and yearly comparisons. Monthly and yearly values were obtained by taking the average of all predictions and measurements for each site and time period. Aggregated comparisons were performed without gap filling to avoid conflating different model errors, which means that not all months and years will have an equivalent number of measurements, and therefore the aggregates do not represent full \coo budgets. We evaluated FAR against several
models without footprint awareness, including XGBoost~\cite{Chen2016XGBoost:System}, which has been widely used for upscaling~\cite{Zhu2024UniformUFLUX}, FAR's MLP flux prediction branch alone without the footprint branch (FAR MLP), and a simple linear model. For these baselines, we took the average
of Landsat bands in a 5~$\times$~5-pixel (150~m~$\times$~150~m) uniform footprint and a larger
33~$\times$~33 (990~m $\times$ 990~m) pixel footprint and used the resulting values and environmental drivers as inputs to these models. We also evaluated $R^2$ for each site in the \textit{test\_site} split independently to assess how model predictions change across sites and IGBP codes.

To examine the importance of how the footprint model is learned, we compared performance at three $\lambda$ points and to a parametrized 2D footprint model using the footprint package
~\cite{Kljun2015AFFP} from Kljun et al. To calculate footprints, we utilized additional
variables, the Monin-Obukhov length and the standard deviation of
lateral wind velocity fluctuations. For sites where these variables were
not available, they were predicted from the same variables used in FAR's
footprint prediction arm. These values were then input into the
footprint package for every half-hour measurement in the AMERI-FAR25 dataset with an assumed boundary layer height of 1~km. These calculated footprints were then used as a drop-in replacement for FAR's footprint prediction branch ($f^\alpha_{\phi,i}(x_t^{footprint})$), allowing the flux prediction arm to train using existing footprints. We refer to this model as FAR+Kljun. The area of the calculated footprints was also compared against the learned footprints. For comparisons, the area for each footprint was calculated by combining the pixels with the largest weights sequentially until their sum exceeded a 0.95 threshold.

A necessary condition for FAR to learn unbiased pixel-level predictions is that some training footprints cover relatively homogeneous regions. To evaluate whether this condition is met, we quantified variation within each footprint using the input Landsat data. Each Landsat band was first standardized across the full dataset to have a mean of zero and a standard deviation of one. As above, let $x^{flux}_{t,i}\in\mathbb R^{B}$ denote the resulting standardized Landsat band vector at pixel $i$ (with $B$ bands), and let $\hat{\alpha}_{t,i,\phi}$ denote the footprint weight for pixel $i$. We compute the footprint-weighted mean vector
\begin{equation}
\bar{x}^{flux}_t = \sum_i \hat{\alpha}_{t,i,\phi}\,x^{flux}_{t,i},
\end{equation}
and the squared deviation at each pixel
\begin{equation}
 d^2_{t,i} = \lVert x^{flux}_{t,i}-\bar{x}^{flux}_t\rVert_2^2 = \sum_{b=1}^{B}\left(x^{flux}_{t,i,b}-\bar{x}^{flux}_{t,b}\right)^2.
\end{equation}
We then define within-footprint variation as a footprint-weighted root-mean-square deviation, adjusted for the effective number of pixels,
\begin{equation}
\sigma^{fp}_t
=
\left(
\frac{\sum_i \hat{\alpha}_{t,i,\phi}\, d^2_{t,i}}{B\left(1-\sum_i \hat{\alpha}_{t,i,\phi}^2\right)}
\right)^{\!1/2}
\end{equation}.
Values near 0 indicate that the pixels covered by a footprint are homogeneous, whereas values near 1 indicate within-footprint variation comparable to the dataset-wide variation of the Landsat bands.

To assess our ability to upscale, we replaced EC
tower measurements with estimates derived from PRISM to test how well meteorological products can replace exact tower data. We then applied our
model to a geographic area of Oregon as a demonstration of FAR's
utility. Notably, some AMERI-FAR25 sites are located in South America and not within PRISM's coverage, so comparisons were made only for sites with PRISM data.
Since these data are complete and we do not need to accommodate missing EC tower measurements, we utilize these predictions for complete NEP estimates. In this region, we also compare NEP predictions with four existing products that provide downloadable predictions: FLUXCOM-X~\cite{bg-21-5079-2024,fluxcomdataset}, GEO3~\cite{han2026geo3nep}, GLO~\cite{yuan2025gloflux,yuan2025glofluxpaper}, and SMAP~\cite{kimball2025smap}.
Like FAR, FLUXCOM-X, GEO3, and GLO use EC tower data to train their models. FLUXCOM-X and GLO are machine-learning-based, whereas GEO3 uses similarity metrics between areas. SMAP is a process model based on satellite data.
FLUXCOM-X provides hourly data at $0.05\deg$ resolution. SMAP provides daily predictions at 9~km resolution. GLO provides monthly predictions at $0.1\deg$ resolution, and GEO3 provides monthly predictions at $0.05\deg$ resolution. To compare the products with FAR, we use January 1, 2016, through December 31, 2021, the range of complete years available for all products. For each product, we use all cells whose centroids fall within the region. In this region, we also estimate the effect of aggregating inputs before prediction (see Section~\ref{sec:resolution_mismatch}). We explicitly compare the mean of the high-resolution predictions within each 1~km cell, $\frac{1}{N}\sum_i f^y_{\theta^*}(x_i)$, with the prediction based on the 1~km aggregated inputs, $f^y_{\theta^*}(\frac{1}{N}\sum_i x_i)$. 
Finally, we also investigated feature importance to assess the model's
reliance on known features, or lack thereof, using Shapley Additive
Explanations (SHAP)~\cite{Lundberg2017APredictions}. SHAP values attribute each
feature's contribution, positive or negative, to a given FC prediction. Specifically, we utilized the SHAP package's KernelExplainer, which is designed for generic ``black-box'' attributions, and applied it to FAR's flux prediction arm to understand feature contributions to predictions.

\begin{table}
\centering
\begin{tabular}{lrrrrrrrrr}
\hline
Model & HH & HH & HH & Monthly & Monthly & Monthly & Yearly & Yearly & Yearly \\

& $R^2$ & RMSE & Bias & $R^2$ & RMSE & Bias & $R^2$ & RMSE & Bias \\

\hline
\multicolumn{7}{c}{\textbf{Footprint-Aware}}\\
\hline
FAR (default;$\lambda=10^{-3}$) & 0.631 & 3.37 & \textbf{0.02} & \textbf{0.765} & \textbf{1.97} & \textbf{0.02} & 0.817 & 1.17 & \textbf{-0.03} \\
FAR+Kljun & \textbf{0.634} & \textbf{3.37} & 0.16 & 0.751 & 2.03 & 0.18 & \textbf{0.822} & \textbf{1.16} & 0.11 \\
FAR $\lambda=0$ & 0.624 & 3.41 & -0.34 & 0.751 & 2.03 & -0.35 & 0.783 & 1.28 & -0.40 \\
FAR $\lambda=-10^{-3}$ & 0.598 & 3.53 & 0.12 & 0.719 & 2.15 & 0.06 & 0.739 & 1.40 & 0.07 \\
\hline
\multicolumn{7}{c}{\textbf{150 m Baselines}}\\
\hline
XGB & \textbf{0.575} & \textbf{3.62} & 0.12 & \textbf{0.731} & \textbf{2.11} & 0.12 & \textbf{0.772} & \textbf{1.31} & \textbf{0.07} \\
FAR MLP & 0.575 & 3.62 & 0.15 & 0.719 & 2.15 & 0.18 & 0.733 & 1.42 & 0.14 \\
Linear & 0.425 & 4.21 & \textbf{-0.01} & 0.686 & 2.27 & \textbf{-0.02} & 0.756 & 1.35 & -0.09 \\
\hline
\multicolumn{7}{c}{\textbf{990 m Baselines}}\\
\hline
XGB & \textbf{0.570} & \textbf{3.64} & \textbf{0.13} & \textbf{0.675} & \textbf{2.32} & 0.17 & 0.689 & 1.53 & 0.14 \\
FAR MLP & 0.472 & 4.04 & 0.30 & 0.592 & 2.60 & 0.29 & 0.568 & 1.80 & 0.24 \\
Linear & 0.414 & 4.25 & 0.14 & 0.654 & 2.39 & \textbf{0.13} & \textbf{0.698} & \textbf{1.50} & \textbf{0.06} \\
\hline
\multicolumn{7}{c}{\textbf{Upscaling}}\\
\hline
FAR (PRISM sites only) & \textbf{0.639} & \textbf{3.54} & \textbf{0.00} & \textbf{0.769} & \textbf{2.08} & \textbf{0.02} & \textbf{0.813} & \textbf{1.27} & \textbf{-0.04} \\
FAR+PRISM & 0.603 & 3.71 & 0.25 & 0.763 & 2.11 & 0.25 & 0.805 & 1.30 & 0.20 \\
\hline
\end{tabular}
\caption{Model performance across experiments. The $R^2$, root mean squared error (RMSE), and bias on all data from withheld sites (\textit{test\_site}) are shown. Models include XGBoost (XGB), FAR with Kljun footprints or different \(\lambda\) values, the FAR flux-prediction arm without a footprint model (FAR MLP), and linear models. Upscaling results describe FAR's performance on sites for which PRISM data is available and FAR's performance replacing tower data with PRISM data (FAR+PRISM). Results are shown for half-hour (HH) predictions, which are then averaged into monthly and yearly predictions. RMSE is reported in \fluxunit. The best-performing result in each group is bolded.}
\label{tab:model_performance}
\end{table}

\section{Results}
For data from withheld sites (\textit{test\_site}; Table~\ref{tab:model_performance}), footprint-aware models generally outperform non-footprint-aware models at all time scales, with FAR+Kljun having slightly better performance than FAR for half-hour predictions and yearly averages, and FAR performing best for monthly averages. Scatter plots comparing FAR's predictions and tower measurements can be seen in supplementary Figure~\ref{fig:supp_scatter}. For fixed footprints, XGBoost performs best at all time scales for both the 150~m and 990~m baselines, with the notable exception of yearly averages at 990~m, where the linear model performs best. As noted in Section~\ref{sec:resolution_mismatch}, linear models are more robust to bias resulting from changes in scale. Since these averages are sensitive to any remaining bias in the half-hour predictor, this may explain why the linear model performs better at larger baselines and when averaging more measurements. While all biases are relatively small compared with the RMSE, linear models generally perform similarly to or better than other fixed-footprint models, with FAR performing best among the footprint-aware models. The 150~m baselines generally outperform the 990~m baselines, particularly for monthly and yearly averages. Taken together, this implies that using a smaller fixed footprint near the tower improves results over a larger one, particularly when aggregating measurements into longer time periods. However, neither baseline outperforms footprint-aware models, with substantial gains in comparison to FAR's MLP branch.

\begin{figure}
    \centering
    \includegraphics[width=0.7\linewidth]{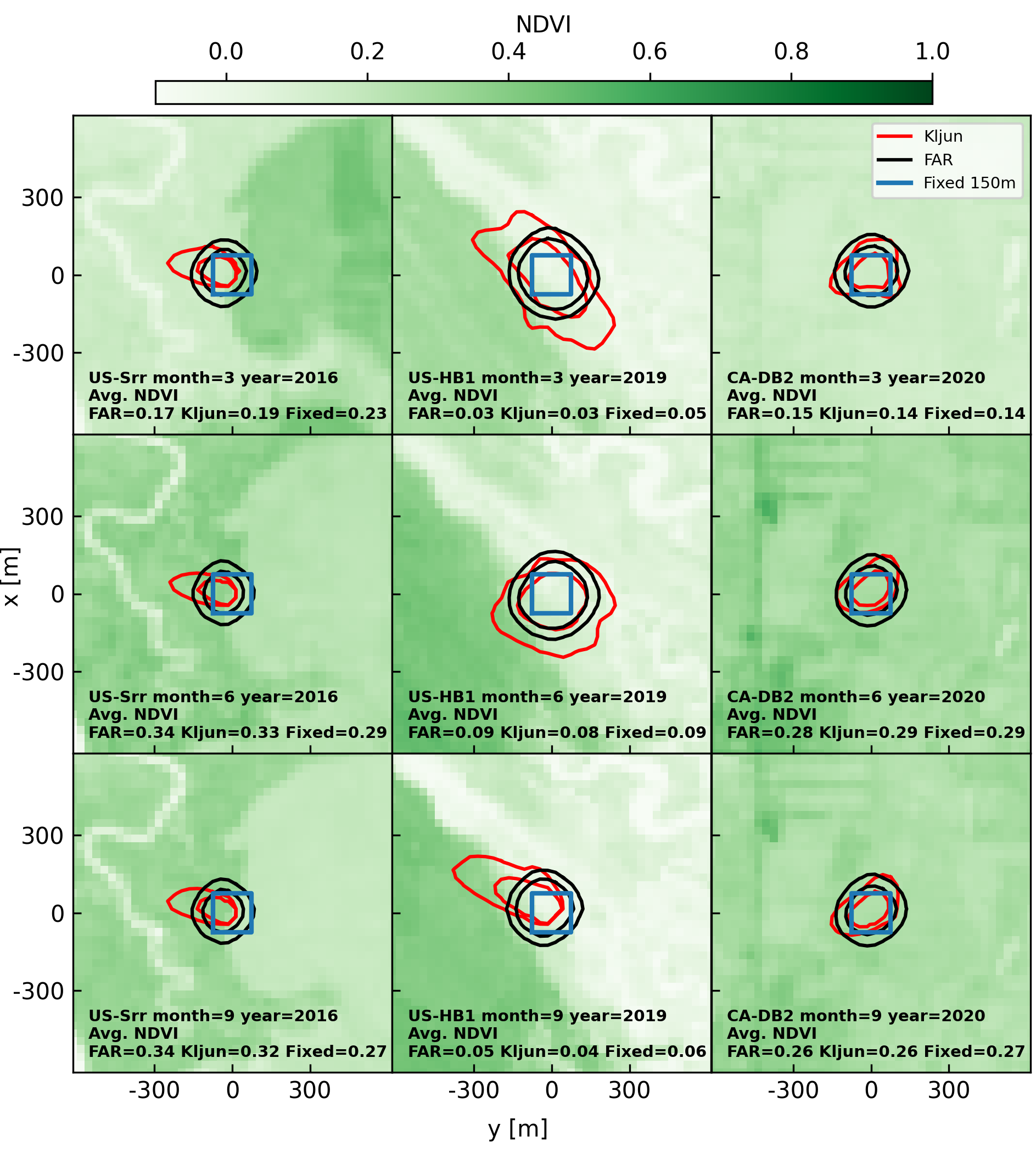}
    \caption{The 68\% and 95\% contours for the monthly footprint climatology (the average of all half-hour footprints) from Kljun, FAR, and a fixed 150~m reference. Given that each model uses multiple features, NDVI is shown as an illustrative proxy.  The background in each panel represents monthly average NDVI. 
    The left column shows site US-Srr, the center column shows site US-HB1, and the right column shows site CA-DB2. Rows represent March, June, and September. The years selected are the most temporally proximal years of data.}
    \label{fig:site_fp}
\end{figure}

FAR generally matches or improves performance relative to XGBoost for each site, with the notable exception of US-HB1, where FAR shows a substantial drop in performance and FAR+Kljun shows comparable performance (Figure~\ref{fig:site_r2}). Similarly, in CA-DB2, none of the three models predict with $R^2>0$, but FAR+Kljun significantly underperforms FAR. US-Srr, US-HB1, and CA-DB2 show contrasting responses to footprint-aware modeling. In Figure~\ref{fig:site_fp}, US-Srr exhibits a relatively consistent footprint that is off-center, with variation in footprint-weighted NDVI between the fixed-footprint and FAR models, which may explain the large performance gains. US-HB1 and CA-DB2, however, show more consistency among the footprint models, which we would expect to limit potential gains from using FAR and FAR+Kljun. While FAR and FAR+Kljun can address the spatial variation in US-Srr, all models may lack sufficient information to learn relationships that can accurately predict both CA-DB2 and US-HB1, with FAR's better performance on CA-DB2 coming with a tradeoff of worse performance on US-HB1, and vice versa for FAR+Kljun. While this is a potential explanation, additional work is needed to verify it.

SHAP results indicate that FAR relies primarily on physically interpretable environmental and spectral features commonly associated with carbon flux, vegetation, and moisture dynamics. The seven most important features (those with the largest mean
absolute SHAP value) are SW\_IN, Red, Near Infrared, SWIR 2, Green, SWIR 1, and Blue. This intuitive result corresponds with solar
radiation being a key driver of carbon flux, followed by frequently used
proxies for vegetation and soil moisture. For example, the normalized
difference vegetation index (NDVI) is calculated from the Red and Near
Infrared channels~\cite{Tucker1979RedVegetation,Rouse1974MonitoringERTS}, and its extensions such as the enhanced
vegetation index (EVI) incorporate the blue channel~\cite{Huete2002OverviewIndices} and GNDVI, which incorporates the green channel~\cite{Smith1995ForecastingSatellite}. Similarly, SWIR 1 and SWIR 2 are used in the normalized difference water index (NDWI)~\cite{Gao1996NDWIASpace}, and normalized burn ratio (NBR)~\cite{Key1999TheSeverity.}, respectively.

\begin{figure}
    \centering
    \includegraphics[width=0.5\linewidth]{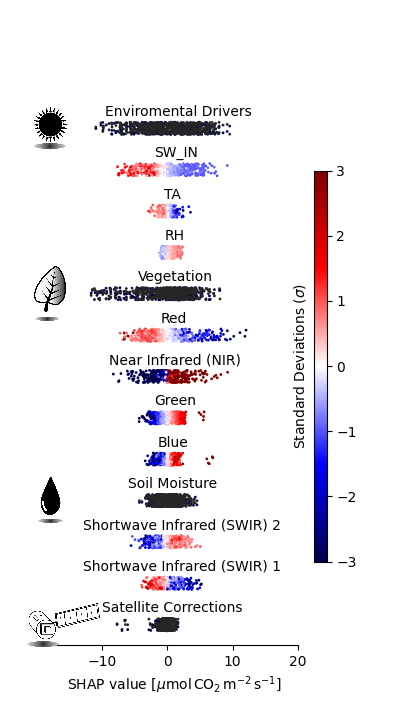}
    \caption{SHAP values for the most important flux prediction features.
    SHAP values are grouped into 4 categories: 1. Environmental drivers
    (SW\_IN, TA, RH), 2. Vegetation proxies (Red, NIR, Green, Blue), 3. Soil
    moisture (SWIR 1/2, TIRS 1/2, Coastal aerosol) and 4. Satellite
    corrections (SAA, SZA, VAA, VZA, TIME\_MASK, Cirrus). Below each group
    is a selection of the highest-impact features. These are shown with
    red/blue points denoting higher/lower than average values. Red points on
    the right of the figure denote that a high value of that feature is associated with flux to the atmosphere, and blue points to the left denote that a low value of that feature is associated with ecosystem drawdown.
    }
    \label{fig:shap}
\end{figure}

\begin{figure}
    \centering
    \subfigure[]{\includegraphics[width=0.49\linewidth]{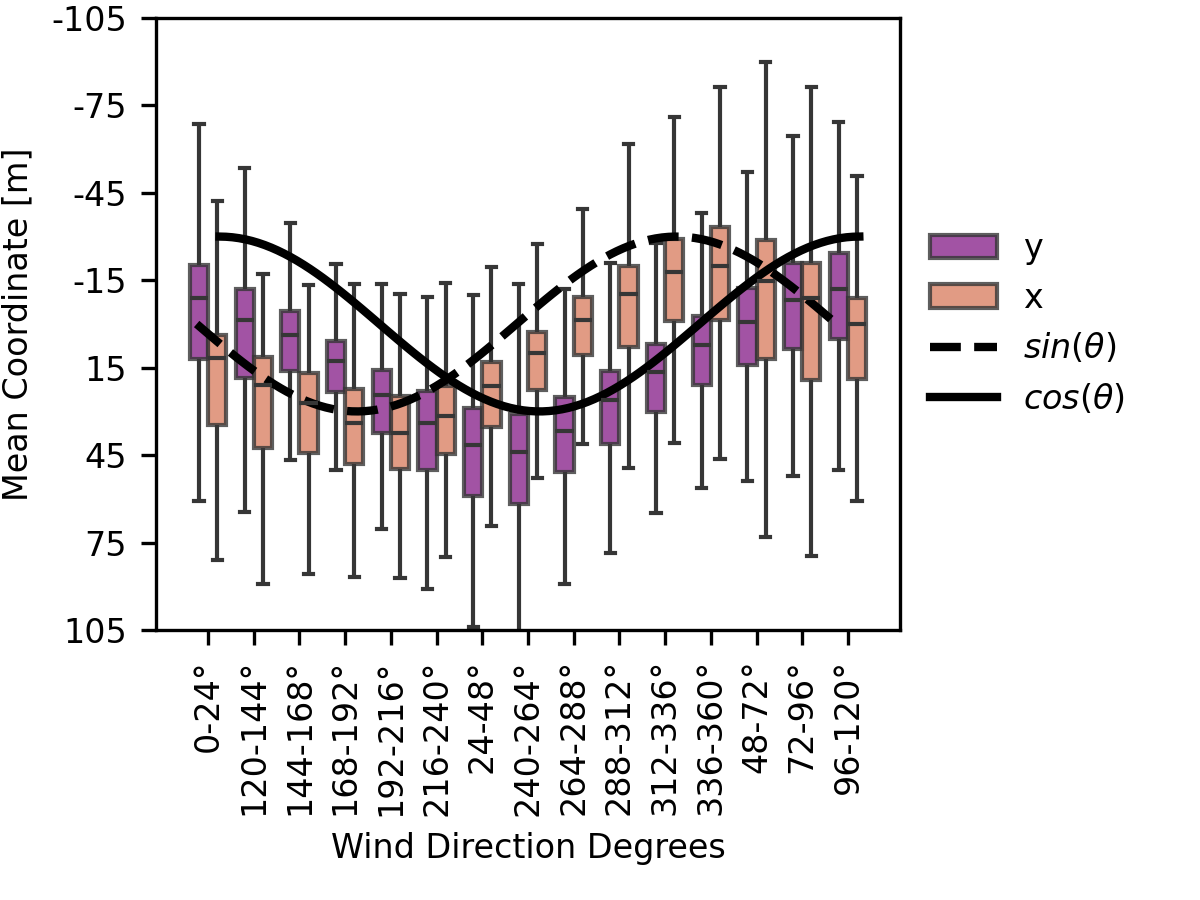}}
    \subfigure[]{\includegraphics[width=0.49\linewidth]{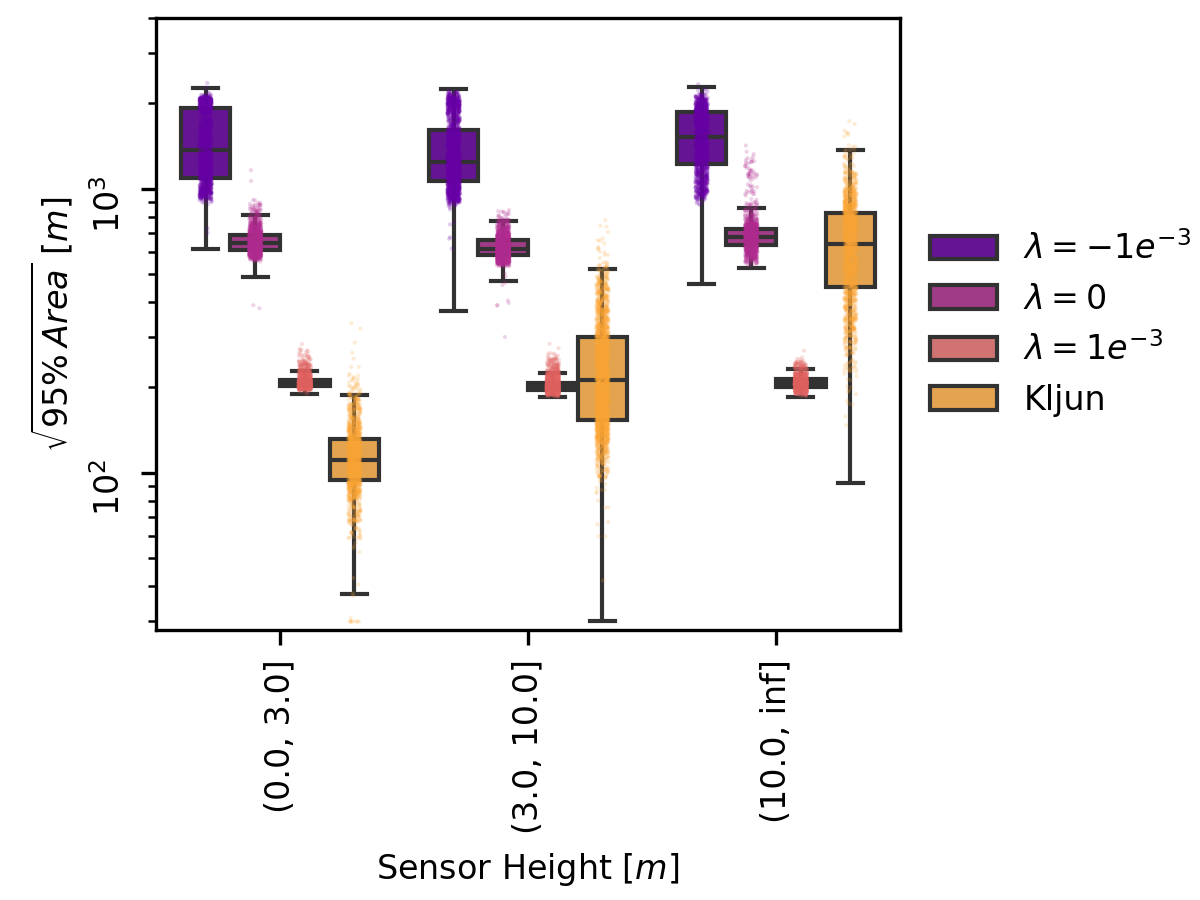}}
    \caption{Footprint Characteristics. (a) The footprint-weighted mean of the coordinates versus wind direction. For context, the sine and cosine of the wind direction scaled by 30~m are overlaid to show the expected upwind pattern.  (b) The square root of the area containing 95\% of the footprint obtained from different models for three tower height groups.}
    \label{fig:wind}
\end{figure}

FAR's learned footprint properties depend strongly on the value \(\lambda\)
used during training to regularize the footprints. Across all
\(\lambda\) values, footprints learn the tower's location at the center of
the image, but their respective areas can
vary substantially (Figure~\ref{fig:wind}) and yield comparable $R^2$ on the test
data (Table~\ref{tab:model_performance}). This implies that FAR's footprints are only weakly constrained, but have a few notable patterns. Figure~\ref{fig:wind} shows that FAR's footprints are generally upwind, and that the area of each learned footprint varies strongly with $\lambda$; it does not vary strongly with tower height when compared to footprints used by FAR+Kljun. Notably, while the learned footprints are only weakly constrained, FAR's default ($\lambda=10^{-3}$), which prefers small footprints, still learns footprints that are larger than those used by FAR+Kljun for towers less than 3~m tall. This result is consistent with tracer studies, which found that the Kljun et al. parameterization in these cases may yield footprints that are smaller than those measured~\cite{Rey-Sanchez2022DetectingMaps}. However, it is important to note that FAR does not guarantee that footprints are physically consistent, but rather only consistent with the observed flux and spatial heterogeneity. For example, the learned areas do not vary meaningfully with tower height, wind speed, or friction velocity.

\begin{figure}
    \centering
    \includegraphics[width=0.47\linewidth]{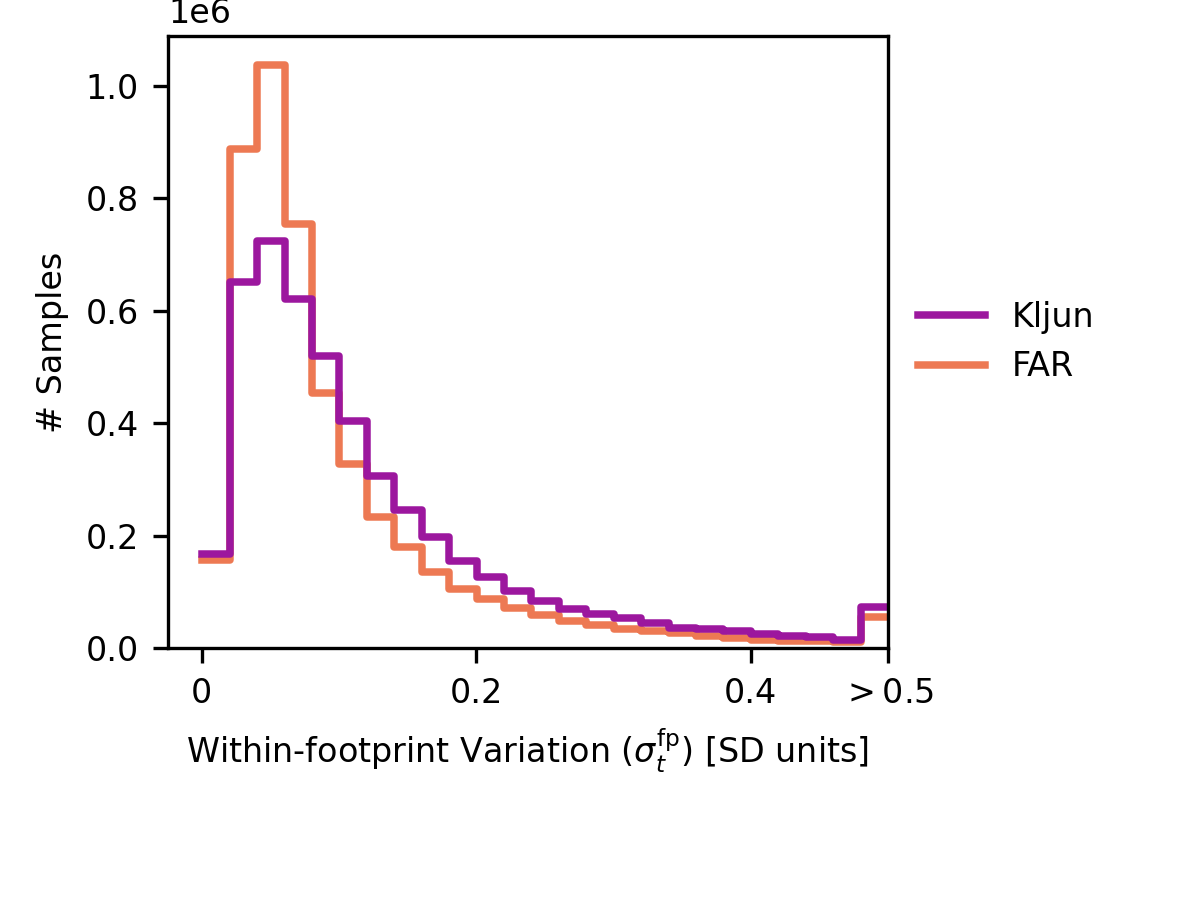}
    \caption{Within-footprint variation of satellite bands, where values near 0 indicate homogeneous contributing regions and values near 1 indicate variation comparable to the full dataset.}
    \label{fig:band}
\end{figure}

As noted in Section~\ref{sec:far_theory}, homogeneous regions are valuable for learning unbiased pixel-level predictors, but cannot constrain footprints within the homogeneous region. The within-footprint variation is shown in Figure~\ref{fig:band}. The data peak at the smallest bin near zero, implying that homogeneous contributing regions exist and make up a considerable part of the AMERI-FAR25 dataset, with footprints from FAR+Kljun exhibiting slightly larger variation. Very few data points exhibit substantial heterogeneity within the footprint for either the learned or calculated footprint, which is consistent with weak constraints on the footprint model and sufficient data to anchor pixel-level predictions.

\subsection{Upscaling Example}

\begin{figure}[pos=!htbp]
    \centering
    \includegraphics[width=0.9\linewidth]{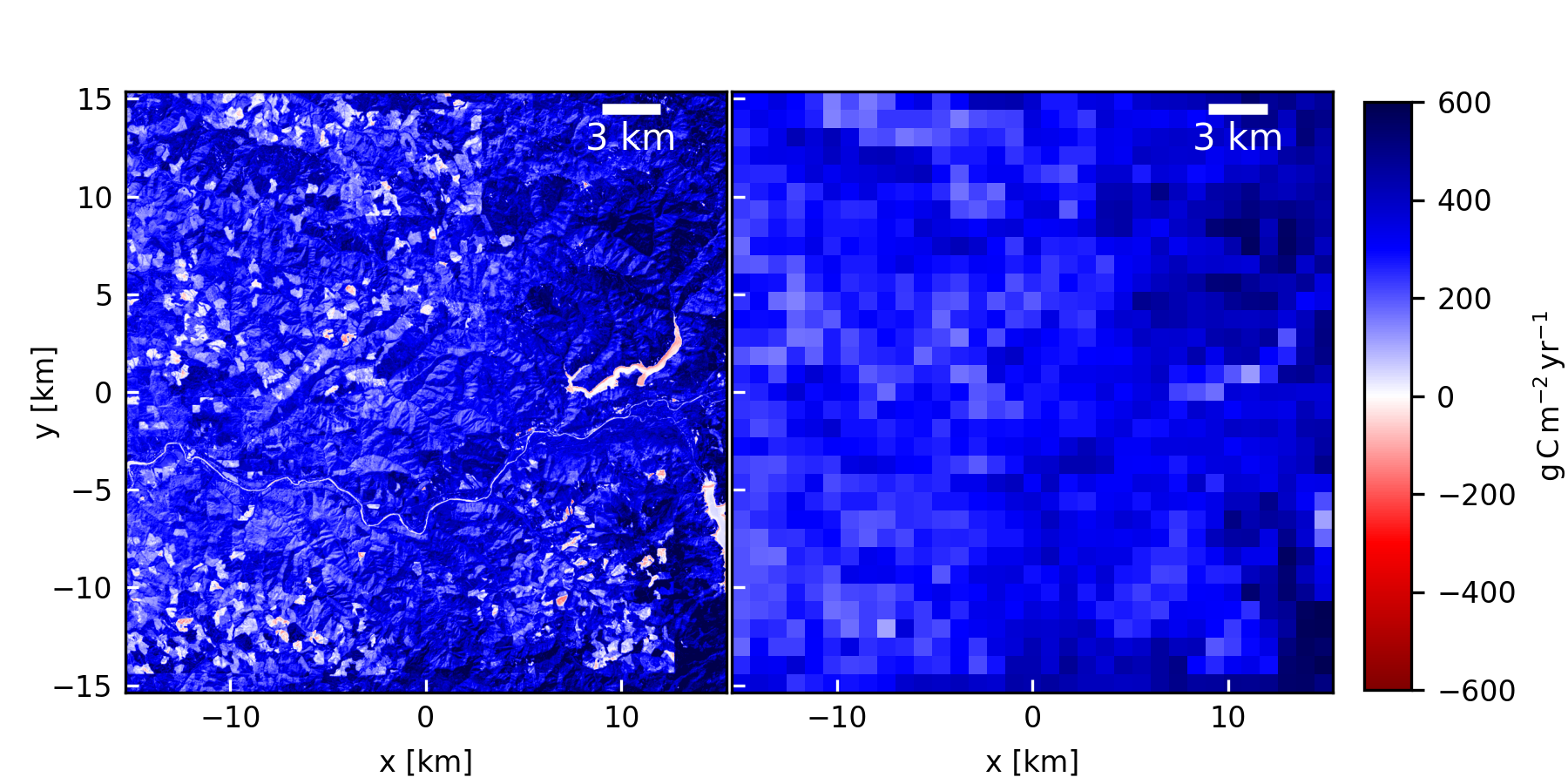}
    \caption{10-Year (2013--2022) average flux predicted by FAR for a region of Oregon's Western Cascades. The left panel shows FAR's 30~m resolution and the right panel shows FAR's results averaged to 990~m. Blue represents net \coo drawdown and red represents net \coo sources.}
    \label{fig:upscaling}
\end{figure}

\begin{figure}[pos=!htbp]
    \centering
    \includegraphics[width=0.5\linewidth]{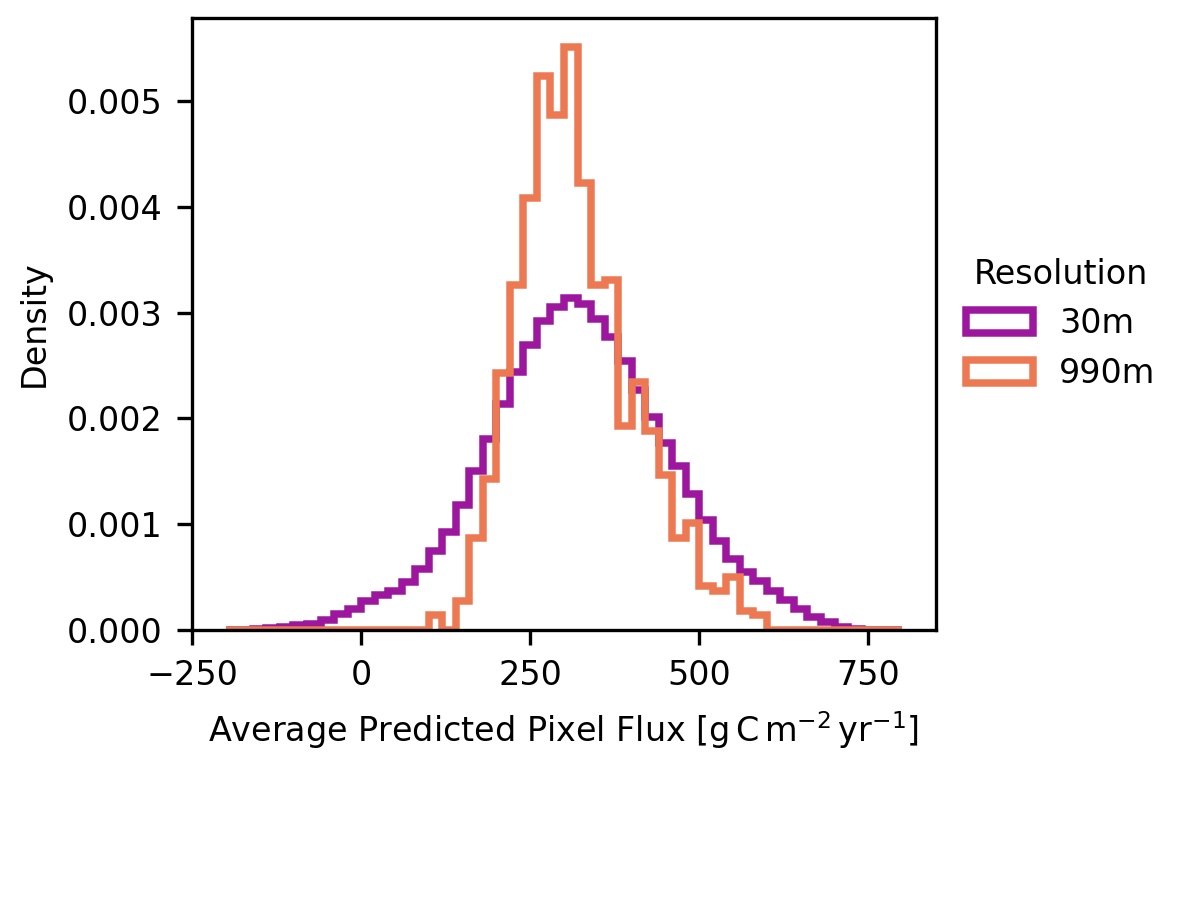}
    \caption{Histogram of the average 10-year predicted fluxes shown in Figure~\ref{fig:upscaling} at both high (30~m) and low (990~m) resolutions. Both histograms are density-normalized such that the total area under each histogram equals one.}
    \label{fig:fluxhist}
\end{figure}

Replacing meteorological variables measured by each tower with values
estimated from PRISM during inference results in a decrease in predictive performance for half-hourly predictions, but provides nearly identical
predictive power on test sites for monthly estimates (see Table~\ref{tab:model_performance}). We attribute this to our use of PRISM's monthly solar transmittance and daily resolution for other environmental drivers. 
However, the good performance for monthly data allows for upscaling. As a demonstration, we utilize PRISM and Landsat data to predict half-hourly fluxes for a $1024\times1024$ pixel region (30.7~km $\times$ 30.7~km) of the Western Cascades centered at 44.172$^\circ$~N, 122.421$^\circ$~W for a 10-year period covering 2013--2022. 
Since our goal for upscaling is just pixel-level flux predictions, we only utilize FAR's flux arm. We selected a small region over a global map due to limitations resulting from FAR's substantial computational requirements. For example, this region alone requires a $1024\times1024$ prediction map be generated for every half hour across the 10-year period exceeding 170,000 prediction maps.
However, this region has a diverse set of forests with varying stand age and management. Previous modeling work in this region has recommended spatial resolutions below 250~m to model this heterogeneity~\cite{Turner2003EffectsLandscape}. Figure~\ref{fig:upscaling} compares FAR’s predictions at native 30~m resolution to results with $\sim$1~km (990~m), a resolution more commonly available in other upscaling products. The high-resolution predictions identify localized \coo sources consistent with recent timber harvesting, which are lost in many of the coarser pixels. Process modeling at similar resolution in this area estimates that NEP spans a range of values, with peaks around 350 \yearlyfluxunit and 50 \yearlyfluxunit with an average of 189 \yearlyfluxunit, with the differences attributed to stand age~\cite{Law2004DisturbanceUSA} and climate. A similar modeling study found NEP per pixel to vary between approximately -200 to 500 \yearlyfluxunit with a mean of 179.5 \yearlyfluxunit~\cite{Turner2006ANEP}.
For the region shown in Figure~\ref{fig:upscaling}, FAR predicts a mean flux of 315 \yearlyfluxunit with the 5th and 95th percentiles at 91.8 \yearlyfluxunit and 540 \yearlyfluxunit (see Figure~\ref{fig:fluxhist}), which generally coincides with the greater NEP peak predicted by~\cite{Law2004DisturbanceUSA} and the NEP per pixel range predicted by~\cite{Turner2006ANEP}. Other flux products produce a wide range of estimates, from 3.2~to~767.3~\yearlyfluxunit (Table~\ref{tab:comp}). GLO, FLUXCOM-X, and GEO3 predict higher values, with GEO3 being the closest to FAR. SMAP produces the smallest value. Aggregating inputs to 1~km before prediction produced an estimate 35.7~\yearlyfluxunit (10.7\%) higher than averaging the high-resolution predictions. This difference is the estimated bias due to Jensen's inequality. The higher estimate slightly reduced the discrepancy between FAR and FLUXCOM-X, GEO3, and GLO but did not explain the full differences among products. However, despite the disagreement in magnitude, seasonal trends are correlated, with linear fits having an $R^2$ between 0.642 and 0.765 and slopes between 0.624 and 1.676. Spatial distributions are also correlated, with the notable anti-correlation between FAR and GEO3. 

Although differences in study areas and time periods preclude exact comparisons, FAR's high-resolution flux estimates are closer to estimates from regional process models than to estimates from other products. As also shown in Figure~\ref{fig:fluxhist}, coarser resolution reduces the tails of the predicted flux distribution, and therefore may hide opportunities for fine-scale interventions. However, substantial differences exist in estimates of NEP for this region across products. The size of this spread is substantially larger than our estimate of bias due to scale mismatch.

\begin{table}
\centering
\caption{Mean NEP from FAR and other products for the region shown in Figure~\ref{fig:upscaling} from January 1, 2016, through December 31, 2021. Mean NEP is averaged over time and across the region. Temporal comparisons are based on linear fits of monthly, regionally averaged NEP between FAR and each specified product. The spatial comparison is based on a linear fit between NEP averaged over time for each product grid cell and FAR predictions averaged over time and over the pixels corresponding to that grid cell. For both comparisons, \(R^2\), bias, and slope quantify agreement with FAR. NEP and bias are reported in \yearlyfluxunit. The FAR 990~m aggregate \(\rightarrow\) predict row quantifies the bias due to Jensen's inequality by averaging FAR's \(X^{landsat}\) inputs to 990~m resolution before prediction.}
\begin{tabular}{lrrrrrrr}
\hline
Product &
Mean NEP &
\multicolumn{3}{c}{Temporal comparison} &
\multicolumn{3}{c}{Spatial comparison} \\
\cline{3-5}\cline{6-8}
& & $R^2$ & Bias & Slope & $R^2$ & Bias & Slope \\
\hline
FAR & 334.5 & - & - & - & - & - & - \\
FLUXCOM-X & 529.6 & 0.681 & 195.3 & 1.092 & 0.464 & 202.1 & 1.019 \\
SMAP & 3.2 & 0.642 & -331.6 & 0.624 & 0.671 & -323.4 & 0.327 \\
GloFlux & 767.3 & 0.765 & 430.3 & 1.676 & 0.254 & 455.7 & 0.770 \\
GEO3-NEP & 420.3 & 0.744 & 84.7 & 0.903 & 0.481 & 92.5 & -0.771 \\
FAR 990 m aggregate \(\rightarrow\) predict
& 370.3 & 0.989 & 35.7 & 1.114 & 0.932 & 35.3 & 0.953 \\
\hline
\end{tabular}
\label{tab:comp}
\end{table}

\section{Conclusions}

FAR presents a principled approach to high-resolution upscaling that can address footprint bias and bias arising from scale mismatch. FAR can provide unbiased pixel-level predictions at spatial scales smaller than an EC tower's footprint and can improve agreement between footprint-aggregated predictions and tower measurements relative to models that lack a footprint model. FAR's learned footprint model performed comparably to the model using footprints calculated with the \textit{footprint} package~\cite{Kljun2015AFFP}, although the learned and calculated footprints differed substantially. Learned footprints aligned with the tower's location and tended to shift upwind, but otherwise varied substantially in area with relatively little effect on prediction quality. This result supports our theoretical conclusion that footprints can be poorly constrained when spatial heterogeneity is limited (probably because investigators tend to place EC towers in homogeneous areas) but nevertheless have relatively little effect on the quality of pixel-level predictions. However, despite being only weakly constrained, all footprint models improved performance relative to fixed-footprint baselines.

In the upscaling demonstration, FAR yielded estimates closest to those from existing process-modeling studies in the region. However, this difference is unlikely to result solely from spatial resolution because the bias due to Jensen's inequality was only about 10\%. One important difference between FAR and its comparators is the dataset used to train or build predictions. FAR uses sites concentrated in North America, whereas the other products use global datasets. We hypothesize that it may be difficult to distinguish the studied forests from other high-productivity areas globally, but substantial work remains to develop a consensus estimate for this region. A downside of FAR is the high computational cost of using high-resolution satellite imagery, both in terms of data requirements and model complexity. We envision FAR as useful for targeted applications that require high-resolution flux maps, but a global FAR product would currently require substantially greater computational resources.

\section{Discussion}

FAR is well suited for broader application in current and future efforts to harmonize EC tower data globally. However, there are several limitations of FAR. The data used to train FAR are concentrated in North America, and similarly, the use of PRISM for upscaling is currently limited to the United States. Another important assumption of FAR is that the
input environmental drivers are sufficient to predict an ecosystem's response. This assumption is unlikely to hold in urban or intensively managed areas, which contain additional \coo sources and may require additional data to predict ecosystem fluxes. FAR's implementation is a simple realization of this concept, and additional model developments for each branch could improve performance. For example, vision transformers~\cite{dosovitskiy2021image} could improve flux predictions, and physics-informed models~\cite{RAISSI2019686} could improve footprint predictions. Similarly, new geospatial foundation models or data from different platforms, including lidar, radar, and higher-resolution satellite imagery, could provide better inputs than Landsat alone. Despite good accuracy in a variety of ecosystems, there remains progress to be made in predicting some sites and determining their reasons for failure. For example, wetlands US-HB1 and CA-DB2 show variation in prediction quality across FAR, FAR+Kljun, and XGBoost, with no tested model achieving positive $R^2$ for site CA-DB2. Additional environmental drivers, site features, or historical information may be required to capture the observed differences at many wetland sites.

The poor constraint on FAR's learned footprints and the improvement from the 990~m to the 150~m baselines imply that appropriate spatial averaging before training may yield comparable predictive performance without FAR's two-armed approach. This approach could substantially reduce the computational burden but would remain subject to bias resulting from Jensen's inequality when aggregating before predicting. Bias from this inequality requires spatial heterogeneity in the inputs during training or inference. Spatial heterogeneity appears limited in the training data, whereas in our test region, aggregating inputs before prediction produced an estimate about 10\% higher than averaging the high-resolution predictions. This 10\% increase in the NEP estimate, however, remains small relative to the spread among predictions from existing models in our region.

Overall, our findings demonstrate that FAR can start to address a long-standing challenge of footprint bias in spatiotemporal predictions and translate EC data into high-resolution maps of carbon fluxes. By integrating \siteyears site-years of
flux tower measurements with Landsat imagery, FAR not only achieves high predictive ability across diverse ecosystems but also provides a framework for estimating ecosystem sinks and sources of \coo at scales directly relevant to land management and policy. While additional validation is needed, comparisons with existing process-based models are encouraging. Importantly, this opens the door to calculating carbon budgets across heterogeneous landscapes, offering an avenue to reconcile ecosystem processes with local and regional assessments of climate mitigation potential. However, the substantial variation in the magnitude of existing NEP estimates should be resolved before these estimates are trusted.

\FloatBarrier

\begin{longtable}{lllccc}
\caption{Sites used in AMERI-FAR25. The complete list of
AmeriFlux sites used in each dataset split in AMERI-FAR25. Sites are
exclusively divided between \textit{train}, \textit{val\_site} and
\textit{test\_site}, with sites in the \textit{train} dataset also appearing in the
\textit{val}, \textit{val\_future}, and \textit{test\_future} datasets.\label{site-table}}\\
\toprule
Site ID & IGBP & Date Range & Train site-days & Test site-days (site/future) & Val site-days (site/future/val) \\
\midrule
AR-CCg \cite{AR-CCg} & GRA & 2018-2021 & 253 & 0/0 & 0/160/61 \\
AR-TF1 \cite{AR-TF1} & WET & 2016-2018 & 75 & 0/0 & 0/102/6 \\
AR-TF2 \cite{AR-TF2} & WET & 2017-2017 & 0 & 0/0 & 28/0/0 \\
BR-CST \cite{BR-CST} & DNF & 2014-2015 & 6 & 0/81 & 0/0/0 \\
BR-Npw \cite{BR-Npw} & WSA & 2014-2017 & 217 & 0/0 & 0/100/32 \\
CA-ARB \cite{CA-ARB} & WET & 2013-2015 & 179 & 0/194 & 0/0/38 \\
CA-ARF \cite{CA-ARF} & WET & 2013-2015 & 230 & 0/0 & 0/127/45 \\
CA-BOU \cite{CA-BOU} & WET & 2018-2020 & 60 & 0/30 & 0/0/18 \\
CA-DB2 \cite{CA-DB2} & WET & 2020-2020 & 0 & 151/0 & 0/0/0 \\
CA-DBB \cite{CA-DBB} & WET & 2016-2020 & 0 & 0/0 & 755/0/0 \\
CA-DSM \cite{CA-DSM} & WET & 2022-2022 & 88 & 0/0 & 0/0/14 \\
CA-ER1 \cite{CA-ER1} & CRO & 2016-2021 & 0 & 567/0 & 0/0/0 \\
CA-Gro \cite{CA-Gro} & MF & 2013-2014 & 0 & 0/166 & 0/0/0 \\
CA-LP1 \cite{CA-LP1} & ENF & 2013-2022 & 674 & 0/0 & 0/185/134 \\
CA-TP1 \cite{CA-TP1} & ENF & 2013-2017 & 0 & 0/0 & 642/0/0 \\
CA-TP3 \cite{CA-TP3} & ENF & 2013-2017 & 496 & 0/0 & 0/192/138 \\
CA-TP4 \cite{CA-TP4} & ENF & 2013-2017 & 0 & 0/0 & 623/0/0 \\
CA-TPD \cite{CA-TPD} & DBF & 2013-2017 & 365 & 0/0 & 0/107/100 \\
CL-SDF \cite{CL-SDF} & EBF & 2014-2022 & 279 & 0/0 & 0/52/66 \\
PE-QFR \cite{PE-QFR} & WET & 2018-2019 & 13 & 0/0 & 0/85/11 \\
PR-xGU \cite{PR-xGU} & EBF & 2019-2022 & 71 & 0/59 & 0/0/29 \\
PR-xLA \cite{PR-xLA} & GRA & 2018-2022 & 92 & 0/63 & 0/0/16 \\
US-A32 \cite{US-A32} & GRA & 2015-2017 & 54 & 0/137 & 0/0/16 \\
US-A74 \cite{US-A74} & CRO & 2016-2017 & 6 & 0/11 & 0/0/1 \\
US-ALQ \cite{US-ALQ} & WET & 2018-2023 & 407 & 0/0 & 0/95/70 \\
US-ARM \cite{US-ARM} & CRO & 2013-2023 & 1029 & 0/169 & 0/0/244 \\
US-ASH \cite{US-ASH} & DBF & 2016-2017 & 0 & 0/53 & 0/0/0 \\
US-ASL \cite{US-ASL} & DBF & 2016-2017 & 0 & 0/0 & 0/19/0 \\
US-ASM \cite{US-ASM} & DBF & 2016-2017 & 0 & 0/58 & 0/0/0 \\
US-An1 \cite{US-An1} & OSH & 2013-2019 & 155 & 0/35 & 0/0/23 \\
US-An2 \cite{US-An2} & OSH & 2013-2019 & 131 & 0/40 & 0/0/50 \\
US-An3 \cite{US-An3} & OSH & 2013-2019 & 190 & 0/0 & 0/44/42 \\
US-BRG \cite{US-BRG} & GRA & 2017-2020 & 268 & 0/0 & 0/82/57 \\
US-BZB \cite{US-BZB} & WET & 2014-2022 & 940 & 0/0 & 0/176/207 \\
US-BZF \cite{US-BZF} & WET & 2014-2022 & 874 & 0/119 & 0/0/176 \\
US-BZS \cite{US-BZS} & ENF & 2015-2021 & 559 & 0/146 & 0/0/188 \\
US-BZo \cite{US-BZo} & WET & 2018-2022 & 0 & 521/0 & 0/0/0 \\
US-Bar \cite{US-Bar} & DBF & 2013-2022 & 0 & 763/0 & 0/0/0 \\
US-Bi1 \cite{US-Bi1} & CRO & 2016-2023 & 0 & 913/0 & 0/0/0 \\
US-Bi2 \cite{US-Bi2} & CRO & 2017-2023 & 391 & 0/0 & 0/87/104 \\
US-CF1 \cite{US-CF1} & CRO & 2017-2021 & 328 & 0/0 & 0/104/66 \\
US-CF2 \cite{US-CF2} & CRO & 2017-2021 & 0 & 0/0 & 359/0/0 \\
US-CF3 \cite{US-CF3} & CRO & 2017-2022 & 0 & 0/0 & 385/0/0 \\
US-CF4 \cite{US-CF4} & CRO & 2017-2021 & 286 & 0/0 & 0/89/52 \\
US-CGG \cite{US-CGG} & GRA & 2019-2022 & 90 & 0/0 & 0/153/17 \\
US-CLF \cite{US-CLF} & CVM & 2017-2020 & 26 & 0/71 & 0/0/6 \\
US-CMW \cite{US-CMW} & DBF & 2013-2022 & 1451 & 0/227 & 0/0/273 \\
US-CPk \cite{US-CPk} & ENF & 2013-2014 & 0 & 0/0 & 41/0/0 \\
US-CRK \cite{US-CRK} & ENF & 2022-2022 & 4 & 0/0 & 0/0/1 \\
US-CRT \cite{US-CRT} & CRO & 2013-2013 & 0 & 0/0 & 59/0/0 \\
US-CS1 \cite{US-CS1} & CRO & 2018-2019 & 0 & 0/100 & 0/0/0 \\
US-CS2 \cite{US-CS2} & ENF & 2018-2022 & 241 & 0/0 & 0/12/79 \\
US-CS3 \cite{US-CS3} & CRO & 2019-2020 & 0 & 0/150 & 0/0/0 \\
US-CS4 \cite{US-CS4} & CRO & 2020-2021 & 0 & 0/130 & 0/0/0 \\
US-CS5 \cite{US-CS5} & CRO & 2021-2021 & 70 & 0/0 & 0/0/15 \\
US-CS6 \cite{US-CS6} & CRO & 2022-2023 & 0 & 119/0 & 0/0/0 \\
US-CS8 \cite{US-CS8} & CRO & 2023-2023 & 50 & 0/0 & 0/0/15 \\
US-Ced \cite{US-Ced} & CSH & 2013-2014 & 74 & 0/0 & 0/142/17 \\
US-DFC \cite{US-DFC} & CRO & 2018-2022 & 0 & 0/0 & 375/0/0 \\
US-DFK \cite{US-DFK} & CRO & 2018-2021 & 0 & 219/0 & 0/0/0 \\
US-DPW \cite{US-DPW} & WET & 2013-2016 & 246 & 0/223 & 0/0/91 \\
US-DS3 \cite{US-DS3} & CRO & 2021-2023 & 55 & 0/62 & 0/0/16 \\
US-Dmg \cite{US-Dmg} & WET & 2021-2022 & 13 & 0/95 & 0/0/3 \\
US-EDN \cite{US-EDN} & WET & 2018-2019 & 56 & 0/105 & 0/0/0 \\
US-Elm \cite{US-Elm} & WET & 2013-2013 & 21 & 0/0 & 0/0/10 \\
US-Esm \cite{US-Esm} & WET & 2013-2013 & 12 & 0/0 & 0/0/0 \\
US-Fcr \cite{US-Fcr} & OSH & 2013-2014 & 31 & 0/97 & 0/0/15 \\
US-GL1 \cite{US-GL1} & WAT & 2013-2021 & 257 & 0/73 & 0/0/71 \\
US-GLE \cite{US-GLE} & ENF & 2018-2020 & 47 & 0/0 & 0/62/12 \\
US-HB1 \cite{US-HB1} & WET & 2019-2023 & 0 & 769/0 & 0/0/0 \\
US-HB2 \cite{US-HB2} & ENF & 2019-2019 & 147 & 0/0 & 0/0/23 \\
US-HB3 \cite{US-HB3} & ENF & 2019-2023 & 532 & 0/0 & 0/241/79 \\
US-HBK \cite{US-HBK} & DBF & 2016-2022 & 0 & 395/0 & 0/0/0 \\
US-HRA \cite{US-HRA} & CRO & 2015-2017 & 0 & 40/0 & 0/0/0 \\
US-HRC \cite{US-HRC} & CRO & 2015-2017 & 16 & 0/19 & 0/0/1 \\
US-Hn2 \cite{US-Hn2} & GRA & 2018-2018 & 27 & 0/0 & 0/0/1 \\
US-Hn3 \cite{US-Hn3} & OSH & 2017-2018 & 1 & 0/81 & 0/0/2 \\
US-Ho1 \cite{US-Ho1} & ENF & 2017-2020 & 183 & 0/123 & 0/0/24 \\
US-Ho2 \cite{US-Ho2} & ENF & 2017-2020 & 0 & 245/0 & 0/0/0 \\
US-Hsm \cite{US-Hsm} & WET & 2021-2023 & 121 & 0/0 & 0/48/19 \\
US-IB1 \cite{US-IB1} & CRO & 2013-2018 & 524 & 0/100 & 0/0/118 \\
US-IB2 \cite{US-IB2} & GRA & 2013-2018 & 0 & 0/0 & 737/0/0 \\
US-ICh \cite{US-ICh} & OSH & 2013-2021 & 0 & 1069/0 & 0/0/0 \\
US-ICs \cite{US-ICs} & WET & 2014-2022 & 0 & 0/0 & 1076/0/0 \\
US-ICt \cite{US-ICt} & OSH & 2013-2022 & 0 & 0/0 & 854/0/0 \\
US-Jo2 \cite{US-Jo2} & OSH & 2014-2020 & 698 & 0/117 & 0/0/149 \\
US-KFS \cite{US-KFS} & GRA & 2013-2018 & 296 & 0/0 & 0/43/67 \\
US-KLS \cite{US-KLS} & GRA & 2013-2019 & 0 & 0/0 & 404/0/0 \\
US-KPL \cite{US-KPL} & WET & 2021-2022 & 0 & 247/0 & 0/0/0 \\
US-KS3 \cite{US-KS3} & WET & 2018-2019 & 0 & 0/146 & 0/0/0 \\
US-Kon \cite{US-Kon} & GRA & 2013-2018 & 0 & 0/0 & 373/0/0 \\
US-MOz \cite{US-MOz} & DBF & 2013-2022 & 1222 & 0/162 & 0/0/272 \\
US-Me2 \cite{US-Me2} & ENF & 2013-2022 & 0 & 0/0 & 250/0/0 \\
US-Me6 \cite{US-Me6} & ENF & 2013-2022 & 1032 & 0/143 & 0/0/181 \\
US-Men \cite{US-Men} & WAT & 2014-2018 & 69 & 0/24 & 0/0/7 \\
US-Mo1 \cite{US-Mo1} & CRO & 2016-2022 & 0 & 0/0 & 1049/0/0 \\
US-Mo2 \cite{US-Mo2} & GRA & 2018-2023 & 0 & 599/0 & 0/0/0 \\
US-Mo3 \cite{US-Mo3} & CRO & 2016-2022 & 695 & 0/0 & 0/189/170 \\
US-Mpj \cite{US-Mpj} & ENF & 2013-2023 & 1380 & 0/0 & 0/185/325 \\
US-Myb \cite{US-Myb} & WET & 2013-2021 & 574 & 0/88 & 0/0/145 \\
US-NC2 \cite{US-NC2} & ENF & 2013-2023 & 0 & 0/0 & 759/0/0 \\
US-NC3 \cite{US-NC3} & ENF & 2013-2021 & 0 & 747/0 & 0/0/0 \\
US-NC4 \cite{US-NC4} & WET & 2015-2021 & 265 & 0/0 & 0/49/72 \\
US-NGB \cite{US-NGB} & SNO & 2013-2023 & 314 & 0/50 & 0/0/58 \\
US-NGC \cite{US-NGC} & GRA & 2017-2023 & 260 & 0/42 & 0/0/53 \\
US-NR1 \cite{US-NR1} & ENF & 2013-2022 & 446 & 0/65 & 0/0/89 \\
US-NR3 \cite{US-NR3} & GRA & 2014-2023 & 0 & 260/0 & 0/0/0 \\
US-NR4 \cite{US-NR4} & GRA & 2014-2023 & 281 & 0/56 & 0/0/86 \\
US-ONA \cite{US-ONA} & GRA & 2016-2021 & 0 & 0/0 & 806/0/0 \\
US-ORv \cite{US-ORv} & WET & 2013-2016 & 117 & 0/0 & 0/50/35 \\
US-OWC \cite{US-OWC} & WET & 2016-2023 & 0 & 0/0 & 128/0/0 \\
US-PSH \cite{US-PSH} & DBF & 2016-2017 & 0 & 55/0 & 0/0/0 \\
US-PSL \cite{US-PSL} & DBF & 2016-2017 & 0 & 0/0 & 0/48/0 \\
US-Pnp \cite{US-Pnp} & WAT & 2016-2022 & 79 & 0/10 & 0/0/10 \\
US-Prr \cite{US-Prr} & ENF & 2013-2022 & 666 & 0/0 & 0/90/151 \\
US-RGA \cite{US-RGA} & CRO & 2021-2023 & 0 & 293/0 & 0/0/0 \\
US-RGB \cite{US-RGB} & CRO & 2021-2023 & 145 & 0/112 & 0/0/32 \\
US-RGF \cite{US-RGF} & CRO & 2023-2023 & 46 & 0/0 & 0/0/19 \\
US-RGW \cite{US-RGW} & CRO & 2022-2023 & 46 & 0/0 & 0/125/16 \\
US-RGo \cite{US-RGo} & CRO & 2021-2023 & 0 & 0/0 & 301/0/0 \\
US-Rls \cite{US-Rls} & CSH & 2014-2022 & 1104 & 0/0 & 0/213/258 \\
US-Rms \cite{US-Rms} & CSH & 2014-2022 & 842 & 0/158 & 0/0/170 \\
US-Ro1 \cite{US-Ro1} & CRO & 2014-2017 & 0 & 305/0 & 0/0/0 \\
US-Ro2 \cite{US-Ro2} & CRO & 2015-2017 & 0 & 0/0 & 223/0/0 \\
US-Ro4 \cite{US-Ro4} & GRA & 2015-2023 & 0 & 0/0 & 1163/0/0 \\
US-Ro5 \cite{US-Ro5} & CRO & 2017-2023 & 0 & 930/0 & 0/0/0 \\
US-Ro6 \cite{US-Ro6} & CRO & 2017-2023 & 0 & 0/0 & 922/0/0 \\
US-Rpf \cite{US-Rpf} & DBF & 2013-2022 & 566 & 0/0 & 0/99/115 \\
US-Rwf \cite{US-Rwf} & CSH & 2014-2022 & 835 & 0/0 & 0/158/169 \\
US-Rws \cite{US-Rws} & OSH & 2014-2022 & 0 & 1383/0 & 0/0/0 \\
US-SP1 \cite{US-SP1} & ENF & 2013-2020 & 528 & 0/0 & 0/107/123 \\
US-SRC \cite{US-SRC} & OSH & 2013-2014 & 0 & 0/0 & 0/67/0 \\
US-SRG \cite{US-SRG} & GRA & 2013-2023 & 0 & 2016/0 & 0/0/0 \\
US-SRM \cite{US-SRM} & WSA & 2013-2023 & 1174 & 0/0 & 0/120/268 \\
US-SRS \cite{US-SRS} & WSA & 2013-2018 & 701 & 0/162 & 0/0/109 \\
US-SSH \cite{US-SSH} & DBF & 2016-2021 & 229 & 0/0 & 0/82/49 \\
US-Seg \cite{US-Seg} & GRA & 2013-2023 & 1413 & 0/0 & 0/185/301 \\
US-Ses \cite{US-Ses} & OSH & 2013-2023 & 0 & 1859/0 & 0/0/0 \\
US-Slt \cite{US-Slt} & DBF & 2013-2014 & 76 & 0/135 & 0/0/15 \\
US-Snd \cite{US-Snd} & GRA & 2013-2014 & 7 & 0/71 & 0/0/5 \\
US-Sne \cite{US-Sne} & GRA & 2016-2020 & 0 & 291/0 & 0/0/0 \\
US-Snf \cite{US-Snf} & GRA & 2018-2020 & 53 & 0/70 & 0/0/16 \\
US-Srr \cite{US-Srr} & WET & 2014-2017 & 0 & 295/0 & 0/0/0 \\
US-StJ \cite{US-StJ} & WET & 2015-2017 & 0 & 489/0 & 0/0/0 \\
US-Syv \cite{US-Syv} & MF & 2013-2023 & 396 & 0/0 & 0/52/71 \\
US-Tw1 \cite{US-Tw1} & WET & 2013-2022 & 0 & 593/0 & 0/0/0 \\
US-Tw2 \cite{US-Tw2} & CRO & 2013-2013 & 5 & 0/0 & 0/0/0 \\
US-Tw3 \cite{US-Tw3} & CRO & 2013-2018 & 422 & 0/144 & 0/0/92 \\
US-Tw4 \cite{US-Tw4} & WET & 2013-2022 & 508 & 0/0 & 0/82/112 \\
US-Tw5 \cite{US-Tw5} & WET & 2018-2020 & 28 & 0/0 & 0/48/6 \\
US-Twt \cite{US-Twt} & CRO & 2013-2017 & 151 & 0/0 & 0/56/30 \\
US-UM3 \cite{US-UM3} & WAT & 2013-2014 & 28 & 0/0 & 0/28/9 \\
US-UMB \cite{US-UMB} & DBF & 2013-2014 & 31 & 0/0 & 0/66/5 \\
US-Uaf \cite{US-Uaf} & ENF & 2013-2022 & 0 & 660/0 & 0/0/0 \\
US-Var \cite{US-Var} & GRA & 2013-2023 & 0 & 0/0 & 1986/0/0 \\
US-Vcm \cite{US-Vcm} & ENF & 2013-2023 & 922 & 0/0 & 0/66/172 \\
US-Vcp \cite{US-Vcp} & ENF & 2013-2023 & 852 & 0/124 & 0/0/194 \\
US-WPT \cite{US-WPT} & WET & 2013-2013 & 0 & 0/0 & 8/0/0 \\
US-Whs \cite{US-Whs} & OSH & 2013-2023 & 0 & 2009/0 & 0/0/0 \\
US-Wjs \cite{US-Wjs} & SAV & 2013-2023 & 1485 & 0/0 & 0/185/301 \\
US-Wkg \cite{US-Wkg} & GRA & 2013-2023 & 1390 & 0/0 & 0/225/346 \\
US-Wrc \cite{US-Wrc} & ENF & 2013-2016 & 272 & 0/177 & 0/0/86 \\
US-xAB \cite{US-xAB} & ENF & 2017-2022 & 136 & 0/0 & 0/87/23 \\
US-xAE \cite{US-xAE} & GRA & 2019-2022 & 215 & 0/118 & 0/0/38 \\
US-xBA \cite{US-xBA} & WET & 2017-2022 & 97 & 0/0 & 0/16/29 \\
US-xBL \cite{US-xBL} & DBF & 2017-2022 & 156 & 0/0 & 0/92/57 \\
US-xBN \cite{US-xBN} & ENF & 2017-2022 & 182 & 0/58 & 0/0/35 \\
US-xBR \cite{US-xBR} & DBF & 2017-2022 & 236 & 0/0 & 0/69/62 \\
US-xCL \cite{US-xCL} & GRA & 2018-2022 & 196 & 0/0 & 0/109/36 \\
US-xCP \cite{US-xCP} & GRA & 2017-2022 & 0 & 0/0 & 353/0/0 \\
US-xDC \cite{US-xDC} & GRA & 2017-2022 & 0 & 561/0 & 0/0/0 \\
US-xDJ \cite{US-xDJ} & ENF & 2018-2022 & 0 & 398/0 & 0/0/0 \\
US-xDL \cite{US-xDL} & MF & 2017-2022 & 118 & 0/0 & 0/73/19 \\
US-xDS \cite{US-xDS} & CVM & 2018-2022 & 257 & 0/0 & 0/69/40 \\
US-xGR \cite{US-xGR} & DBF & 2017-2022 & 155 & 0/0 & 0/77/25 \\
US-xHA \cite{US-xHA} & DBF & 2017-2022 & 0 & 276/0 & 0/0/0 \\
US-xHE \cite{US-xHE} & OSH & 2017-2022 & 178 & 0/0 & 0/98/62 \\
US-xJE \cite{US-xJE} & ENF & 2017-2022 & 254 & 0/0 & 0/104/49 \\
US-xJR \cite{US-xJR} & OSH & 2017-2022 & 0 & 0/0 & 434/0/0 \\
US-xKA \cite{US-xKA} & GRA & 2017-2022 & 159 & 0/86 & 0/0/68 \\
US-xKZ \cite{US-xKZ} & GRA & 2017-2022 & 203 & 0/73 & 0/0/65 \\
US-xLE \cite{US-xLE} & DBF & 2017-2022 & 33 & 0/0 & 0/69/15 \\
US-xMB \cite{US-xMB} & OSH & 2019-2022 & 153 & 0/101 & 0/0/30 \\
US-xML \cite{US-xML} & DBF & 2019-2022 & 0 & 0/0 & 83/0/0 \\
US-xNG \cite{US-xNG} & GRA & 2017-2022 & 246 & 0/87 & 0/0/39 \\
US-xNQ \cite{US-xNQ} & OSH & 2019-2022 & 0 & 0/0 & 280/0/0 \\
US-xNW \cite{US-xNW} & ENF & 2018-2022 & 0 & 0/0 & 91/0/0 \\
US-xRM \cite{US-xRM} & ENF & 2017-2022 & 261 & 0/0 & 0/84/58 \\
US-xRN \cite{US-xRN} & DBF & 2019-2022 & 0 & 0/0 & 186/0/0 \\
US-xSB \cite{US-xSB} & ENF & 2019-2022 & 0 & 0/0 & 217/0/0 \\
US-xSC \cite{US-xSC} & DBF & 2020-2022 & 0 & 0/0 & 100/0/0 \\
US-xSE \cite{US-xSE} & DBF & 2017-2022 & 0 & 287/0 & 0/0/0 \\
US-xSJ \cite{US-xSJ} & SAV & 2019-2022 & 155 & 0/0 & 0/84/26 \\
US-xSL \cite{US-xSL} & CRO & 2017-2022 & 154 & 0/0 & 0/108/27 \\
US-xSP \cite{US-xSP} & ENF & 2017-2022 & 0 & 0/0 & 348/0/0 \\
US-xSR \cite{US-xSR} & OSH & 2017-2022 & 219 & 0/116 & 0/0/57 \\
US-xST \cite{US-xST} & DBF & 2017-2022 & 0 & 612/0 & 0/0/0 \\
US-xTA \cite{US-xTA} & ENF & 2017-2022 & 0 & 0/0 & 318/0/0 \\
US-xTL \cite{US-xTL} & WET & 2017-2022 & 86 & 0/50 & 0/0/32 \\
US-xTR \cite{US-xTR} & DBF & 2017-2022 & 284 & 0/77 & 0/0/45 \\
US-xUK \cite{US-xUK} & DBF & 2017-2022 & 310 & 0/157 & 0/0/91 \\
US-xUN \cite{US-xUN} & MF & 2017-2022 & 276 & 0/0 & 0/84/65 \\
US-xWD \cite{US-xWD} & GRA & 2017-2022 & 222 & 0/92 & 0/0/28 \\
US-xWR \cite{US-xWR} & ENF & 2018-2022 & 185 & 0/73 & 0/0/59 \\
US-xYE \cite{US-xYE} & ENF & 2019-2022 & 153 & 0/0 & 0/94/35 \\
\bottomrule
\end{longtable}

\section{Acknowledgments}

\subsection{Funding}

This material is based upon work supported by the National Science
Foundation under Grant No. 2319597 (LS, JS, AC, LA, BB). We acknowledge
the AmeriFlux data portal as critical to this work. Funding for the
AmeriFlux data portal was provided by the U.S. Department of Energy
Office of Science.

\subsection{Competing interests}

The authors declare that they have no competing interests.

\subsection{Data and materials availability}

The datasets and models used in this study are available from online
repositories. FAR's source code and pretrained weights will be available on
GitHub and Zenodo when published.


\subsection{Resource Use Acknowledgment}

There are increasing calls for AI scientists to acknowledge energy use
and the tensions created in work using AI with promises to improve climate
mitigation ~\cite{Tamburrini2022TheScientists,Henderson2020TowardsLearning}. To support this ethical standard, we
report and acknowledge FAR's resource usage here. FAR
was trained at the University of Oregon on a high-performance computing
cluster utilizing one A100 GPU and associated compute infrastructure.
While resources are shared on UO's HPC, this node's peak power usage is
2.4~kW and training took approximately 7 days, resulting in a maximum
energy usage of 403~kWh. Talapas uses Eugene's relatively low
carbon-footprint supply of power, which is primarily sourced from
hydropower, followed by nuclear and biomass ~\cite{eweb_power_sources}. We
utilize the IPCC's median technology-specific performance parameters
~\cite{schlomer2014technology} for this power mix and estimate equivalent
emissions of 22.4~kg $CO_2e$, or the equivalent of using
about 2.5~gallons of gasoline. While this is relatively small, the
generally extensive use of hydropower has greatly impacted the well-being of
Indigenous Peoples in the Columbia Gorge and across the state of Oregon.
Salmon habitats have been dramatically changed by the presence of
hydropower in the Columbia and Snake Rivers for the past 60 years
~\cite{Dauble2003ImpactsSalmon,Lee1989TheSustainability}. The fall Chinook salmon, a species that reached
near extinction in Oregon in the early 2020s and who are sacred kin of
Indigenous Peoples, has struggled with spawning in the alluvial river
ecosystems due to insufficient modifications of dams and artificial
flows.

\subsection{Declaration of generative AI and AI-assisted technologies in the
manuscript preparation process}

During the preparation of this work, the authors used ChatGPT-5 and Claude Code to
improve the clarity and grammar of the manuscript. After using these
tools, the authors reviewed and edited the content as needed and take
full responsibility for the content of the published article.

\section{Supplementary Figures}
\setcounter{figure}{0}
\renewcommand{\thefigure}{S\arabic{figure}}
\begin{figure}[pos=!htbp]
    \centering
    \includegraphics[width=1\linewidth]{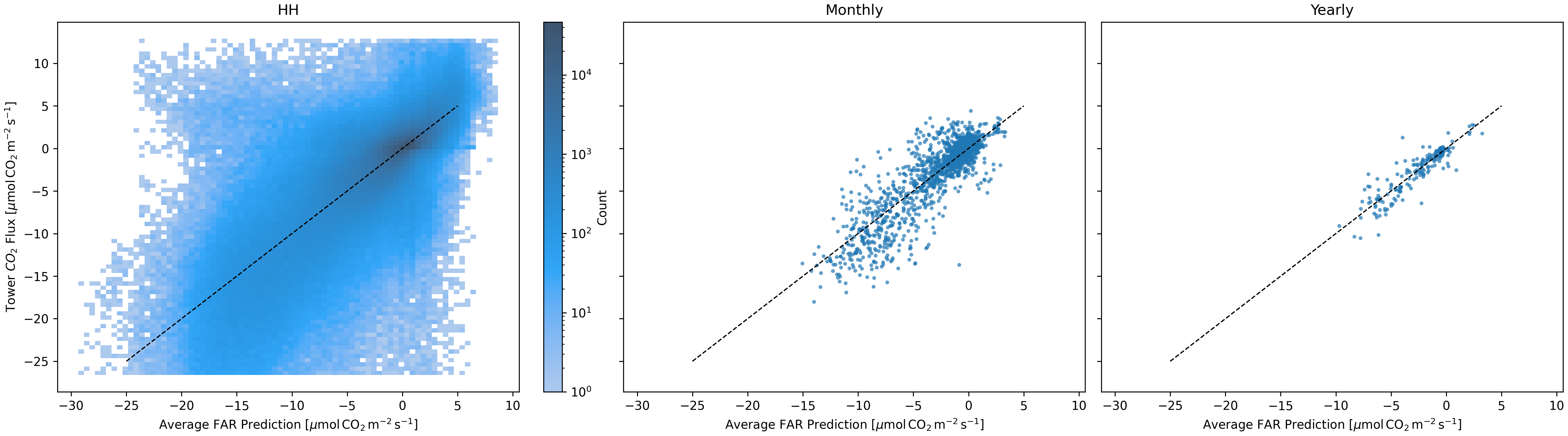}
    \caption{Tower measurements are compared against FAR predictions. The dashed black line denotes a perfect agreement reference.  Half-hour results (HH) are plotted as a histogram due to the large number of estimates with counts concentrated near the reference line and shown on a log color scale. Monthly and yearly values are plotted as scatterplots. These values correspond to the FAR row in Table~\ref{tab:model_performance}.}
    \label{fig:supp_scatter}
\end{figure}
\FloatBarrier
\printcredits

\bibliographystyle{model1-num-names.bst}
\bibliography{site,references,extra} 

@misc{eros2020landsat,
  author    = {{Earth Resources Observation and Science (EROS) Center}},
  title     = {Landsat 8--9 Operational Land Imager / Thermal Infrared Sensor Level-1, Collection 2},
  year      = {2020},
  publisher = {U.S. Geological Survey},
  doi       = {10.5066/P975CC9B},
  note      = {Dataset}
}

@article{justice1998modis_land_remote_sensing,
  author  = {Justice, Christopher O. and Vermote, Eric and Townshend, John R. G. and Defries, Ruth and Roy, David P. and Hall, Dorothy K. and Salomonson, Vincent V. and Privette, Jeffrey L. and Riggs, George and Strahler, Alan and Lucht, Wolfgang and Myneni, Ranga B. and Knyazikhin, Yuri and Running, Steve W. and Nemani, Rama R. and Wan, Zhengming and Huete, Alfredo R. and Van Leeuwen, Wim and Wolfe, Robert E. and Giglio, Louis and Muller, Jan Peter and Lewis, Philip and Barnsley, Michael J.},
  title   = {The moderate resolution imaging spectroradiometer ({MODIS}): Land remote sensing for global change research},
  journal = {IEEE Transactions on Geoscience and Remote Sensing},
  volume  = {36},
  number  = {4},
  pages   = {1228--1249},
  year    = {1998},
  doi     = {10.1109/36.701075},
}

@inproceedings{dosovitskiy2021image,
  title     = {An Image is Worth 16x16 Words: Transformers for Image Recognition at Scale},
  author    = {Dosovitskiy, Alexey and Beyer, Lucas and Kolesnikov, Alexander and Weissenborn, Dirk and Zhai, Xiaohua and Unterthiner, Thomas and Dehghani, Mostafa and Minderer, Matthias and Heigold, Georg and Gelly, Sylvain and Uszkoreit, Jakob and Houlsby, Neil},
  booktitle = {International Conference on Learning Representations},
  year      = {2021},
  url       = {https://openreview.net/forum?id=YicbFdNTTy}
}

@inproceedings{vaswani2017attention,
  title     = {Attention Is All You Need},
  author    = {Vaswani, Ashish and Shazeer, Noam and Parmar, Niki and Uszkoreit, Jakob and Jones, Llion and Gomez, Aidan N. and Kaiser, {\L}ukasz and Polosukhin, Illia},
  booktitle = {Advances in Neural Information Processing Systems},
  volume    = {30},
  year      = {2017},
  url       = {https://papers.nips.cc/paper/7181-attention-is-all-you-need}
}

@misc{PRISMUniversity,
  key          = {PRISM Group},
  title        = {{PRISM Climate Data}},
  howpublished = {{PRISM Group, Oregon State University, https://prism.oregonstate.edu}},
  year = {n.d.},
  note         = {Accessed March 30, 2026}
}

@misc{stafford_pysolar,
  author       = {Stafford, Brandon},
  title        = {{Pysolar: Staring Directly at the Sun Since 2007}},
  howpublished = {\url{https://pysolar.readthedocs.io/en/latest/}},
  version = {0.13},
  year = {2025},
  note         = {Documentation. Accessed April 24, 2025}
}

@misc{forget_landsatxplore,
  author       = {Forget, Yann},
  title        = {{landsatxplore: Search and download Landsat scenes from EarthExplorer}},
  howpublished = {\url{https://github.com/yannforget/landsatxplore}},
  year = {2023},
  version = {0.15.0},
  note         = {Software repository. Accessed  April 24, 2025}

}

@misc{eweb_power_sources,
  author = {{Eugene Water \& Electric Board}},
  title  = {{Where Your Power Comes From}},
  year = {2024},
  howpublished={\url{https://www.eweb.org/your-public-utility/power-supply/where-your-power-comes-from}},
  note   = {Accessed April 24, 2025}
}

@incollection{schlomer2014technology,
  author    = {Schl{\"o}mer, Steffen and Bruckner, Thomas and Fulton, Lew and Hertwich, Edgar and McKinnon, Alan and Perczyk, Daniel and Roy, Joyashree and Schaeffer, Roberto and Sims, Ralph and Smith, Pete and Wiser, Ryan},
  title     = {{Annex III: Technology-specific Cost and Performance Parameters}},
  booktitle = {{Climate Change 2014: Mitigation of Climate Change. Contribution of Working Group III to the Fifth Assessment Report of the Intergovernmental Panel on Climate Change}},
  editor    = {Edenhofer, Ottmar and Pichs-Madruga, Ram{\'o}n and Sokona, Youba and Farahani, Ellie and Kadner, Susanne and Seyboth, Kristin and Adler, Anna and Baum, Ina and Brunner, Steffen and Eickemeier, Patrick and Kriemann, Benjamin and Savolainen, Jussi and Schl{\"o}mer, Steffen and von Stechow, Christoph and Zwickel, Timm and Minx, Jan C.},
  publisher = {Cambridge University Press},
  address   = {Cambridge, United Kingdom and New York, NY, USA},
  year      = {2014},
  pages     = {1329--1356},
doi = {10.1017/CBO9781107415416.025},
}

@inproceedings{fortier2025carbonsense,
  title     = {{CarbonSense: A Multimodal Dataset and Baseline for Carbon Flux Modelling}},
  author    = {Fortier, Matthew and Richter, Mats L. and Sonnentag, Oliver and Pal, Christopher},
  booktitle = {International Conference on Learning Representations},
  year      = {2025},
  url       = {https://proceedings.iclr.cc/paper_files/paper/2025/hash/6752ced903c3f0265108caa10933965f-Abstract-Conference.html}
}

@book{Durrett2019ProbabilityExamples,
  author    = {Durrett, Richard},
  title     = {{Probability: Theory and Examples}},
  edition   = {5},
  publisher = {Cambridge University Press},
  year      = {2019},
  isbn      = {9781108473682}
}

@article{nash1970river,
  title={River flow forecasting through conceptual models part I — A discussion of principles},
  author={Nash, J. E. and Sutcliffe, J. V.},
  journal={Journal of Hydrology},
  volume={10},
  number={3},
  pages={282--290},
  year={1970},
  doi={10.1016/0022-1694(70)90255-6}
}

@article{Loveland01012000,
author = {T. R. Loveland and B. C. Reed and J. F. Brown and D. O. Ohlen and Z. Zhu and L. Yang and J. W. Merchant},
title = {Development of a global land cover characteristics database and IGBP DISCover from 1 km AVHRR data},
journal = {International Journal of Remote Sensing},
volume = {21},
number = {6-7},
pages = {1303--1330},
year = {2000},
publisher = {Taylor \& Francis},
doi = {10.1080/014311600210191},
}

@article{RAISSI2019686,
title = {Physics-informed neural networks: A deep learning framework for solving forward and inverse problems involving nonlinear partial differential equations},
journal = {Journal of Computational Physics},
volume = {378},
pages = {686-707},
year = {2019},
issn = {0021-9991},
doi = {https://doi.org/10.1016/j.jcp.2018.10.045},
url = {https://www.sciencedirect.com/science/article/pii/S0021999118307125},
author = {M. Raissi and P. Perdikaris and G.E. Karniadakis}
}

@misc{fluxcomdataset,
  doi = {10.18160/5NZG-JMJE},
  url = {https://meta.icos-cp.eu/collections/zfwf1Ak2I7OlziGDTX8Xl6_T},
  author = {Nelson, Jacob A. and Walther, Sophia and Jung, Martin and Gans, Fabian and Kraft, Basil and Weber, Ulrich and Hamdi, Zayd and Duveiller, Gregory and Zhang, Weijie},
  title = {FLUXCOM-X-BASE},
  publisher = {ICOS ERIC -- Carbon Portal},
  year = {2023},
  copyright = {Creative Commons Attribution 4.0 International},
  urldate   = {2026-09-22}

}

@Article{bg-21-5079-2024,
AUTHOR = {Nelson, J. A. and Walther, S. and Gans, F. and Kraft, B. and Weber, U. and Novick, K. and Buchmann, N. and Migliavacca, M. and Wohlfahrt, G. and \v{S}igut, L. and Ibrom, A. and Papale, D. and G\"ockede, M. and Duveiller, G. and Knohl, A. and H\"ortnagl, L. and Scott, R. L. and Du\v{s}ek, J. and Zhang, W. and Hamdi, Z. M. and Reichstein, M. and Aranda-Barranco, S. and Ard\"o, J. and Op de Beeck, M. and Billesbach, D. and Bowling, D. and Bracho, R. and Br\"ummer, C. and Camps-Valls, G. and Chen, S. and Cleverly, J. R. and Desai, A. and Dong, G. and El-Madany, T. S. and Euskirchen, E. S. and Feigenwinter, I. and Galvagno, M. and Gerosa, G. A. and Gielen, B. and Goded, I. and Goslee, S. and Gough, C. M. and Heinesch, B. and Ichii, K. and Jackowicz-Korczynski, M. A. and Klosterhalfen, A. and Knox, S. and Kobayashi, H. and Kohonen, K.-M. and Korkiakoski, M. and Mammarella, I. and Gharun, M. and Marzuoli, R. and Matamala, R. and Metzger, S. and Montagnani, L. and Nicolini, G. and O'Halloran, T. and Ourcival, J.-M. and Peichl, M. and Pendall, E. and Ruiz Reverter, B. and Roland, M. and Sabbatini, S. and Sachs, T. and Schmidt, M. and Schwalm, C. R. and Shekhar, A. and Silberstein, R. and Silveira, M. L. and Spano, D. and Tagesson, T. and Tramontana, G. and Trotta, C. and Turco, F. and Vesala, T. and Vincke, C. and Vitale, D. and Vivoni, E. R. and Wang, Y. and Woodgate, W. and Yepez, E. A. and Zhang, J. and Zona, D. and Jung, M.},
TITLE = {X-BASE: the first terrestrial carbon and water flux products from
an extended data-driven scaling framework, FLUXCOM-X},
JOURNAL = {Biogeosciences},
VOLUME = {21},
YEAR = {2024},
NUMBER = {22},
PAGES = {5079--5115},
URL = {https://bg.copernicus.org/articles/21/5079/2024/},
DOI = {10.5194/bg-21-5079-2024}
}

@misc{kimball2025smap,
    author       = {Kimball, John S. and Endsley, Avery and Jones, Lucas A.
                    and Kundig, Tyler and Reichle, Rolf},
    title        = {{SMAP L4 Global Daily 9 km EASE-Grid Carbon Net Ecosystem
                    Exchange, Version 8}},
    year         = {2025},
    howpublished = {SPL4CMDL, Version 8},
    publisher    = {NASA National Snow and Ice Data Center Distributed Active
                    Archive Center},
    address      = {Boulder, Colorado, USA},
    doi          = {10.5067/U7SN8JDZL0UC},
    url          = {https://doi.org/10.5067/U7SN8JDZL0UC},
    note         = {Dataset; subset centered at 44.172 degrees N,
                    122.421 degrees W; accessed 2026-09-22},
    urldate   = {2026-09-22}

  }

@misc{yuan2025gloflux,
    author       = {Yuan, Qiwang and Wang, Xufeng and Che, Tao and Li, Jun},
    title        = {Global Carbon Flux Dataset Generated by Fusing Remote
                    Sensing and Multiple Flux Networks Observation},
    year         = {2025},
    publisher    = {National Tibetan Plateau Data Center},
    doi          = {10.11888/Terre.tpdc.301870},
    url          = {https://doi.org/10.11888/Terre.tpdc.301870},
    note         = {GloFlux dataset}
  }

@article{yuan2025glofluxpaper,
    author       = {Yuan, Qiwang and Wang, Xufeng and Che, Tao and Li, Jun},
    title        = {Global Carbon Flux Dataset Generated by Fusing Remote
                    Sensing and Multiple Flux Networks Observation},
    journal      = {Scientific Data},
    year         = {2025},
    volume       = {12},
    pages        = {1359},
    doi          = {10.1038/s41597-025-05672-8},
    url          = {https://doi.org/10.1038/s41597-025-05672-8}
  }

@article{han2026geo3nep,
    author  = {Han, Qizhi and Liu, Liangyun and Liu, Xinjie and Zhu, A-xing},
    title   = {A Geo-Prediction Model for Estimating Global Terrestrial
               Net Ecosystem Productivity Based on the Third Law of Geography},
    journal = {Ecological Informatics},
    year    = {2026},
    month   = {6},
    volume  = {96},
    pages   = {103844},
    issn    = {1574-9541},
    doi     = {10.1016/j.ecoinf.2026.103844},
    url     = {https://doi.org/10.1016/j.ecoinf.2026.103844}
  }

@article{Turner2006ANEP,
    title = {{A diagnostic carbon flux model to monitor the effects of disturbance and interannual variation in climate on regional NEP}},
    year = {2006},
    journal = {Tellus B},
    author = {Turner, David P. and Ritts, W. D. and Styles, J. M. and Yang, Z. and Cohen, W. B. and Law, B. E. and Thornton, P. E.},
    number = {5},
    month = {11},
    pages = {476--490},
    volume = {58},
    publisher = {John Wiley {\&} Sons, Ltd},
    url = {https://onlinelibrary.wiley.com/doi/full/10.1111/j.1600-0889.2006.00221.x https://onlinelibrary.wiley.com/doi/abs/10.1111/j.1600-0889.2006.00221.x https://onlinelibrary.wiley.com/doi/10.1111/j.1600-0889.2006.00221.x},
    doi = {10.1111/J.1600-0889.2006.00221.X},
    issn = {1600-0889}
}

@article{Kljun2015AFFP,
    title = {{A simple two-dimensional parameterisation for Flux Footprint Prediction (FFP)}},
    year = {2015},
    journal = {Geoscientific Model Development},
    author = {Kljun, N. and Calanca, P. and Rotach, M. W. and Schmid, H. P.},
    number = {11},
    month = {11},
    pages = {3695--3713},
    volume = {8},
    publisher = {Copernicus GmbH},
    doi = {10.5194/GMD-8-3695-2015},
    issn = {19919603}
}

@article{Lundberg2017APredictions,
    title = {{A Unified Approach to Interpreting Model Predictions}},
    year = {2017},
    journal = {Advances in Neural Information Processing Systems},
    author = {Lundberg, Scott M and Allen, Paul G and Lee, Su-In},
    volume = {30},
    url = {https://github.com/slundberg/shap}
}

@article{Kingma2014Adam:Optimization,
    title = {{Adam: A Method for Stochastic Optimization}},
    year = {2014},
    journal = {3rd International Conference on Learning Representations, ICLR 2015 - Conference Track Proceedings},
    author = {Kingma, Diederik P. and Ba, Jimmy Lei},
    month = {12},
    publisher = {International Conference on Learning Representations, ICLR},
    url = {https://arxiv.org/abs/1412.6980v9},
    arxivId = {1412.6980}
}

@article{Chu2023AmeriFluxSharing,
    title = {{AmeriFlux BASE data pipeline to support network growth and data sharing}},
    year = {2023},
    journal = {Scientific Data 2023 10:1},
    author = {Chu, Housen and Christianson, Danielle S. and Cheah, You Wei and Pastorello, Gilberto and O’Brien, Fianna and Geden, Joshua and Ngo, Sy Toan and Hollowgrass, Rachel and Leibowitz, Karla and Beekwilder, Norman F. and Sandesh, Megha and Dengel, Sigrid and Chan, Stephen W. and Santos, André and Delwiche, Kyle and Yi, Koong and Buechner, Christin and Baldocchi, Dennis and Papale, Dario and Keenan, Trevor F. and Biraud, Sébastien C. and Agarwal, Deborah A. and Torn, Margaret S.},
    number = {1},
    month = {9},
    pages = {1--13},
    volume = {10},
    publisher = {Nature Publishing Group},
    url = {https://www.nature.com/articles/s41597-023-02531-2},
    doi = {10.1038/s41597-023-02531-2},
    issn = {2052-4463},
    pmid = {37696825}
}

@article{Fang2024AssessingFLUXNET2015,
    title = {{Assessing Spatial Representativeness of Global Flux Tower Eddy-Covariance Measurements Using Data from FLUXNET2015}},
    year = {2024},
    journal = {Scientific Data 2024 11:1},
    author = {Fang, Junjun and Fang, Jingchun and Chen, Baozhang and Zhang, Huifang and Dilawar, Adil and Guo, Man and Liu, Shu’an},
    number = {1},
    month = {6},
    pages = {1--16},
    volume = {11},
    publisher = {Nature Publishing Group},
    url = {https://www.nature.com/articles/s41597-024-03291-3},
    doi = {10.1038/s41597-024-03291-3},
    issn = {2052-4463},
    pmid = {38830898}
}

@article{Ilse2018Attention-basedLearning,
    title = {{Attention-based Deep Multiple Instance Learning}},
    year = {2018},
    journal = {35th International Conference on Machine Learning, ICML 2018},
    author = {Ilse, Maximilian and Tomczak, Jakub M. and Welling, Max},
    month = {2},
    pages = {3376--3391},
    volume = {5},
    publisher = {International Machine Learning Society (IMLS)},
    url = {https://arxiv.org/abs/1802.04712v4},
    isbn = {9781510867963},
    arxivId = {1802.04712}
}

@article{Rey-Sanchez2022DetectingMaps,
    title = {{Detecting Hot Spots of Methane Flux Using Footprint-Weighted Flux Maps}},
    year = {2022},
    journal = {Journal of Geophysical Research: Biogeosciences},
    author = {Rey-Sanchez, Camilo and Arias-Ortiz, Ariane and Kasak, Kuno and Chu, Housen and Szutu, Daphne and Verfaillie, Joseph and Baldocchi, Dennis},
    number = {8},
    month = {8},
    pages = {e2022JG006977},
    volume = {127},
    publisher = {John Wiley {\&} Sons, Ltd},
    url = {https://onlinelibrary.wiley.com/doi/full/10.1029/2022JG006977 https://onlinelibrary.wiley.com/doi/abs/10.1029/2022JG006977 https://agupubs.onlinelibrary.wiley.com/doi/10.1029/2022JG006977},
    doi = {10.1029/2022JG006977},
    issn = {2169-8961}
}

@article{Law2004DisturbanceUSA,
    title = {{Disturbance and climate effects on carbon stocks and fluxes across Western Oregon USA}},
    year = {2004},
    journal = {Global Change Biology},
    author = {Law, Beverly E. and Turner, D. and Campbell, J. and Sun, O. J. and Van Tuyl, S. and Ritts, W. D. and Cohen, W. B.},
    number = {9},
    month = {9},
    pages = {1429--1444},
    volume = {10},
    publisher = {John Wiley {\&} Sons, Ltd},
    url = {https://onlinelibrary.wiley.com/doi/full/10.1111/j.1365-2486.2004.00822.x https://onlinelibrary.wiley.com/doi/abs/10.1111/j.1365-2486.2004.00822.x https://onlinelibrary.wiley.com/doi/10.1111/j.1365-2486.2004.00822.x},
    doi = {10.1111/J.1365-2486.2004.00822.X},
    issn = {1365-2486}
}

@article{Turner2003EffectsLandscape,
    title = {{Effects of land use and fine-scale environmental heterogeneity on net ecosystem production over a temperate coniferous forest landscape}},
    year = {2003},
    journal = {Tellus B: Chemical and Physical Meteorology},
    author = {Turner, David P. and Guzy, Michael and Lefsky, Michael A. and Van Tuyl, Steve and Sun, Osbert and Daly, Chris and Law, Beverly E.},
    number = {2},
    month = {1},
    pages = {657},
    volume = {55},
    publisher = {Stockholm University Press},
    doi = {10.3402/TELLUSB.V55I2.16756},
    issn = {0280-6509}
}

@article{Smith1995ForecastingSatellite,
    title = {{Forecasting wheat yield in a Mediterranean-type environment from the NOAA satellite}},
    year = {1995},
    journal = {Australian Journal of Agricultural Research},
    author = {Smith, R. C.G. and Adams, J. and Stephens, D. J. and Hick, P. T.},
    number = {1},
    month = {2},
    pages = {113--125},
    volume = {46},
    publisher = {CSIRO Publishing},
    url = {/cp/article/46/1/113/32746/Forecasting-wheat-yield-in-a-Mediterranean-type https://dx.doi.org/10.1071/AR9950113},
    doi = {10.1071/AR9950113},
    issn = {0004-9409}
}

@article{Dauble2003ImpactsSalmon,
    title = {{Impacts of the Columbia River hydroelectric system on main-stem habitats of fall chinook salmon}},
    year = {2003},
    journal = {North American Journal of Fisheries Management},
    author = {Dauble, D. D. and Hanrahan, T. P. and Geist, D. R. and Parsley, M. J.},
    number = {3},
    month = {8},
    pages = {641--659},
    volume = {23},
    publisher = {Wiley},
    doi = {10.1577/M02-013},
    issn = {0275-5947}
}

@article{Reyes-Munoz2024InferringModels,
    title = {{Inferring global terrestrial carbon fluxes from the synergy of Sentinel 3 {\&} 5P with Gaussian process hybrid models}},
    year = {2024},
    journal = {Remote Sensing of Environment},
    author = {Reyes-Mu{\~{n}}oz, Pablo and D.Kov{\'{a}}cs, Dávid and Berger, Katja and Pipia, Luca and Belda, Santiago and Rivera-Caicedo, Juan Pablo and Verrelst, Jochem},
    month = {5},
    pages = {114072},
    volume = {305},
    publisher = {Elsevier},
    doi = {10.1016/J.RSE.2024.114072},
    issn = {0034-4257}
}

@article{Rumelhart1986LearningErrors,
    title = {{Learning representations by back-propagating errors}},
    year = {1986},
    journal = {Nature 1986 323:6088},
    author = {Rumelhart, David E. and Hinton, Geoffrey E. and Williams, Ronald J.},
    number = {6088},
    pages = {533--536},
    volume = {323},
    publisher = {Nature Publishing Group},
    url = {https://www.nature.com/articles/323533a0},
    doi = {10.1038/323533a0},
    issn = {1476-4687}
}

@article{MartinBelda2022LPJ-GUESS/LSMv1.0:Realism,
    title = {{LPJ-GUESS/LSMv1.0: A next-generation land surface model with high ecological realism}},
    year = {2022},
    journal = {Geoscientific Model Development},
    author = {Mart{\'{i}}n Belda, David and Anthoni, Peter and W{\aa}rlind, David and Olin, Stefan and Schurgers, Guy and Tang, Jing and Smith, Benjamin and Arneth, Almut},
    number = {17},
    month = {9},
    pages = {6709--6745},
    volume = {15},
    publisher = {Copernicus Publications},
    doi = {10.5194/GMD-15-6709-2022},
    issn = {19919603}
}

@article{Kong2022MatchingEstimates,
    title = {{Matching high resolution satellite data and flux tower footprints improves their agreement in photosynthesis estimates}},
    year = {2022},
    journal = {Agricultural and Forest Meteorology},
    author = {Kong, Juwon and Ryu, Youngryel and Liu, Jiangong and Dechant, Benjamin and Rey-Sanchez, Camilo and Shortt, Robert and Szutu, Daphne and Verfaillie, Joe and Houborg, Rasmus and Baldocchi, Dennis D.},
    month = {4},
    pages = {108878},
    volume = {316},
    publisher = {Elsevier},
    doi = {10.1016/J.AGRFORMET.2022.108878},
    issn = {0168-1923}
}

@article{Rouse1974MonitoringERTS,
    title = {{Monitoring vegetation systems in the Great Plains with ERTS}},
    year = {1974},
    journal = {NASA.  Goddard Space Flight Center  3d ERTS-1 Symp., Vol. 1, Sect. A},
    author = {Rouse, J. W., Jr. and Haas, R. H. and Schell, J. A. and Deering, D. W.}
}

@article{Gao1996NDWIASpace,
    title = {{NDWI—A normalized difference water index for remote sensing of vegetation liquid water from space}},
    year = {1996},
    journal = {Remote Sensing of Environment},
    author = {Gao, Bo Cai},
    number = {3},
    month = {12},
    pages = {257--266},
    volume = {58},
    publisher = {Elsevier},
    doi = {10.1016/S0034-4257(96)00067-3},
    issn = {0034-4257}
}

@article{Huete2002OverviewIndices,
    title = {{Overview of the radiometric and biophysical performance of the MODIS vegetation indices}},
    year = {2002},
    journal = {Remote Sensing of Environment},
    author = {Huete, A. and Didan, K. and Miura, T. and Rodriguez, E. P. and Gao, X. and Ferreira, L. G.},
    number = {1-2},
    month = {11},
    pages = {195--213},
    volume = {83},
    publisher = {Elsevier},
    doi = {10.1016/S0034-4257(02)00096-2},
    issn = {0034-4257}
}

@article{Paszke2019PyTorch:Library,
    title = {{PyTorch: An Imperative Style, High-Performance Deep Learning Library}},
    year = {2019},
    author = {Paszke, Adam and Gross, Sam and Massa, Francisco and Lerer, Adam and Bradbury Google, James and Chanan, Gregory and Killeen, Trevor and Lin, Zeming and Gimelshein, Natalia and Antiga, Luca and Desmaison, Alban and Xamla, Andreas Köpf and Yang, Edward and Devito, Zach and Raison Nabla, Martin and Tejani, Alykhan and Chilamkurthy, Sasank and Ai, Qure and Steiner, Benoit and Facebook, Lu Fang and Facebook, Junjie Bai and Chintala, Soumith},
    url = {https://dl.acm.org/doi/10.5555/3454287.3455008},
    doi = {10.5555/3454287.3455008}
}

@article{Tucker1979RedVegetation,
    title = {{Red and photographic infrared linear combinations for monitoring vegetation}},
    year = {1979},
    journal = {Remote Sensing of Environment},
    author = {Tucker, Compton J.},
    number = {2},
    month = {5},
    pages = {127--150},
    volume = {8},
    publisher = {Elsevier},
    doi = {10.1016/0034-4257(79)90013-0},
    issn = {0034-4257}
}

@article{Chu2021RepresentativenessSites,
    title = {{Representativeness of Eddy-Covariance flux footprints for areas surrounding AmeriFlux sites}},
    year = {2021},
    journal = {Agricultural and Forest Meteorology},
    author = {Chu, Housen and Luo, Xiangzhong and Ouyang, Zutao and Chan, W. Stephen and Dengel, Sigrid and Biraud, Sébastien C. and Torn, Margaret S. and Metzger, Stefan and Kumar, Jitendra and Arain, M. Altaf and Arkebauer, Tim J. and Baldocchi, Dennis and Bernacchi, Carl and Billesbach, Dave and Black, T. Andrew and Blanken, Peter D. and Bohrer, Gil and Bracho, Rosvel and Brown, Shannon and Brunsell, Nathaniel A. and Chen, Jiquan and Chen, Xingyuan and Clark, Kenneth and Desai, Ankur R. and Duman, Tomer and Durden, David and Fares, Silvano and Forbrich, Inke and Gamon, John A. and Gough, Christopher M. and Griffis, Timothy and Helbig, Manuel and Hollinger, David and Humphreys, Elyn and Ikawa, Hiroki and Iwata, Hiroki and Ju, Yang and Knowles, John F. and Knox, Sara H. and Kobayashi, Hideki and Kolb, Thomas and Law, Beverly and Lee, Xuhui and Litvak, Marcy and Liu, Heping and Munger, J. William and Noormets, Asko and Novick, Kim and Oberbauer, Steven F. and Oechel, Walter and Oikawa, Patty and Papuga, Shirley A. and Pendall, Elise and Prajapati, Prajaya and Prueger, John and Quinton, William L. and Richardson, Andrew D. and Russell, Eric S. and Scott, Russell L. and Starr, Gregory and Staebler, Ralf and Stoy, Paul C. and Stuart-Ha{\"{e}}ntjens, Ellen and Sonnentag, Oliver and Sullivan, Ryan C. and Suyker, Andy and Ueyama, Masahito and Vargas, Rodrigo and Wood, Jeffrey D. and Zona, Donatella},
    month = {5},
    pages = {108350},
    volume = {301-302},
    publisher = {Elsevier},
    doi = {10.1016/J.AGRFORMET.2021.108350},
    issn = {0168-1923}
}

@article{Jung2020ScalingApproach,
    title = {{Scaling carbon fluxes from eddy covariance sites to globe: Synthesis and evaluation of the FLUXCOM approach}},
    year = {2020},
    journal = {Biogeosciences},
    author = {Jung, Martin and Schwalm, Christopher and Migliavacca, Mirco and Walther, Sophia and Camps-Valls, Gustau and Koirala, Sujan and Anthoni, Peter and Besnard, Simon and Bodesheim, Paul and Carvalhais, Nuno and Chevallier, Frederic and Gans, Fabian and S Goll, Daniel and Haverd, Vanessa and K{\"{o}}hler, Philipp and Ichii, Kazuhito and K Jain, Atul and Liu, Junzhi and Lombardozzi, Danica and E M S Nabel, Julia and A Nelson, Jacob and O'Sullivan, Michael and Pallandt, Martijn and Papale, Dario and Peters, Wouter and Pongratz, Julia and R{\"{o}}denbeck, Christian and Sitch, Stephen and Tramontana, Gianluca and Walker, Anthony and Weber, Ulrich and Reichstein, Markus},
    number = {5},
    month = {3},
    pages = {1343--1365},
    volume = {17},
    publisher = {Copernicus GmbH},
    doi = {10.5194/BG-17-1343-2020},
    issn = {17264189}
}

@article{Ran2016SpatialScale,
    title = {{Spatial representativeness and uncertainty of eddy covariance carbon flux measurements for upscaling net ecosystem productivity to the grid scale}},
    year = {2016},
    journal = {Agricultural and Forest Meteorology},
    author = {Ran, Youhua and Li, Xin and Sun, Rui and Kljun, Natascha and Zhang, Lei and Wang, Xufeng and Zhu, Gaofeng},
    month = {12},
    pages = {114--127},
    volume = {230-231},
    publisher = {Elsevier},
    doi = {10.1016/J.AGRFORMET.2016.05.008},
    issn = {0168-1923}
}

@article{Peltola2015StudyingHeight,
    title = {{Studying the spatial variability of methane flux with five eddy covariance towers of varying height}},
    year = {2015},
    journal = {Agricultural and Forest Meteorology},
    author = {Peltola, O. and Hensen, A. and Belelli Marchesini, L. and Helfter, C. and Bosveld, F. C. and van den Bulk, W. C.M. and Haapanala, S. and van Huissteden, J. and Laurila, T. and Lindroth, A. and Nemitz, E. and R{\"{o}}ckmann, T. and Vermeulen, A. T. and Mammarella, I.},
    month = {12},
    pages = {456--472},
    volume = {214-215},
    publisher = {Elsevier},
    doi = {10.1016/J.AGRFORMET.2015.09.007},
    issn = {0168-1923}
}

@article{Tamburrini2022TheScientists,
    title = {{The AI Carbon Footprint and Responsibilities of AI Scientists}},
    year = {2022},
    journal = {Philosophies 2022, Vol. 7, Page 4},
    author = {Tamburrini, Guglielmo},
    number = {1},
    month = {1},
    pages = {4},
    volume = {7},
    publisher = {Multidisciplinary Digital Publishing Institute},
    url = {https://www.mdpi.com/2409-9287/7/1/4/htm https://www.mdpi.com/2409-9287/7/1/4},
    doi = {10.3390/PHILOSOPHIES7010004},
    issn = {2409-9287}
}

@misc{Lee1989TheSustainability,
    title = {{The Columbia river basin. Experimenting with sustainability}},
    year = {1989},
    booktitle = {Environment},
    author = {Lee, K. N.},
    number = {6},
    volume = {31},
    doi = {10.1080/00139157.1989.9928950},
    issn = {00139157}
}

@inproceedings{Key1999TheSeverity.,
    title = {{The Normalized Burn Ratio (NBR): A Landsat TM radiometric index of burn severity.}},
    year = {1999},
    booktitle = {USGS/NPS Joint Fire Science Conference Proceedings},
    author = {Key, Christopher and Benson, Nancy},
    publisher = {U.S. Geological Survey, National Park Service, Remote Sensing Applications Center.},
    address = {Boise, Idaho}
}

@article{Papale2006TowardsEstimation,
    title = {{Towards a standardized processing of Net Ecosystem Exchange measured with eddy covariance technique: Algorithms and uncertainty estimation}},
    year = {2006},
    journal = {Biogeosciences},
    author = {Papale, D. and Reichstein, M. and Aubinet, M. and Canfora, E. and Bernhofer, C. and Kutsch, W. and Longdoz, B. and Rambal, S. and Valentini, R. and Vesala, T. and Yakir, D.},
    number = {4},
    pages = {571--583},
    volume = {3},
    publisher = {European Geosciences Union},
    doi = {10.5194/BG-3-571-2006},
    issn = {17264189}
}

@article{Henderson2020TowardsLearning,
    title = {{Towards the Systematic Reporting of the Energy and Carbon Footprints of Machine Learning}},
    year = {2020},
    journal = {Journal of Machine Learning Research},
    author = {Henderson, Peter and Hu, Jieru and Romoff, Joshua and Brunskill, Emma and Jurafsky, Dan and Pineau, Joelle},
    pages = {1--43},
    volume = {21},
    url = {http://jmlr.org/papers/v21/20-312.html.}
}

@article{Reyes-Munoz2025Tower-to-globalData,
    title = {{Tower-to-global upscaling of terrestrial carbon fluxes driven by MODIS-LAI, Sentinel-3-LAI and ERA5-Land data}},
    year = {2025},
    journal = {Ecological Indicators},
    author = {Reyes-Mu{\~{n}}oz, Pablo and D.Kov{\'{a}}cs, Dávid and Verrelst, Jochem},
    month = {8},
    pages = {113597},
    volume = {177},
    publisher = {Elsevier},
    doi = {10.1016/J.ECOLIND.2025.113597},
    issn = {1470-160X}
}

@article{Ronneberger2015U-Net:Segmentation,
    title = {{U-Net: Convolutional Networks for Biomedical Image Segmentation}},
    year = {2015},
    journal = {Lecture Notes in Computer Science (including subseries Lecture Notes in Artificial Intelligence and Lecture Notes in Bioinformatics)},
    author = {Ronneberger, Olaf and Fischer, Philipp and Brox, Thomas},
    pages = {234--241},
    volume = {9351},
    publisher = {Springer, Cham},
    url = {https://link.springer.com/chapter/10.1007/978-3-319-24574-4_28},
    isbn = {978-3-319-24574-4},
    doi = {10.1007/978-3-319-24574-4{\_}28},
    issn = {1611-3349}
}

@article{Zhu2024UniformUFLUX,
    title = {{Uniform upscaling techniques for eddy covariance FLUXes (UFLUX)}},
    year = {2024},
    journal = {International Journal of Remote Sensing},
    author = {Zhu, Songyan and Quaife, Tristan and Hill, Timothy},
    number = {5},
    month = {3},
    pages = {1450--1476},
    volume = {45},
    publisher = {Taylor {\&} Francis},
    url = {https://www.tandfonline.com/doi/abs/10.1080/01431161.2024.2312266},
    doi = {10.1080/01431161.2024.2312266},
    issn = {13665901}
}

@article{Xu2017UpscalingFunctions,
    title = {{Upscaling tower-observed turbulent exchange at fine spatio-temporal resolution using environmental response functions}},
    year = {2017},
    journal = {Agricultural and Forest Meteorology},
    author = {Xu, Ke and Metzger, Stefan and Desai, Ankur R.},
    month = {1},
    pages = {10--22},
    volume = {232},
    publisher = {Elsevier},
    doi = {10.1016/J.AGRFORMET.2016.07.019},
    issn = {0168-1923}
}

@article{Chen2016XGBoost:System,
    title = {{XGBoost: A Scalable Tree Boosting System}},
    year = {2016},
    journal = {Proceedings of the 22nd ACM SIGKDD International Conference on Knowledge Discovery and Data Mining},
    author = {Chen, Tianqi and Guestrin, Carlos},
    publisher = {ACM},
    url = {http://dx.doi.org/10.1145/2939672.2939785},
    address = {New York, NY, USA},
    isbn = {9781450342322},
    doi = {10.1145/2939672}
}

@misc{AR-CCg,
  title = {{AmeriFlux BASE AR-CCg Carlos Casares grassland, Ver. 1-5}},
  year = {2022},
  author = {Posse, Gabriela},
  publisher = {AmeriFlux AMP},
  doi = {10.17190/AMF/1865474},
  note = {Dataset},
  site = {AR-CCg}
}

@misc{AR-TF1,
  title = {{AmeriFlux BASE AR-TF1 Rio Moat bog, Ver. 2-5}},
  year = {2021},
  author = {Kutzbach, Lars},
  publisher = {AmeriFlux AMP},
  doi = {10.17190/AMF/1543389},
  note = {Dataset},
  site = {AR-TF1}
}

@misc{AR-TF2,
  title = {{AmeriFlux BASE AR-TF2 Rio Pipo bog, Ver. 1-5}},
  year = {2019},
  author = {Kutzbach, Lars},
  publisher = {AmeriFlux AMP},
  doi = {10.17190/AMF/1543388},
  note = {Dataset},
  site = {AR-TF2}
}

@misc{BR-CST,
  title = {{AmeriFlux BASE BR-CST Caatinga Serra Talhada, Ver. 1-5}},
  year = {2019},
  author = {Antonino, Antonio},
  publisher = {AmeriFlux AMP},
  doi = {10.17190/AMF/1562386},
  note = {Dataset},
  site = {BR-CST}
}

@misc{BR-Npw,
  title = {{AmeriFlux BASE BR-Npw Northern Pantanal Wetland, Ver. 1-5}},
  year = {2019},
  author = {Vourlitis, George and Dalmagro, Higo and Nogueira, José de S. and Johnson, Mark and Arruda, Paulo},
  publisher = {AmeriFlux AMP},
  doi = {10.17190/AMF/1579716},
  note = {Dataset},
  site = {BR-Npw}
}

@misc{CA-ARB,
  title = {{AmeriFlux BASE CA-ARB Attawapiskat River Bog, Ver. 1-5}},
  year = {2018},
  author = {Todd, Aaron and Humphreys, Elyn},
  publisher = {AmeriFlux AMP},
  doi = {10.17190/AMF/1480319},
  note = {Dataset},
  site = {CA-ARB}
}

@misc{CA-ARF,
  title = {{AmeriFlux BASE CA-ARF Attawapiskat River Fen, Ver. 1-5}},
  year = {2018},
  author = {Todd, Aaron and Humphreys, Elyn},
  publisher = {AmeriFlux AMP},
  doi = {10.17190/AMF/1480318},
  note = {Dataset},
  site = {CA-ARF}
}

@misc{CA-BOU,
  title = {{AmeriFlux BASE CA-BOU Bouleau Peatland, Ver. 2-5}},
  year = {2023},
  author = {Garneau, Michelle},
  publisher = {AmeriFlux AMP},
  doi = {10.17190/AMF/1987599},
  note = {Dataset},
  site = {CA-BOU}
}

@misc{CA-DB2,
  title = {{AmeriFlux BASE CA-DB2 Delta Burns Bog 2, Ver. 1-5}},
  year = {2021},
  author = {Knox, Sara},
  publisher = {AmeriFlux AMP},
  doi = {10.17190/AMF/1811362},
  note = {Dataset},
  site = {CA-DB2}
}

@misc{CA-DBB,
  title = {{AmeriFlux BASE CA-DBB Delta Burns Bog, Ver. 2-5}},
  year = {2021},
  author = {Christen, Andreas and Knox, Sara},
  publisher = {AmeriFlux AMP},
  doi = {10.17190/AMF/1543378},
  note = {Dataset},
  site = {CA-DBB}
}

@misc{CA-DSM,
  title = {{AmeriFlux BASE CA-DSM Delta Salt Marsh, Ver. 1-5}},
  year = {2023},
  author = {Knox, Sara},
  publisher = {AmeriFlux AMP},
  doi = {10.17190/AMF/1964085},
  note = {Dataset},
  site = {CA-DSM}
}

@misc{CA-ER1,
  title = {{AmeriFlux BASE CA-ER1 Elora Research Station, Ver. 3-5}},
  year = {2021},
  author = {Wagner-Riddle, Claudia},
  publisher = {AmeriFlux AMP},
  doi = {10.17190/AMF/1579541},
  note = {Dataset},
  site = {CA-ER1}
}

@misc{CA-Gro,
  title = {{AmeriFlux BASE CA-Gro Ontario - Groundhog River, Boreal Mixedwood Forest, Ver. 2-5}},
  year = {2019},
  author = {McCaughey, Harry},
  publisher = {AmeriFlux AMP},
  doi = {10.17190/AMF/1245996},
  note = {Dataset},
  site = {CA-Gro}
}

@misc{CA-LP1,
  title = {{AmeriFlux BASE CA-LP1 British Columbia - Mountain pine beetle-attacked lodgepole pine stand , Ver. 4-5}},
  year = {2023},
  author = {Black, Thomas Andrew},
  publisher = {AmeriFlux AMP},
  doi = {10.17190/AMF/1660337},
  note = {Dataset},
  site = {CA-LP1}
}

@misc{CA-TP1,
  title = {{AmeriFlux BASE CA-TP1 Ontario - Turkey Point 2002 Plantation White Pine, Ver. 3-5}},
  year = {2018},
  author = {Arain, M. Altaf},
  publisher = {AmeriFlux AMP},
  doi = {10.17190/AMF/1246009},
  note = {Dataset},
  site = {CA-TP1}
}

@misc{CA-TP3,
  title = {{AmeriFlux BASE CA-TP3 Ontario - Turkey Point 1974 Plantation White Pine, Ver. 3-5}},
  year = {2018},
  author = {Arain, M. Altaf},
  publisher = {AmeriFlux AMP},
  doi = {10.17190/AMF/1246011},
  note = {Dataset},
  site = {CA-TP3}
}

@misc{CA-TP4,
  title = {{AmeriFlux BASE CA-TP4 Ontario - Turkey Point 1939 Plantation White Pine, Ver. 4-5}},
  year = {2018},
  author = {Arain, M. Altaf},
  publisher = {AmeriFlux AMP},
  doi = {10.17190/AMF/1246012},
  note = {Dataset},
  site = {CA-TP4}
}

@misc{CA-TPD,
  title = {{AmeriFlux BASE CA-TPD Ontario - Turkey Point Mature Deciduous, Ver. 2-5}},
  year = {2018},
  author = {Arain, M. Altaf},
  publisher = {AmeriFlux AMP},
  doi = {10.17190/AMF/1246152},
  note = {Dataset},
  site = {CA-TPD}
}

@misc{CL-SDF,
  title = {{AmeriFlux BASE CL-SDF Senda Darwin Forest, Ver. 1-5}},
  year = {2022},
  author = {Perez-Quezada, Jorge and Armesto, Juan J.},
  publisher = {AmeriFlux AMP},
  doi = {10.17190/AMF/1902273},
  note = {Dataset},
  site = {CL-SDF}
}

@misc{PE-QFR,
  title = {{AmeriFlux BASE PE-QFR Quistococha Forest Reserve, Ver. 2-5}},
  year = {2021},
  author = {Roman, Tyler and Griffis, Timothy and Kolka, Randy and Wayson, Craig and Lilleskov, Erik and Torres, Dennis del Castillo and Malaverri, Lizardo Fachin and Marin, Jhon Ever Rengifo},
  publisher = {AmeriFlux AMP},
  doi = {10.17190/AMF/1671889},
  note = {Dataset},
  site = {PE-QFR}
}

@misc{PR-xGU,
  title = {{AmeriFlux BASE PR-xGU NEON Guanica Forest (GUAN), Ver. 6-5}},
  year = {2023},
  author = {{NEON (National Ecological Observatory Network)}},
  publisher = {AmeriFlux AMP},
  doi = {10.17190/AMF/1773393},
  note = {Dataset},
  site = {PR-xGU}
}

@misc{PR-xLA,
  title = {{AmeriFlux BASE PR-xLA NEON Lajas Experimental Station (LAJA), Ver. 6-5}},
  year = {2023},
  author = {{NEON (National Ecological Observatory Network)}},
  publisher = {AmeriFlux AMP},
  doi = {10.17190/AMF/1773394},
  note = {Dataset},
  site = {PR-xLA}
}

@misc{US-A32,
  title = {{AmeriFlux BASE US-A32 ARM-SGP Medford hay pasture, Ver. 1-5}},
  year = {2018},
  author = {Billesbach, Dave and Kueppers, Lara and Torn, Margaret and Biraud, Sebastien},
  publisher = {AmeriFlux AMP},
  doi = {10.17190/AMF/1436327},
  note = {Dataset},
  site = {US-A32}
}

@misc{US-A74,
  title = {{AmeriFlux BASE US-A74 ARM SGP milo field, Ver. 1-5}},
  year = {2018},
  author = {Billesbach, Dave and Kueppers, Lara and Torn, Margaret and Biraud, Sebastien},
  publisher = {AmeriFlux AMP},
  doi = {10.17190/AMF/1436328},
  note = {Dataset},
  site = {US-A74}
}

@misc{US-ALQ,
  title = {{AmeriFlux BASE US-ALQ Allequash Creek Site, Ver. 15-5}},
  year = {2023},
  author = {Olson, Brent},
  publisher = {AmeriFlux AMP},
  doi = {10.17190/AMF/1480323},
  note = {Dataset},
  site = {US-ALQ}
}

@misc{US-ARM,
  title = {{AmeriFlux BASE US-ARM ARM Southern Great Plains site- Lamont, Ver. 12-5}},
  year = {2023},
  author = {Biraud, Sebastien and Fischer, Marc and Chan, Stephen and Torn, Margaret},
  publisher = {AmeriFlux AMP},
  doi = {10.17190/AMF/1246027},
  note = {Dataset},
  site = {US-ARM}
}

@misc{US-ASH,
  title = {{AmeriFlux BASE US-ASH USSL San Joaquin Valley Almond High Salinity, Ver. 1-5}},
  year = {2020},
  author = {Anderson, Ray G.},
  publisher = {AmeriFlux AMP},
  doi = {10.17190/AMF/1634880},
  note = {Dataset},
  site = {US-ASH}
}

@misc{US-ASL,
  title = {{AmeriFlux BASE US-ASL USSL San Joaquin Valley Almond Low Salinity, Ver. 1-5}},
  year = {2020},
  author = {Anderson, Ray G.},
  publisher = {AmeriFlux AMP},
  doi = {10.17190/AMF/1617706},
  note = {Dataset},
  site = {US-ASL}
}

@misc{US-ASM,
  title = {{AmeriFlux BASE US-ASM USSL San Joaquin Valley Almond Medium Salinity, Ver. 1-5}},
  year = {2020},
  author = {Anderson, Ray G.},
  publisher = {AmeriFlux AMP},
  doi = {10.17190/AMF/1617709},
  note = {Dataset},
  site = {US-ASM}
}

@misc{US-An1,
  title = {{AmeriFlux BASE US-An1 Anaktuvuk River Severe Burn, Ver. 2-5}},
  year = {2020},
  author = {Rocha, Adrian and Shaver, Gaius and Hobbie, John},
  publisher = {AmeriFlux AMP},
  doi = {10.17190/AMF/1246142},
  note = {Dataset},
  site = {US-An1}
}

@misc{US-An2,
  title = {{AmeriFlux BASE US-An2 Anaktuvuk River Moderate Burn, Ver. 2-5}},
  year = {2020},
  author = {Rocha, Adrian and Shaver, Gaius and Hobbie, John},
  publisher = {AmeriFlux AMP},
  doi = {10.17190/AMF/1246143},
  note = {Dataset},
  site = {US-An2}
}

@misc{US-An3,
  title = {{AmeriFlux BASE US-An3 Anaktuvuk River Unburned, Ver. 2-5}},
  year = {2020},
  author = {Rocha, Adrian and Shaver, Gaius and Hobbie, John},
  publisher = {AmeriFlux AMP},
  doi = {10.17190/AMF/1246144},
  note = {Dataset},
  site = {US-An3}
}

@misc{US-BRG,
  title = {{AmeriFlux BASE US-BRG Bayles Road Grassland Tower, Ver. 1-5}},
  year = {2020},
  author = {Novick, Kimberly},
  publisher = {AmeriFlux AMP},
  doi = {10.17190/AMF/1756416},
  note = {Dataset},
  site = {US-BRG}
}

@misc{US-BZB,
  title = {{AmeriFlux BASE US-BZB Bonanza Creek Thermokarst Bog, Ver. 4-5}},
  year = {2022},
  author = {Euskirchen, Eugenie},
  publisher = {AmeriFlux AMP},
  doi = {10.17190/AMF/1773401},
  note = {Dataset},
  site = {US-BZB}
}

@misc{US-BZF,
  title = {{AmeriFlux BASE US-BZF Bonanza Creek Rich Fen, Ver. 4-5}},
  year = {2022},
  author = {Euskirchen, Eugenie},
  publisher = {AmeriFlux AMP},
  doi = {10.17190/AMF/1756433},
  note = {Dataset},
  site = {US-BZF}
}

@misc{US-BZS,
  title = {{AmeriFlux BASE US-BZS Bonanza Creek Black Spruce, Ver. 3-5}},
  year = {2022},
  author = {Euskirchen, Eugenie},
  publisher = {AmeriFlux AMP},
  doi = {10.17190/AMF/1756434},
  note = {Dataset},
  site = {US-BZS}
}

@misc{US-BZo,
  title = {{AmeriFlux BASE US-BZo Bonanza Creek Old Thermokarst Bog, Ver. 3-5}},
  year = {2022},
  author = {Euskirchen, Eugenie},
  publisher = {AmeriFlux AMP},
  doi = {10.17190/AMF/1846662},
  note = {Dataset},
  site = {US-BZo}
}

@misc{US-Bar,
  title = {{AmeriFlux BASE US-Bar Bartlett Experimental Forest, Ver. 6-5}},
  year = {2023},
  author = {Richardson, Andrew and Hollinger, David},
  publisher = {AmeriFlux AMP},
  doi = {10.17190/AMF/1246030},
  note = {Dataset},
  site = {US-Bar}
}

@misc{US-Bi1,
  title = {{AmeriFlux BASE US-Bi1 Bouldin Island Alfalfa, Ver. 10-5}},
  year = {2023},
  author = {Rey-Sanchez, Camilo and Wang, Carlos Tianxin and Szutu, Daphne and Shortt, Robert and Chamberlain, Samuel D. and Verfaillie, Joseph and Baldocchi, Dennis},
  publisher = {AmeriFlux AMP},
  doi = {10.17190/AMF/1480317},
  note = {Dataset},
  site = {US-Bi1}
}

@misc{US-Bi2,
  title = {{AmeriFlux BASE US-Bi2 Bouldin Island corn, Ver. 15-5}},
  year = {2023},
  author = {Rey-Sanchez, Camilo and Wang, Carlos Tianxin and Szutu, Daphne and Hemes, Kyle and Verfaillie, Joseph and Baldocchi, Dennis},
  publisher = {AmeriFlux AMP},
  doi = {10.17190/AMF/1419513},
  note = {Dataset},
  site = {US-Bi2}
}

@misc{US-CF1,
  title = {{AmeriFlux BASE US-CF1 CAF-LTAR Cook East, Ver. 3-5}},
  year = {2022},
  author = {Phillips, Claire L. and Huggins, Dave},
  publisher = {AmeriFlux AMP},
  doi = {10.17190/AMF/1543382},
  note = {Dataset},
  site = {US-CF1}
}

@misc{US-CF2,
  title = {{AmeriFlux BASE US-CF2 CAF-LTAR Cook West, Ver. 2-5}},
  year = {2021},
  author = {Huggins, Dave},
  publisher = {AmeriFlux AMP},
  doi = {10.17190/AMF/1543383},
  note = {Dataset},
  site = {US-CF2}
}

@misc{US-CF3,
  title = {{AmeriFlux BASE US-CF3 CAF-LTAR Boyd North, Ver. 3-5}},
  year = {2022},
  author = {Huggins, Dave},
  publisher = {AmeriFlux AMP},
  doi = {10.17190/AMF/1543385},
  note = {Dataset},
  site = {US-CF3}
}

@misc{US-CF4,
  title = {{AmeriFlux BASE US-CF4 CAF-LTAR Boyd South, Ver. 3-5}},
  year = {2022},
  author = {Huggins, Dave},
  publisher = {AmeriFlux AMP},
  doi = {10.17190/AMF/1543384},
  note = {Dataset},
  site = {US-CF4}
}

@misc{US-CGG,
  title = {{AmeriFlux BASE US-CGG Concord Grazed Grassland, Ver. 1-5}},
  year = {2023},
  author = {Oikawa, Patty},
  publisher = {AmeriFlux AMP},
  doi = {10.17190/AMF/1987600},
  note = {Dataset},
  site = {US-CGG}
}

@misc{US-CLF,
  title = {{AmeriFlux BASE US-CLF Cole Farm, Ver. 1-5}},
  year = {2023},
  author = {Davis, Kenneth J.},
  publisher = {AmeriFlux AMP},
  doi = {10.17190/AMF/1987601},
  note = {Dataset},
  site = {US-CLF}
}

@misc{US-CMW,
  title = {{AmeriFlux BASE US-CMW Charleston Mesquite Woodland, Ver. 2-5}},
  year = {2022},
  author = {Scott, Russell},
  publisher = {AmeriFlux AMP},
  doi = {10.17190/AMF/1660339},
  note = {Dataset},
  site = {US-CMW}
}

@misc{US-CPk,
  title = {{AmeriFlux BASE US-CPk Chimney Park, Ver. 2-1}},
  year = {2016},
  author = {Ewers, Brent and Bretfeld, Mario and Pendall, Elise},
  publisher = {AmeriFlux AMP},
  doi = {10.17190/AMF/1246150},
  note = {Dataset},
  site = {US-CPk}
}

@misc{US-CRK,
  title = {{AmeriFlux BASE US-CRK Davy Crockett National Forest, Ver. 2-5}},
  year = {2023},
  author = {Noormets, Asko},
  publisher = {AmeriFlux AMP},
  doi = {10.17190/AMF/2204055},
  note = {Dataset},
  site = {US-CRK}
}

@misc{US-CRT,
  title = {{AmeriFlux BASE US-CRT Curtice Walter-Berger cropland, Ver. 5-5}},
  year = {2021},
  author = {Chen, Jiquan and Chu, Housen},
  publisher = {AmeriFlux AMP},
  doi = {10.17190/AMF/1246156},
  note = {Dataset},
  site = {US-CRT}
}

@misc{US-CS1,
  title = {{AmeriFlux BASE US-CS1 Central Sands Irrigated Agricultural Field, Ver. 3-5}},
  year = {2023},
  author = {Desai, Ankur},
  publisher = {AmeriFlux AMP},
  doi = {10.17190/AMF/1617710},
  note = {Dataset},
  site = {US-CS1}
}

@misc{US-CS2,
  title = {{AmeriFlux BASE US-CS2 Tri county school Pine Forest, Ver. 5-5}},
  year = {2023},
  author = {Desai, Ankur},
  publisher = {AmeriFlux AMP},
  doi = {10.17190/AMF/1617711},
  note = {Dataset},
  site = {US-CS2}
}

@misc{US-CS3,
  title = {{AmeriFlux BASE US-CS3 Central Sands Irrigated Agricultural Field, Ver. 4-5}},
  year = {2023},
  author = {Desai, Ankur},
  publisher = {AmeriFlux AMP},
  doi = {10.17190/AMF/1617713},
  note = {Dataset},
  site = {US-CS3}
}

@misc{US-CS4,
  title = {{AmeriFlux BASE US-CS4 Central Sands Irrigated Agricultural Field, Ver. 4-5}},
  year = {2023},
  author = {Desai, Ankur},
  publisher = {AmeriFlux AMP},
  doi = {10.17190/AMF/1756417},
  note = {Dataset},
  site = {US-CS4}
}

@misc{US-CS5,
  title = {{AmeriFlux BASE US-CS5 Central Sands Irrigated Agricultural Field, Ver. 1-5}},
  year = {2022},
  author = {Desai, Ankur},
  publisher = {AmeriFlux AMP},
  doi = {10.17190/AMF/1846663},
  note = {Dataset},
  site = {US-CS5}
}

@misc{US-CS6,
  title = {{AmeriFlux BASE US-CS6 Central Sands Irrigated Agricultural Field, Ver. 1-5}},
  year = {2023},
  author = {Desai, Ankur},
  publisher = {AmeriFlux AMP},
  doi = {10.17190/AMF/2001297},
  note = {Dataset},
  site = {US-CS6}
}

@misc{US-CS8,
  title = {{AmeriFlux BASE US-CS8 Central Sands Irrigated Agricultural Field, Ver. 2-5}},
  year = {2023},
  author = {Desai, Ankur},
  publisher = {AmeriFlux AMP},
  doi = {10.17190/AMF/2001298},
  note = {Dataset},
  site = {US-CS8}
}

@misc{US-Ced,
  title = {{AmeriFlux BASE US-Ced Cedar Bridge, Ver. 7-1}},
  year = {2016},
  author = {Clark, Ken},
  publisher = {AmeriFlux AMP},
  doi = {10.17190/AMF/1246043},
  note = {Dataset},
  site = {US-Ced}
}

@misc{US-DFC,
  title = {{AmeriFlux BASE US-DFC US Dairy Forage Research Center, Prairie du Sac, Ver. 2-5}},
  year = {2023},
  author = {Duff, Alison and Desai, Ankur},
  publisher = {AmeriFlux AMP},
  doi = {10.17190/AMF/1660340},
  note = {Dataset},
  site = {US-DFC}
}

@misc{US-DFK,
  title = {{AmeriFlux BASE US-DFK Dairy Forage Research Center - Kernza, Ver. 1-5}},
  year = {2021},
  author = {Duff, Alison and Desai, Ankur and Risso, Valentin Picasso},
  publisher = {AmeriFlux AMP},
  doi = {10.17190/AMF/1825937},
  note = {Dataset},
  site = {US-DFK}
}

@misc{US-DPW,
  title = {{AmeriFlux BASE US-DPW Disney Wilderness Preserve Wetland, Ver. 1-5}},
  year = {2019},
  author = {Hinkle, Charles Ross and Bracho, Rosvel},
  publisher = {AmeriFlux AMP},
  doi = {10.17190/AMF/1562387},
  note = {Dataset},
  site = {US-DPW}
}

@misc{US-DS3,
  title = {{AmeriFlux BASE US-DS3 Staten Rice 1, Ver. 2-5}},
  year = {2023},
  author = {Schuppenhauer, Michael R. and Biraud, Sebastien C. and Deverel, Steve and Chan, Stephen},
  publisher = {AmeriFlux AMP},
  doi = {10.17190/AMF/1890490},
  note = {Dataset},
  site = {US-DS3}
}

@misc{US-Dmg,
  title = {{AmeriFlux BASE US-Dmg Dutch Slough Marsh Gilbert Tract, Ver. 1-5}},
  year = {2023},
  author = {Arias-Ortiz, Ariane and Baldocchi, Dennis},
  publisher = {AmeriFlux AMP},
  doi = {10.17190/AMF/1964086},
  note = {Dataset},
  site = {US-Dmg}
}

@misc{US-EDN,
  title = {{AmeriFlux BASE US-EDN Eden Landing Ecological Reserve, Ver. 2-5}},
  year = {2020},
  author = {Oikawa, Patty},
  publisher = {AmeriFlux AMP},
  doi = {10.17190/AMF/1543381},
  note = {Dataset},
  site = {US-EDN}
}

@misc{US-Elm,
  title = {{AmeriFlux BASE US-Elm Everglades (long hydroperiod marsh), Ver. 4-1}},
  year = {2016},
  author = {Starr, Gregory and Oberbauer, Steve},
  publisher = {AmeriFlux AMP},
  doi = {10.17190/AMF/1246118},
  note = {Dataset},
  site = {US-Elm}
}

@misc{US-Esm,
  title = {{AmeriFlux BASE US-Esm Everglades (short hydroperiod marsh), Ver. 5-1}},
  year = {2016},
  author = {Starr, Gregory and Oberbauer, Steve},
  publisher = {AmeriFlux AMP},
  doi = {10.17190/AMF/1246119},
  note = {Dataset},
  site = {US-Esm}
}

@misc{US-Fcr,
  title = {{AmeriFlux BASE US-Fcr Cascaden Ridge Fire Scar, Ver. 3-5}},
  year = {2023},
  author = {Ueyama, Masahito and Iwata, Hiroki and Harazono, Yoshinobu},
  publisher = {AmeriFlux AMP},
  doi = {10.17190/AMF/1562388},
  note = {Dataset},
  site = {US-Fcr}
}

@misc{US-GL1,
  title = {{AmeriFlux BASE US-GL1 Stannard Rock, Ver. 2-5}},
  year = {2023},
  author = {Spence, Chris},
  publisher = {AmeriFlux AMP},
  doi = {10.17190/AMF/2204057},
  note = {Dataset},
  site = {US-GL1}
}

@misc{US-GLE,
  title = {{AmeriFlux BASE US-GLE GLEES, Ver. 8-5}},
  year = {2021},
  author = {Frank, John and Massman, Bill},
  publisher = {AmeriFlux AMP},
  doi = {10.17190/AMF/1246056},
  note = {Dataset},
  site = {US-GLE}
}

@misc{US-HB1,
  title = {{AmeriFlux BASE US-HB1 North Inlet Crab Haul Creek, Ver. 3-5}},
  year = {2023},
  author = {Forsythe, Jeremy D. and Kline, Michael A. and O'Halloran, Thomas L.},
  publisher = {AmeriFlux AMP},
  doi = {10.17190/AMF/1660341},
  note = {Dataset},
  site = {US-HB1}
}

@misc{US-HB2,
  title = {{AmeriFlux BASE US-HB2 Hobcaw Barony Mature Longleaf Pine, Ver. 1-5}},
  year = {2020},
  author = {Forsythe, Jeremy D. and Kline, Michael A. and O'Halloran, Thomas L.},
  publisher = {AmeriFlux AMP},
  doi = {10.17190/AMF/1660342},
  note = {Dataset},
  site = {US-HB2}
}

@misc{US-HB3,
  title = {{AmeriFlux BASE US-HB3 Hobcaw Barony Longleaf Pine Restoration, Ver. 2-5}},
  year = {2023},
  author = {Forsythe, Jeremy D. and Kline, Michael A. and O'Halloran, Thomas L.},
  publisher = {AmeriFlux AMP},
  doi = {10.17190/AMF/1660343},
  note = {Dataset},
  site = {US-HB3}
}

@misc{US-HBK,
  title = {{AmeriFlux BASE US-HBK Hubbard Brook Experimental Forest, Ver. 2-5}},
  year = {2023},
  author = {Kelsey, Eric and Green, Mark},
  publisher = {AmeriFlux AMP},
  doi = {10.17190/AMF/1634881},
  note = {Dataset},
  site = {US-HBK}
}

@misc{US-HRA,
  title = {{AmeriFlux BASE US-HRA Humnoke Farm Rice Field – Field A, Ver. 3-5}},
  year = {2021},
  author = {Runkle, Benjamin R. K.},
  publisher = {AmeriFlux AMP},
  doi = {10.17190/AMF/1543376},
  note = {Dataset},
  site = {US-HRA}
}

@misc{US-HRC,
  title = {{AmeriFlux BASE US-HRC Humnoke Farm Rice Field – Field C, Ver. 3-5}},
  year = {2021},
  author = {Reba, Michele L.},
  publisher = {AmeriFlux AMP},
  doi = {10.17190/AMF/1543375},
  note = {Dataset},
  site = {US-HRC}
}

@misc{US-Hn2,
  title = {{AmeriFlux BASE US-Hn2 Hanford 100H grassland, Ver. 1-5}},
  year = {2019},
  author = {Liu, Heping and Huang, Maoyi and Chen, Xingyuan},
  publisher = {AmeriFlux AMP},
  doi = {10.17190/AMF/1562389},
  note = {Dataset},
  site = {US-Hn2}
}

@misc{US-Hn3,
  title = {{AmeriFlux BASE US-Hn3 Hanford 100H sagebrush, Ver. 1-5}},
  year = {2019},
  author = {Liu, Heping and Huang, Maoyi and Chen, Xingyuan},
  publisher = {AmeriFlux AMP},
  doi = {10.17190/AMF/1543379},
  note = {Dataset},
  site = {US-Hn3}
}

@misc{US-Ho1,
  title = {{AmeriFlux BASE US-Ho1 Howland Forest (main tower), Ver. 7-5}},
  year = {2021},
  author = {Hollinger, David},
  publisher = {AmeriFlux AMP},
  doi = {10.17190/AMF/1246061},
  note = {Dataset},
  site = {US-Ho1}
}

@misc{US-Ho2,
  title = {{AmeriFlux BASE US-Ho2 Howland Forest (west tower), Ver. 4-5}},
  year = {2021},
  author = {Hollinger, David},
  publisher = {AmeriFlux AMP},
  doi = {10.17190/AMF/1246062},
  note = {Dataset},
  site = {US-Ho2}
}

@misc{US-Hsm,
  title = {{AmeriFlux BASE US-Hsm Hill Slough Marsh, Ver. 3-5}},
  year = {2023},
  author = {Arias-Ortiz, Ariane and Szutu, Daphne and Verfaillie, Joseph and Baldocchi, Dennis},
  publisher = {AmeriFlux AMP},
  doi = {10.17190/AMF/1890483},
  note = {Dataset},
  site = {US-Hsm}
}

@misc{US-IB1,
  title = {{AmeriFlux BASE US-IB1 Fermi National Accelerator Laboratory- Batavia (Agricultural site), Ver. 8-5}},
  year = {2019},
  author = {Matamala, Roser},
  publisher = {AmeriFlux AMP},
  doi = {10.17190/AMF/1246065},
  note = {Dataset},
  site = {US-IB1}
}

@misc{US-IB2,
  title = {{AmeriFlux BASE US-IB2 Fermi National Accelerator Laboratory- Batavia (Prairie site), Ver. 8-5}},
  year = {2019},
  author = {Matamala, Roser},
  publisher = {AmeriFlux AMP},
  doi = {10.17190/AMF/1246066},
  note = {Dataset},
  site = {US-IB2}
}

@misc{US-ICh,
  title = {{AmeriFlux BASE US-ICh Imnavait Creek Watershed Heath Tundra, Ver. 4-5}},
  year = {2022},
  author = {Euskirchen, Eugenie and Shaver, Gaius and Bret-Harte, Syndonia},
  publisher = {AmeriFlux AMP},
  doi = {10.17190/AMF/1246133},
  note = {Dataset},
  site = {US-ICh}
}

@misc{US-ICs,
  title = {{AmeriFlux BASE US-ICs Imnavait Creek Watershed Wet Sedge Tundra, Ver. 7-5}},
  year = {2022},
  author = {Euskirchen, Eugenie and Shaver, Gaius and Bret-Harte, Syndonia},
  publisher = {AmeriFlux AMP},
  doi = {10.17190/AMF/1246130},
  note = {Dataset},
  site = {US-ICs}
}

@misc{US-ICt,
  title = {{AmeriFlux BASE US-ICt Imnavait Creek Watershed Tussock Tundra, Ver. 5-5}},
  year = {2022},
  author = {Euskirchen, Eugenie and Shaver, Gaius and Bret-Harte, Syndonia},
  publisher = {AmeriFlux AMP},
  doi = {10.17190/AMF/1246131},
  note = {Dataset},
  site = {US-ICt}
}

@misc{US-Jo2,
  title = {{AmeriFlux BASE US-Jo2 Jornada Experimental Range Mixed Shrubland, Ver. 2-5}},
  year = {2022},
  author = {Vivoni, Enrique R. and Perez-Ruiz, Eli R.},
  publisher = {AmeriFlux AMP},
  doi = {10.17190/AMF/1617696},
  note = {Dataset},
  site = {US-Jo2}
}

@misc{US-KFS,
  title = {{AmeriFlux BASE US-KFS Kansas Field Station, Ver. 7-5}},
  year = {2020},
  author = {Brunsell, Nathaniel},
  publisher = {AmeriFlux AMP},
  doi = {10.17190/AMF/1246132},
  note = {Dataset},
  site = {US-KFS}
}

@misc{US-KLS,
  title = {{AmeriFlux BASE US-KLS Kansas Land Institute, Ver. 2-5}},
  year = {2021},
  author = {Brunsell, Nathaniel},
  publisher = {AmeriFlux AMP},
  doi = {10.17190/AMF/1498745},
  note = {Dataset},
  site = {US-KLS}
}

@misc{US-KPL,
  title = {{AmeriFlux BASE US-KPL Lily Lake Fen, Ver. 2-5}},
  year = {2023},
  author = {Sullivan, Patrick},
  publisher = {AmeriFlux AMP},
  doi = {10.17190/AMF/1865478},
  note = {Dataset},
  site = {US-KPL}
}

@misc{US-KS3,
  title = {{AmeriFlux BASE US-KS3 Kennedy Space Center (salt marsh), Ver. 1-5}},
  year = {2019},
  author = {Bracho, Rosvel and Hinkle, Charles Ross},
  publisher = {AmeriFlux AMP},
  doi = {10.17190/AMF/1562390},
  note = {Dataset},
  site = {US-KS3}
}

@misc{US-Kon,
  title = {{AmeriFlux BASE US-Kon Konza Prairie LTER (KNZ), Ver. 5-5}},
  year = {2020},
  author = {Brunsell, Nathaniel},
  publisher = {AmeriFlux AMP},
  doi = {10.17190/AMF/1246068},
  note = {Dataset},
  site = {US-Kon}
}

@misc{US-MOz,
  title = {{AmeriFlux BASE US-MOz Missouri Ozark Site, Ver. 11-5}},
  year = {2022},
  author = {Wood, Jeffrey and Gu, Lianhong},
  publisher = {AmeriFlux AMP},
  doi = {10.17190/AMF/1246081},
  note = {Dataset},
  site = {US-MOz}
}

@misc{US-Me2,
  title = {{AmeriFlux BASE US-Me2 Metolius mature ponderosa pine, Ver. 18-5}},
  year = {2022},
  author = {Law, Bev},
  publisher = {AmeriFlux AMP},
  doi = {10.17190/AMF/1246076},
  note = {Dataset},
  site = {US-Me2}
}

@misc{US-Me6,
  title = {{AmeriFlux BASE US-Me6 Metolius Young Pine Burn, Ver. 16-5}},
  year = {2023},
  author = {Law, Bev},
  publisher = {AmeriFlux AMP},
  doi = {10.17190/AMF/1246128},
  note = {Dataset},
  site = {US-Me6}
}

@misc{US-Men,
  title = {{AmeriFlux BASE US-Men Lake Mendota, Center for Limnology Site, Ver. 3-5}},
  year = {2018},
  author = {Desai, Ankur},
  publisher = {AmeriFlux AMP},
  doi = {10.17190/AMF/1433375},
  note = {Dataset},
  site = {US-Men}
}

@misc{US-Mo1,
  title = {{AmeriFlux BASE US-Mo1 LTAR CMRB Field 1 (CMRB ASP), Ver. 2-5}},
  year = {2023},
  author = {Schreiner-McGraw, Adam},
  publisher = {AmeriFlux AMP},
  doi = {10.17190/AMF/1870588},
  note = {Dataset},
  site = {US-Mo1}
}

@misc{US-Mo2,
  title = {{AmeriFlux BASE US-Mo2 LTAR CMRB Tucker Prairie (CMRB TP), Ver. 2-5}},
  year = {2023},
  author = {Schreiner-McGraw, Adam},
  publisher = {AmeriFlux AMP},
  doi = {10.17190/AMF/1902276},
  note = {Dataset},
  site = {US-Mo2}
}

@misc{US-Mo3,
  title = {{AmeriFlux BASE US-Mo3 LTAR CMRB Field 3 (CMRB BAU), Ver. 2-5}},
  year = {2023},
  author = {Schreiner-McGraw, Adam},
  publisher = {AmeriFlux AMP},
  doi = {10.17190/AMF/1870589},
  note = {Dataset},
  site = {US-Mo3}
}

@misc{US-Mpj,
  title = {{AmeriFlux BASE US-Mpj Mountainair Pinyon-Juniper Woodland, Ver. 22-5}},
  year = {2023},
  author = {Litvak, Marcy},
  publisher = {AmeriFlux AMP},
  doi = {10.17190/AMF/1246123},
  note = {Dataset},
  site = {US-Mpj}
}

@misc{US-Myb,
  title = {{AmeriFlux BASE US-Myb Mayberry Wetland, Ver. 13-5}},
  year = {2022},
  author = {Matthes, Jaclyn Hatala and Sturtevant, Cove and Oikawa, Patty and Chamberlain, Samuel D and Szutu, Daphne and Arias-Ortiz, Ariane and Verfaillie, Joseph and Baldocchi, Dennis},
  publisher = {AmeriFlux AMP},
  doi = {10.17190/AMF/1246139},
  note = {Dataset},
  site = {US-Myb}
}

@misc{US-NC2,
  title = {{AmeriFlux BASE US-NC2 NC\_Loblolly Plantation, Ver. 11-5}},
  year = {2023},
  author = {Noormets, Asko and Sun, Ge and Gavazzi, Michael and Domec, Jean-Christophe and McNulty, Steve and Miao, Guofang and Aguilos, Maricar and Mitra, Bhaskar and Minick, Kevan and King, John and Yang, Linqing and Prajapati, Prajaya},
  publisher = {AmeriFlux AMP},
  doi = {10.17190/AMF/1246083},
  note = {Dataset},
  site = {US-NC2}
}

@misc{US-NC3,
  title = {{AmeriFlux BASE US-NC3 NC\_Clearcut\#3, Ver. 4-5}},
  year = {2022},
  author = {Noormets, Asko and Gavazzi, Michael and Aguilos, Maricar and King, John and Mitra, Bhaskar and Domec, Jean-Christophe},
  publisher = {AmeriFlux AMP},
  doi = {10.17190/AMF/1419506},
  note = {Dataset},
  site = {US-NC3}
}

@misc{US-NC4,
  title = {{AmeriFlux BASE US-NC4 NC\_AlligatorRiver, Ver. 5-5}},
  year = {2022},
  author = {Noormets, Asko and King, John and Mitra, Bhaskar and Miao, Guofang and Aguilos, Maricar and Minick, Kevan and Prajapati, Prajaya and Domec, Jean-Christophe},
  publisher = {AmeriFlux AMP},
  doi = {10.17190/AMF/1480314},
  note = {Dataset},
  site = {US-NC4}
}

@misc{US-NGB,
  title = {{AmeriFlux BASE US-NGB NGEE Arctic Barrow, Ver. 5-5}},
  year = {2023},
  author = {Torn, Margaret and Dengel, Sigrid},
  publisher = {AmeriFlux AMP},
  doi = {10.17190/AMF/1436326},
  note = {Dataset},
  site = {US-NGB}
}

@misc{US-NGC,
  title = {{AmeriFlux BASE US-NGC NGEE Arctic Council, Ver. 3-5}},
  year = {2023},
  author = {Torn, Margaret and Dengel, Sigrid},
  publisher = {AmeriFlux AMP},
  doi = {10.17190/AMF/1634883},
  note = {Dataset},
  site = {US-NGC}
}

@misc{US-NR1,
  title = {{AmeriFlux BASE US-NR1 Niwot Ridge Forest (LTER NWT1), Ver. 20-5}},
  year = {2023},
  author = {Blanken, Peter D. and Monson, Russel K. and Burns, Sean P. and Bowling, David R. and Turnipseed, Andrew A.},
  publisher = {AmeriFlux AMP},
  doi = {10.17190/AMF/1246088},
  note = {Dataset},
  site = {US-NR1}
}

@misc{US-NR3,
  title = {{AmeriFlux BASE US-NR3 Niwot Ridge Alpine (T-Van West), Ver. 4-5}},
  year = {2023},
  author = {Knowles, John},
  publisher = {AmeriFlux AMP},
  doi = {10.17190/AMF/1804491},
  note = {Dataset},
  site = {US-NR3}
}

@misc{US-NR4,
  title = {{AmeriFlux BASE US-NR4 Niwot Ridge Alpine (T-Van East), Ver. 4-5}},
  year = {2023},
  author = {Knowles, John},
  publisher = {AmeriFlux AMP},
  doi = {10.17190/AMF/1804492},
  note = {Dataset},
  site = {US-NR4}
}

@misc{US-ONA,
  title = {{AmeriFlux BASE US-ONA Florida pine flatwoods, Ver. 3-5}},
  year = {2022},
  author = {Silveira, Maria L. and Bracho, Rosvel},
  publisher = {AmeriFlux AMP},
  doi = {10.17190/AMF/1660350},
  note = {Dataset},
  site = {US-ONA}
}

@misc{US-ORv,
  title = {{AmeriFlux BASE US-ORv Olentangy River Wetland Research Park, Ver. 3-5}},
  year = {2020},
  author = {Bohrer, Gil},
  publisher = {AmeriFlux AMP},
  doi = {10.17190/AMF/1246135},
  note = {Dataset},
  site = {US-ORv}
}

@misc{US-OWC,
  title = {{AmeriFlux BASE US-OWC Old Woman Creek, Ver. 4-5}},
  year = {2023},
  author = {Bohrer, Gil and Kerns, Janice},
  publisher = {AmeriFlux AMP},
  doi = {10.17190/AMF/1418679},
  note = {Dataset},
  site = {US-OWC}
}

@misc{US-PSH,
  title = {{AmeriFlux BASE US-PSH USSL San Joaquin Valley Pistachio High, Ver. 1-5}},
  year = {2020},
  author = {Anderson, Ray G.},
  publisher = {AmeriFlux AMP},
  doi = {10.17190/AMF/1617719},
  note = {Dataset},
  site = {US-PSH}
}

@misc{US-PSL,
  title = {{AmeriFlux BASE US-PSL USSL San Joaquin Valley Pistachio Low, Ver. 1-5}},
  year = {2020},
  author = {Anderson, Ray G.},
  publisher = {AmeriFlux AMP},
  doi = {10.17190/AMF/1617720},
  note = {Dataset},
  site = {US-PSL}
}

@misc{US-Pnp,
  title = {{AmeriFlux BASE US-Pnp Lake Mendota, Picnic Point Site, Ver. 8-5}},
  year = {2023},
  author = {Desai, Ankur},
  publisher = {AmeriFlux AMP},
  doi = {10.17190/AMF/1433376},
  note = {Dataset},
  site = {US-Pnp}
}

@misc{US-Prr,
  title = {{AmeriFlux BASE US-Prr Poker Flat Research Range Black Spruce Forest, Ver. 4-5}},
  year = {2023},
  author = {Iwahana, Go and Kobayashi, Hideki and Ikawa, Hiroki and Suzuki, Rikie},
  publisher = {AmeriFlux AMP},
  doi = {10.17190/AMF/1246153},
  note = {Dataset},
  site = {US-Prr}
}

@misc{US-RGA,
  title = {{AmeriFlux BASE US-RGA Arkansas Corn Farm, Ver. 3-5}},
  year = {2023},
  author = {Schuppenhauer, Michael R. and Biraud, Sebastien C. and Chan, Stephen},
  publisher = {AmeriFlux AMP},
  doi = {10.17190/AMF/1880913},
  note = {Dataset},
  site = {US-RGA}
}

@misc{US-RGB,
  title = {{AmeriFlux BASE US-RGB Butte County Rice Farm, Ver. 4-5}},
  year = {2023},
  author = {Schuppenhauer, Michael and Biraud, Sebastien C. and Chan, Stephen},
  publisher = {AmeriFlux AMP},
  doi = {10.17190/AMF/1870591},
  note = {Dataset},
  site = {US-RGB}
}

@misc{US-RGF,
  title = {{AmeriFlux BASE US-RGF Stanislaus County Forage Farm, Ver. 1-5}},
  year = {2023},
  author = {Schuppenhauer, Michael R. and Biraud, Sebastien C. and Chan, Stephen},
  publisher = {AmeriFlux AMP},
  doi = {10.17190/AMF/2001310},
  note = {Dataset},
  site = {US-RGF}
}

@misc{US-RGW,
  title = {{AmeriFlux BASE US-RGW Desha County Rice Farm, Ver. 3-5}},
  year = {2023},
  author = {Schuppenhauer, Michael R. and Biraud, Sebastien C. and Chan, Stephen},
  publisher = {AmeriFlux AMP},
  doi = {10.17190/AMF/1880915},
  note = {Dataset},
  site = {US-RGW}
}

@misc{US-RGo,
  title = {{AmeriFlux BASE US-RGo Glenn County Organic Rice Farm, Ver. 3-5}},
  year = {2023},
  author = {Schuppenhauer, Michael R. and Biraud, Sebastien C. and Chan, Stephen},
  publisher = {AmeriFlux AMP},
  doi = {10.17190/AMF/1880914},
  note = {Dataset},
  site = {US-RGo}
}

@misc{US-Rls,
  title = {{AmeriFlux BASE US-Rls RCEW Low Sagebrush, Ver. 5-5}},
  year = {2023},
  author = {Flerchinger, Gerald},
  publisher = {AmeriFlux AMP},
  doi = {10.17190/AMF/1418682},
  note = {Dataset},
  site = {US-Rls}
}

@misc{US-Rms,
  title = {{AmeriFlux BASE US-Rms RCEW Mountain Big Sagebrush, Ver. 5-5}},
  year = {2023},
  author = {Flerchinger, Gerald},
  publisher = {AmeriFlux AMP},
  doi = {10.17190/AMF/1375202},
  note = {Dataset},
  site = {US-Rms}
}

@misc{US-Ro1,
  title = {{AmeriFlux BASE US-Ro1 Rosemount- G21, Ver. 5-5}},
  year = {2018},
  author = {Baker, John and Griffis, Tim and Griffis, Timothy},
  publisher = {AmeriFlux AMP},
  doi = {10.17190/AMF/1246092},
  note = {Dataset},
  site = {US-Ro1}
}

@misc{US-Ro2,
  title = {{AmeriFlux BASE US-Ro2 Rosemount- C7, Ver. 2-5}},
  year = {2023},
  author = {Baker, John and Griffis, Tim},
  publisher = {AmeriFlux AMP},
  doi = {10.17190/AMF/1418683},
  note = {Dataset},
  site = {US-Ro2}
}

@misc{US-Ro4,
  title = {{AmeriFlux BASE US-Ro4 Rosemount Prairie, Ver. 22-5}},
  year = {2023},
  author = {Baker, John and Griffis, Tim},
  publisher = {AmeriFlux AMP},
  doi = {10.17190/AMF/1419507},
  note = {Dataset},
  site = {US-Ro4}
}

@misc{US-Ro5,
  title = {{AmeriFlux BASE US-Ro5 Rosemount I18\_South, Ver. 22-5}},
  year = {2023},
  author = {Baker, John and Griffis, Tim},
  publisher = {AmeriFlux AMP},
  doi = {10.17190/AMF/1419508},
  note = {Dataset},
  site = {US-Ro5}
}

@misc{US-Ro6,
  title = {{AmeriFlux BASE US-Ro6 Rosemount I18\_North, Ver. 22-5}},
  year = {2023},
  author = {Baker, John and Griffis, Tim},
  publisher = {AmeriFlux AMP},
  doi = {10.17190/AMF/1419509},
  note = {Dataset},
  site = {US-Ro6}
}

@misc{US-Rpf,
  title = {{AmeriFlux BASE US-Rpf Poker Flat Research Range: Succession from fire scar to deciduous forest, Ver. 9-5}},
  year = {2023},
  author = {Ueyama, Masahito and Iwata, Hiroki and Harazono, Yoshinobu},
  publisher = {AmeriFlux AMP},
  doi = {10.17190/AMF/1579540},
  note = {Dataset},
  site = {US-Rpf}
}

@misc{US-Rwf,
  title = {{AmeriFlux BASE US-Rwf RCEW Upper Sheep Prescibed Fire, Ver. 3-5}},
  year = {2023},
  author = {Flerchinger, Gerald},
  publisher = {AmeriFlux AMP},
  doi = {10.17190/AMF/1617724},
  note = {Dataset},
  site = {US-Rwf}
}

@misc{US-Rws,
  title = {{AmeriFlux BASE US-Rws Reynolds Creek Wyoming big sagebrush, Ver. 5-5}},
  year = {2023},
  author = {Flerchinger, Gerald},
  publisher = {AmeriFlux AMP},
  doi = {10.17190/AMF/1375201},
  note = {Dataset},
  site = {US-Rws}
}

@misc{US-SP1,
  title = {{AmeriFlux BASE US-SP1 Slashpine-Austin Cary- 65yrs nat regen, Ver. 5-5}},
  year = {2023},
  author = {Bracho, Rosvel and Martin, Timothy A.},
  publisher = {AmeriFlux AMP},
  doi = {10.17190/AMF/1246100},
  note = {Dataset},
  site = {US-SP1}
}

@misc{US-SRC,
  title = {{AmeriFlux BASE US-SRC Santa Rita Creosote, Ver. 6-5}},
  year = {2019},
  author = {Kurc, Shirley},
  publisher = {AmeriFlux AMP},
  doi = {10.17190/AMF/1246127},
  note = {Dataset},
  site = {US-SRC}
}

@misc{US-SRG,
  title = {{AmeriFlux BASE US-SRG Santa Rita Grassland, Ver. 15-5}},
  year = {2023},
  author = {Scott, Russell},
  publisher = {AmeriFlux AMP},
  doi = {10.17190/AMF/1246154},
  note = {Dataset},
  site = {US-SRG}
}

@misc{US-SRM,
  title = {{AmeriFlux BASE US-SRM Santa Rita Mesquite, Ver. 26-5}},
  year = {2023},
  author = {Scott, Russell},
  publisher = {AmeriFlux AMP},
  doi = {10.17190/AMF/1246104},
  note = {Dataset},
  site = {US-SRM}
}

@misc{US-SRS,
  title = {{AmeriFlux BASE US-SRS Santa Rita Savanna, Ver. 3-5}},
  year = {2022},
  author = {Vivoni, Enrique R.},
  publisher = {AmeriFlux AMP},
  doi = {10.17190/AMF/1660351},
  note = {Dataset},
  site = {US-SRS}
}

@misc{US-SSH,
  title = {{AmeriFlux BASE US-SSH Susquehanna Shale Hills Critical Zone Observatory, Ver. 1-5}},
  year = {2022},
  author = {Forsythe, Brandon R. and Horne, Jason and Davis, Kenneth J.},
  publisher = {AmeriFlux AMP},
  doi = {10.17190/AMF/1880916},
  note = {Dataset},
  site = {US-SSH}
}

@misc{US-Seg,
  title = {{AmeriFlux BASE US-Seg Sevilleta grassland, Ver. 23-5}},
  year = {2023},
  author = {Litvak, Marcy},
  publisher = {AmeriFlux AMP},
  doi = {10.17190/AMF/1246124},
  note = {Dataset},
  site = {US-Seg}
}

@misc{US-Ses,
  title = {{AmeriFlux BASE US-Ses Sevilleta shrubland, Ver. 22-5}},
  year = {2023},
  author = {Litvak, Marcy},
  publisher = {AmeriFlux AMP},
  doi = {10.17190/AMF/1246125},
  note = {Dataset},
  site = {US-Ses}
}

@misc{US-Slt,
  title = {{AmeriFlux BASE US-Slt Silas Little- New Jersey, Ver. 5-1}},
  year = {2016},
  author = {Clark, Ken},
  publisher = {AmeriFlux AMP},
  doi = {10.17190/AMF/1246096},
  note = {Dataset},
  site = {US-Slt}
}

@misc{US-Snd,
  title = {{AmeriFlux BASE US-Snd Sherman Island, Ver. 2-1}},
  year = {2016},
  author = {Detto, Matteo and Sturtevant, Cove and Oikawa, Patty and Verfaillie, Joseph and Baldocchi, Dennis},
  publisher = {AmeriFlux AMP},
  doi = {10.17190/AMF/1246094},
  note = {Dataset},
  site = {US-Snd}
}

@misc{US-Sne,
  title = {{AmeriFlux BASE US-Sne Sherman Island Restored Wetland, Ver. 7-5}},
  year = {2021},
  author = {Shortt, Robert and Hemes, Kyle and Szutu, Daphne and Verfaillie, Joseph and Baldocchi, Dennis},
  publisher = {AmeriFlux AMP},
  doi = {10.17190/AMF/1418684},
  note = {Dataset},
  site = {US-Sne}
}

@misc{US-Snf,
  title = {{AmeriFlux BASE US-Snf Sherman Barn, Ver. 3-5}},
  year = {2020},
  author = {Kusak, Kuno and Sanchez, Camilo Rey and Szutu, Daphne and Baldocchi, Dennis},
  publisher = {AmeriFlux AMP},
  doi = {10.17190/AMF/1579718},
  note = {Dataset},
  site = {US-Snf}
}

@misc{US-Srr,
  title = {{AmeriFlux BASE US-Srr Suisun marsh - Rush Ranch, Ver. 1-5}},
  year = {2018},
  author = {Bergamaschi, Brian and Windham-Myers, Lisamarie},
  publisher = {AmeriFlux AMP},
  doi = {10.17190/AMF/1418685},
  note = {Dataset},
  site = {US-Srr}
}

@misc{US-StJ,
  title = {{AmeriFlux BASE US-StJ St Jones Reserve, Ver. 2-5}},
  year = {2020},
  author = {Vargas, Rodrigo},
  publisher = {AmeriFlux AMP},
  doi = {10.17190/AMF/1480316},
  note = {Dataset},
  site = {US-StJ}
}

@misc{US-Syv,
  title = {{AmeriFlux BASE US-Syv Sylvania Wilderness Area, Ver. 26-5}},
  year = {2023},
  author = {Desai, Ankur},
  publisher = {AmeriFlux AMP},
  doi = {10.17190/AMF/1246106},
  note = {Dataset},
  site = {US-Syv}
}

@misc{US-Tw1,
  title = {{AmeriFlux BASE US-Tw1 Twitchell Wetland West Pond, Ver. 10-5}},
  year = {2023},
  author = {Valach, Alex and Shortt, Robert and Szutu, Daphne and Eichelmann, Elke and Knox, Sara and Hemes, Kyle and Verfaillie, Joseph and Baldocchi, Dennis},
  publisher = {AmeriFlux AMP},
  doi = {10.17190/AMF/1246147},
  note = {Dataset},
  site = {US-Tw1}
}

@misc{US-Tw2,
  title = {{AmeriFlux BASE US-Tw2 Twitchell Corn, Ver. 2-5}},
  year = {2019},
  author = {Sturtevant, Cove and Verfaillie, Joseph and Baldocchi, Dennis},
  publisher = {AmeriFlux AMP},
  doi = {10.17190/AMF/1246148},
  note = {Dataset},
  site = {US-Tw2}
}

@misc{US-Tw3,
  title = {{AmeriFlux BASE US-Tw3 Twitchell Alfalfa, Ver. 5-5}},
  year = {2018},
  author = {Chamberlain, Samuel D and Oikawa, Patty and Sturtevant, Cove and Szutu, Daphne and Verfaillie, Joseph and Baldocchi, Dennis},
  publisher = {AmeriFlux AMP},
  doi = {10.17190/AMF/1246149},
  note = {Dataset},
  site = {US-Tw3}
}

@misc{US-Tw4,
  title = {{AmeriFlux BASE US-Tw4 Twitchell East End Wetland, Ver. 13-5}},
  year = {2023},
  author = {Eichelmann, Elke and Shortt, Robert and Knox, Sara and Sanchez, Camilo Rey and Valach, Alex and Sturtevant, Cove and Szutu, Daphne and Verfaillie, Joseph and Baldocchi, Dennis},
  publisher = {AmeriFlux AMP},
  doi = {10.17190/AMF/1246151},
  note = {Dataset},
  site = {US-Tw4}
}

@misc{US-Tw5,
  title = {{AmeriFlux BASE US-Tw5 East Pond Wetland, Ver. 3-5}},
  year = {2020},
  author = {Valach, Alex and Kasak, Kuno and Szutu, Daphne and Verfaillie, Joseph and Baldocchi, Dennis},
  publisher = {AmeriFlux AMP},
  doi = {10.17190/AMF/1543380},
  note = {Dataset},
  site = {US-Tw5}
}

@misc{US-Twt,
  title = {{AmeriFlux BASE US-Twt Twitchell Island, Ver. 7-5}},
  year = {2023},
  author = {Knox, Sara and Matthes, Jaclyn Hatala and Verfaillie, Joseph and Baldocchi, Dennis},
  publisher = {AmeriFlux AMP},
  doi = {10.17190/AMF/1246140},
  note = {Dataset},
  site = {US-Twt}
}

@misc{US-UM3,
  title = {{AmeriFlux BASE US-UM3 Douglas Lake, Ver. 1-5}},
  year = {2018},
  author = {Bohrer, Gil},
  publisher = {AmeriFlux AMP},
  doi = {10.17190/AMF/1480315},
  note = {Dataset},
  site = {US-UM3}
}

@misc{US-UMB,
  title = {{AmeriFlux BASE US-UMB Univ. of Mich. Biological Station, Ver. 20-5}},
  year = {2023},
  author = {Gough, Christopher and Bohrer, Gil and Curtis, Peter},
  publisher = {AmeriFlux AMP},
  doi = {10.17190/AMF/1246107},
  note = {Dataset},
  site = {US-UMB}
}

@misc{US-Uaf,
  title = {{AmeriFlux BASE US-Uaf University of Alaska, Fairbanks, Ver. 11-5}},
  year = {2023},
  author = {Ueyama, Masahito and Iwata, Hiroki and Harazono, Yoshinobu},
  publisher = {AmeriFlux AMP},
  doi = {10.17190/AMF/1480322},
  note = {Dataset},
  site = {US-Uaf}
}

@misc{US-Var,
  title = {{AmeriFlux BASE US-Var Vaira Ranch- Ione, Ver. 19-5}},
  year = {2023},
  author = {Ma, Siyan and Xu, Liukang and Verfaillie, Joseph and Baldocchi, Dennis},
  publisher = {AmeriFlux AMP},
  doi = {10.17190/AMF/1245984},
  note = {Dataset},
  site = {US-Var}
}

@misc{US-Vcm,
  title = {{AmeriFlux BASE US-Vcm Valles Caldera Mixed Conifer, Ver. 24-5}},
  year = {2023},
  author = {Litvak, Marcy},
  publisher = {AmeriFlux AMP},
  doi = {10.17190/AMF/1246121},
  note = {Dataset},
  site = {US-Vcm}
}

@misc{US-Vcp,
  title = {{AmeriFlux BASE US-Vcp Valles Caldera Ponderosa Pine, Ver. 22-5}},
  year = {2023},
  author = {Litvak, Marcy},
  publisher = {AmeriFlux AMP},
  doi = {10.17190/AMF/1246122},
  note = {Dataset},
  site = {US-Vcp}
}

@misc{US-WPT,
  title = {{AmeriFlux BASE US-WPT Winous Point North Marsh, Ver. 4-5}},
  year = {2019},
  author = {Chen, Jiquan and Chu, Housen},
  publisher = {AmeriFlux AMP},
  doi = {10.17190/AMF/1246155},
  note = {Dataset},
  site = {US-WPT}
}

@misc{US-Whs,
  title = {{AmeriFlux BASE US-Whs Walnut Gulch Lucky Hills Shrub, Ver. 21-5}},
  year = {2023},
  author = {Scott, Russ},
  publisher = {AmeriFlux AMP},
  doi = {10.17190/AMF/1246113},
  note = {Dataset},
  site = {US-Whs}
}

@misc{US-Wjs,
  title = {{AmeriFlux BASE US-Wjs Willard Juniper Savannah, Ver. 22-5}},
  year = {2023},
  author = {Litvak, Marcy},
  publisher = {AmeriFlux AMP},
  doi = {10.17190/AMF/1246120},
  note = {Dataset},
  site = {US-Wjs}
}

@misc{US-Wkg,
  title = {{AmeriFlux BASE US-Wkg Walnut Gulch Kendall Grasslands, Ver. 21-5}},
  year = {2023},
  author = {Scott, Russell},
  publisher = {AmeriFlux AMP},
  doi = {10.17190/AMF/1246112},
  note = {Dataset},
  site = {US-Wkg}
}

@misc{US-Wrc,
  title = {{AmeriFlux BASE US-Wrc Wind River Crane Site, Ver. 8-1}},
  year = {2016},
  author = {Wharton, Sonia},
  publisher = {AmeriFlux AMP},
  doi = {10.17190/AMF/1246114},
  note = {Dataset},
  site = {US-Wrc}
}

@misc{US-xAB,
  title = {{AmeriFlux BASE US-xAB NEON Abby Road (ABBY), Ver. 8-5}},
  year = {2023},
  author = {{NEON (National Ecological Observatory Network)}},
  publisher = {AmeriFlux AMP},
  doi = {10.17190/AMF/1617726},
  note = {Dataset},
  site = {US-xAB}
}

@misc{US-xAE,
  title = {{AmeriFlux BASE US-xAE NEON Klemme Range Research Station (OAES), Ver. 7-5}},
  year = {2023},
  author = {{NEON (National Ecological Observatory Network)}},
  publisher = {AmeriFlux AMP},
  doi = {10.17190/AMF/1671891},
  note = {Dataset},
  site = {US-xAE}
}

@misc{US-xBA,
  title = {{AmeriFlux BASE US-xBA NEON Barrow Environmental Observatory (BARR), Ver. 7-5}},
  year = {2023},
  author = {{NEON (National Ecological Observatory Network)}},
  publisher = {AmeriFlux AMP},
  doi = {10.17190/AMF/1671892},
  note = {Dataset},
  site = {US-xBA}
}

@misc{US-xBL,
  title = {{AmeriFlux BASE US-xBL NEON Blandy Experimental Farm (BLAN), Ver. 7-5}},
  year = {2023},
  author = {{NEON (National Ecological Observatory Network)}},
  publisher = {AmeriFlux AMP},
  doi = {10.17190/AMF/1671893},
  note = {Dataset},
  site = {US-xBL}
}

@misc{US-xBN,
  title = {{AmeriFlux BASE US-xBN NEON Caribou Creek - Poker Flats Watershed (BONA), Ver. 8-5}},
  year = {2023},
  author = {{NEON (National Ecological Observatory Network)}},
  publisher = {AmeriFlux AMP},
  doi = {10.17190/AMF/1617727},
  note = {Dataset},
  site = {US-xBN}
}

@misc{US-xBR,
  title = {{AmeriFlux BASE US-xBR NEON Bartlett Experimental Forest (BART), Ver. 8-5}},
  year = {2023},
  author = {{NEON (National Ecological Observatory Network)}},
  publisher = {AmeriFlux AMP},
  doi = {10.17190/AMF/1579542},
  note = {Dataset},
  site = {US-xBR}
}

@misc{US-xCL,
  title = {{AmeriFlux BASE US-xCL NEON LBJ National Grassland (CLBJ), Ver. 7-5}},
  year = {2023},
  author = {{NEON (National Ecological Observatory Network)}},
  publisher = {AmeriFlux AMP},
  doi = {10.17190/AMF/1671894},
  note = {Dataset},
  site = {US-xCL}
}

@misc{US-xCP,
  title = {{AmeriFlux BASE US-xCP NEON Central Plains Experimental Range (CPER), Ver. 8-5}},
  year = {2023},
  author = {{NEON (National Ecological Observatory Network)}},
  publisher = {AmeriFlux AMP},
  doi = {10.17190/AMF/1579720},
  note = {Dataset},
  site = {US-xCP}
}

@misc{US-xDC,
  title = {{AmeriFlux BASE US-xDC NEON Dakota Coteau Field School (DCFS), Ver. 8-5}},
  year = {2023},
  author = {{NEON (National Ecological Observatory Network)}},
  publisher = {AmeriFlux AMP},
  doi = {10.17190/AMF/1617728},
  note = {Dataset},
  site = {US-xDC}
}

@misc{US-xDJ,
  title = {{AmeriFlux BASE US-xDJ NEON Delta Junction (DEJU), Ver. 8-5}},
  year = {2023},
  author = {{NEON (National Ecological Observatory Network)}},
  publisher = {AmeriFlux AMP},
  doi = {10.17190/AMF/1634884},
  note = {Dataset},
  site = {US-xDJ}
}

@misc{US-xDL,
  title = {{AmeriFlux BASE US-xDL NEON Dead Lake (DELA), Ver. 8-5}},
  year = {2023},
  author = {{NEON (National Ecological Observatory Network)}},
  publisher = {AmeriFlux AMP},
  doi = {10.17190/AMF/1579721},
  note = {Dataset},
  site = {US-xDL}
}

@misc{US-xDS,
  title = {{AmeriFlux BASE US-xDS NEON Disney Wilderness Preserve (DSNY), Ver. 7-5}},
  year = {2023},
  author = {{NEON (National Ecological Observatory Network)}},
  publisher = {AmeriFlux AMP},
  doi = {10.17190/AMF/1671895},
  note = {Dataset},
  site = {US-xDS}
}

@misc{US-xGR,
  title = {{AmeriFlux BASE US-xGR NEON Great Smoky Mountains National Park, Twin Creeks (GRSM), Ver. 8-5}},
  year = {2023},
  author = {{NEON (National Ecological Observatory Network)}},
  publisher = {AmeriFlux AMP},
  doi = {10.17190/AMF/1634885},
  note = {Dataset},
  site = {US-xGR}
}

@misc{US-xHA,
  title = {{AmeriFlux BASE US-xHA NEON Harvard Forest (HARV), Ver. 9-5}},
  year = {2023},
  author = {{NEON (National Ecological Observatory Network)}},
  publisher = {AmeriFlux AMP},
  doi = {10.17190/AMF/1562391},
  note = {Dataset},
  site = {US-xHA}
}

@misc{US-xHE,
  title = {{AmeriFlux BASE US-xHE NEON Healy (HEAL), Ver. 8-5}},
  year = {2023},
  author = {{NEON (National Ecological Observatory Network)}},
  publisher = {AmeriFlux AMP},
  doi = {10.17190/AMF/1617729},
  note = {Dataset},
  site = {US-xHE}
}

@misc{US-xJE,
  title = {{AmeriFlux BASE US-xJE NEON Jones Ecological Research Center (JERC), Ver. 8-5}},
  year = {2023},
  author = {{NEON (National Ecological Observatory Network)}},
  publisher = {AmeriFlux AMP},
  doi = {10.17190/AMF/1617730},
  note = {Dataset},
  site = {US-xJE}
}

@misc{US-xJR,
  title = {{AmeriFlux BASE US-xJR NEON Jornada LTER (JORN), Ver. 8-5}},
  year = {2023},
  author = {{NEON (National Ecological Observatory Network)}},
  publisher = {AmeriFlux AMP},
  doi = {10.17190/AMF/1617731},
  note = {Dataset},
  site = {US-xJR}
}

@misc{US-xKA,
  title = {{AmeriFlux BASE US-xKA NEON Konza Prairie Biological Station - Relocatable (KONA), Ver. 8-5}},
  year = {2023},
  author = {{NEON (National Ecological Observatory Network)}},
  publisher = {AmeriFlux AMP},
  doi = {10.17190/AMF/1579722},
  note = {Dataset},
  site = {US-xKA}
}

@misc{US-xKZ,
  title = {{AmeriFlux BASE US-xKZ NEON Konza Prairie Biological Station (KONZ), Ver. 9-5}},
  year = {2023},
  author = {{NEON (National Ecological Observatory Network)}},
  publisher = {AmeriFlux AMP},
  doi = {10.17190/AMF/1562392},
  note = {Dataset},
  site = {US-xKZ}
}

@misc{US-xLE,
  title = {{AmeriFlux BASE US-xLE NEON Lenoir Landing (LENO), Ver. 6-5}},
  year = {2023},
  author = {{NEON (National Ecological Observatory Network)}},
  publisher = {AmeriFlux AMP},
  doi = {10.17190/AMF/1773398},
  note = {Dataset},
  site = {US-xLE}
}

@misc{US-xMB,
  title = {{AmeriFlux BASE US-xMB NEON Moab (MOAB), Ver. 7-5}},
  year = {2023},
  author = {{NEON (National Ecological Observatory Network)}},
  publisher = {AmeriFlux AMP},
  doi = {10.17190/AMF/1671896},
  note = {Dataset},
  site = {US-xMB}
}

@misc{US-xML,
  title = {{AmeriFlux BASE US-xML NEON Mountain Lake Biological Station (MLBS), Ver. 7-5}},
  year = {2023},
  author = {{NEON (National Ecological Observatory Network)}},
  publisher = {AmeriFlux AMP},
  doi = {10.17190/AMF/1671897},
  note = {Dataset},
  site = {US-xML}
}

@misc{US-xNG,
  title = {{AmeriFlux BASE US-xNG NEON Northern Great Plains Research Laboratory (NOGP), Ver. 8-5}},
  year = {2023},
  author = {{NEON (National Ecological Observatory Network)}},
  publisher = {AmeriFlux AMP},
  doi = {10.17190/AMF/1617732},
  note = {Dataset},
  site = {US-xNG}
}

@misc{US-xNQ,
  title = {{AmeriFlux BASE US-xNQ NEON Onaqui-Ault (ONAQ), Ver. 8-5}},
  year = {2023},
  author = {{NEON (National Ecological Observatory Network)}},
  publisher = {AmeriFlux AMP},
  doi = {10.17190/AMF/1617733},
  note = {Dataset},
  site = {US-xNQ}
}

@misc{US-xNW,
  title = {{AmeriFlux BASE US-xNW NEON Niwot Ridge Mountain Research Station (NIWO), Ver. 7-5}},
  year = {2023},
  author = {{NEON (National Ecological Observatory Network)}},
  publisher = {AmeriFlux AMP},
  doi = {10.17190/AMF/1671898},
  note = {Dataset},
  site = {US-xNW}
}

@misc{US-xRM,
  title = {{AmeriFlux BASE US-xRM NEON Rocky Mountain National Park, CASTNET (RMNP), Ver. 8-5}},
  year = {2023},
  author = {{NEON (National Ecological Observatory Network)}},
  publisher = {AmeriFlux AMP},
  doi = {10.17190/AMF/1579723},
  note = {Dataset},
  site = {US-xRM}
}

@misc{US-xRN,
  title = {{AmeriFlux BASE US-xRN NEON Oak Ridge National Lab (ORNL), Ver. 6-5}},
  year = {2023},
  author = {{NEON (National Ecological Observatory Network)}},
  publisher = {AmeriFlux AMP},
  doi = {10.17190/AMF/1773400},
  note = {Dataset},
  site = {US-xRN}
}

@misc{US-xSB,
  title = {{AmeriFlux BASE US-xSB NEON Ordway-Swisher Biological Station (OSBS), Ver. 7-5}},
  year = {2023},
  author = {{NEON (National Ecological Observatory Network)}},
  publisher = {AmeriFlux AMP},
  doi = {10.17190/AMF/1671899},
  note = {Dataset},
  site = {US-xSB}
}

@misc{US-xSC,
  title = {{AmeriFlux BASE US-xSC NEON Smithsonian Conservation Biology Institute (SCBI), Ver. 7-5}},
  year = {2023},
  author = {{NEON (National Ecological Observatory Network)}},
  publisher = {AmeriFlux AMP},
  doi = {10.17190/AMF/1671900},
  note = {Dataset},
  site = {US-xSC}
}

@misc{US-xSE,
  title = {{AmeriFlux BASE US-xSE NEON Smithsonian Environmental Research Center (SERC), Ver. 8-5}},
  year = {2023},
  author = {{NEON (National Ecological Observatory Network)}},
  publisher = {AmeriFlux AMP},
  doi = {10.17190/AMF/1617734},
  note = {Dataset},
  site = {US-xSE}
}

@misc{US-xSJ,
  title = {{AmeriFlux BASE US-xSJ NEON San Joaquin Experimental Range (SJER), Ver. 7-5}},
  year = {2023},
  author = {{NEON (National Ecological Observatory Network)}},
  publisher = {AmeriFlux AMP},
  doi = {10.17190/AMF/1671901},
  note = {Dataset},
  site = {US-xSJ}
}

@misc{US-xSL,
  title = {{AmeriFlux BASE US-xSL NEON North Sterling, CO (STER), Ver. 8-5}},
  year = {2023},
  author = {{NEON (National Ecological Observatory Network)}},
  publisher = {AmeriFlux AMP},
  doi = {10.17190/AMF/1617735},
  note = {Dataset},
  site = {US-xSL}
}

@misc{US-xSP,
  title = {{AmeriFlux BASE US-xSP NEON Soaproot Saddle (SOAP), Ver. 8-5}},
  year = {2023},
  author = {{NEON (National Ecological Observatory Network)}},
  publisher = {AmeriFlux AMP},
  doi = {10.17190/AMF/1617736},
  note = {Dataset},
  site = {US-xSP}
}

@misc{US-xSR,
  title = {{AmeriFlux BASE US-xSR NEON Santa Rita Experimental Range (SRER), Ver. 8-5}},
  year = {2023},
  author = {{NEON (National Ecological Observatory Network)}},
  publisher = {AmeriFlux AMP},
  doi = {10.17190/AMF/1579543},
  note = {Dataset},
  site = {US-xSR}
}

@misc{US-xST,
  title = {{AmeriFlux BASE US-xST NEON Steigerwaldt Land Services (STEI), Ver. 8-5}},
  year = {2023},
  author = {{NEON (National Ecological Observatory Network)}},
  publisher = {AmeriFlux AMP},
  doi = {10.17190/AMF/1617737},
  note = {Dataset},
  site = {US-xST}
}

@misc{US-xTA,
  title = {{AmeriFlux BASE US-xTA NEON Talladega National Forest (TALL), Ver. 7-5}},
  year = {2023},
  author = {{NEON (National Ecological Observatory Network)}},
  publisher = {AmeriFlux AMP},
  doi = {10.17190/AMF/1671902},
  note = {Dataset},
  site = {US-xTA}
}

@misc{US-xTL,
  title = {{AmeriFlux BASE US-xTL NEON Toolik (TOOL), Ver. 8-5}},
  year = {2023},
  author = {{NEON (National Ecological Observatory Network)}},
  publisher = {AmeriFlux AMP},
  doi = {10.17190/AMF/1617739},
  note = {Dataset},
  site = {US-xTL}
}

@misc{US-xTR,
  title = {{AmeriFlux BASE US-xTR NEON Treehaven (TREE), Ver. 8-5}},
  year = {2023},
  author = {{NEON (National Ecological Observatory Network)}},
  publisher = {AmeriFlux AMP},
  doi = {10.17190/AMF/1634886},
  note = {Dataset},
  site = {US-xTR}
}

@misc{US-xUK,
  title = {{AmeriFlux BASE US-xUK NEON The University of Kansas Field Station (UKFS), Ver. 8-5}},
  year = {2023},
  author = {{NEON (National Ecological Observatory Network)}},
  publisher = {AmeriFlux AMP},
  doi = {10.17190/AMF/1617740},
  note = {Dataset},
  site = {US-xUK}
}

@misc{US-xUN,
  title = {{AmeriFlux BASE US-xUN NEON University of Notre Dame Environmental Research Center (UNDE), Ver. 8-5}},
  year = {2023},
  author = {{NEON (National Ecological Observatory Network)}},
  publisher = {AmeriFlux AMP},
  doi = {10.17190/AMF/1617741},
  note = {Dataset},
  site = {US-xUN}
}

@misc{US-xWD,
  title = {{AmeriFlux BASE US-xWD NEON Woodworth (WOOD), Ver. 8-5}},
  year = {2023},
  author = {{NEON (National Ecological Observatory Network)}},
  publisher = {AmeriFlux AMP},
  doi = {10.17190/AMF/1579724},
  note = {Dataset},
  site = {US-xWD}
}

@misc{US-xWR,
  title = {{AmeriFlux BASE US-xWR NEON Wind River Experimental Forest (WREF), Ver. 8-5}},
  year = {2023},
  author = {{NEON (National Ecological Observatory Network)}},
  publisher = {AmeriFlux AMP},
  doi = {10.17190/AMF/1617742},
  note = {Dataset},
  site = {US-xWR}
}

@misc{US-xYE,
  title = {{AmeriFlux BASE US-xYE NEON Yellowstone Northern Range (Frog Rock) (YELL), Ver. 8-5}},
  year = {2023},
  author = {{NEON (National Ecological Observatory Network)}},
  publisher = {AmeriFlux AMP},
  doi = {10.17190/AMF/1617743},
  note = {Dataset},
  site = {US-xYE}
}

\end{document}